%% file: main_arxiv.tex
\RequirePackage{etoolbox}
\newtoggle{is_LNCS}
\togglefalse{is_LNCS}
\iftoggle{is_LNCS}{
  \documentclass[runningheads,envcountsame,orivec]{llncs}
}{
  \documentclass{article}
}

\input{preamble}

\title{Sparse PCA with Overlapping Supports: NP-Hardness and an FPTAS}
\iftoggle{is_LNCS}{
  \author{Anonymous Author(s)}
  \authorrunning{Anonymous Author(s)}
  \titlerunning{Sparse PCA with Overlapping Supports}
  \institute{}
  \hypersetup{hidelinks,pdfauthor={Anonymous Author(s)},pdftitle={Sparse PCA with Overlapping Supports: NP-Hardness and an FPTAS}}
}{
  \author{Alberto Del Pia\thanks{Department of Industrial and Systems Engineering \& Wisconsin Institute for Discovery, University of Wisconsin--Madison, Madison, WI, USA. E-mail: \href{mailto:delpia@wisc.edu}{\texttt{delpia@wisc.edu}}.} \and Dekun Zhou\thanks{E-mail: \href{mailto:dekun.zhou6@gmail.com}{\texttt{dekun.zhou6@gmail.com}}.}}
  \date{November 11, 2026}
  \hypersetup{colorlinks, linkcolor=blue, citecolor=blue, urlcolor=blue, pdftitle={Sparse PCA with Overlapping Supports: NP-Hardness and an FPTAS}, pdfauthor={Alberto Del Pia and Dekun Zhou}}
}

\begin{document}
\maketitle

\begin{abstract}
In sparse principal component analysis (SPCA) with several components, each must be sparse and all must be pairwise orthogonal.
When the supports of the components overlap, these two requirements interact.
For a covariance matrix of fixed rank, polynomial-time algorithms with optimality or approximation guarantees are known for one component, and for several components with disjoint supports or with a common support.
We study the general model, with individual sparsity constants and arbitrary support overlaps.
We are given a rational $n\times r$ matrix $R$ and $d$ sparsity constants $s_i \le n$, and the goal is to find an $n\times d$ matrix $X$ with orthonormal columns that maximizes $\fnorm{R^\top X}^2$, such that column $i$ has at most $s_i$ non-zero entries.
Here $RR^\top$ plays the role of the covariance matrix and has rank at most $r$.
We show that SPCA is NP-hard even when $r = d = 2$, but admits a fully polynomial-time approximation scheme (FPTAS) when $r$ and $d$ are fixed.
The returned solution has exactly orthonormal columns, and its entries are real algebraic numbers in an exact representation.
For fixed $\epsilon$, the number of steps of the algorithm is linear in $n$ up to a polylogarithmic factor.
Our approach reduces SPCA to a sparse version of the orthogonal Procrustes problem, that is, to maximizing a linear function over the feasible region of SPCA.
We show that this problem is NP-hard even for $d = 2$, and we give an FPTAS for it when $d$ is fixed.
This FPTAS combines approximate dynamic programming over the rows, two-column updates that restore exact orthogonality, and an exact computation of the few variables for which these updates are not available.
\end{abstract}

\noindent\textbf{Keywords:} Sparse PCA $\cdot$ Procrustes problem $\cdot$ Approximation scheme $\cdot$ NP-hardness $\cdot$ Approximate dynamic programming.

\input{sections/intro}
\input{sections/motivation}
\input{sections/guarantees}
\input{sections/acknowledgement}

{\small
\bibliographystyle{plainurl}
\bibliography{biblio}
}

\clearpage
\appendix
\input{sections/relatedwork}
\input{sections/algorithm}
\clearpage
\input{sections/appendix}

\end{document}

%% file: preamble.tex
\usepackage[toc,page]{appendix} 

\iftoggle{is_LNCS}{
  \usepackage[T1]{fontenc}
  \usepackage{amsmath,amsfonts,amssymb}
  \usepackage{newtxtext}
  \usepackage[varvw]{newtxmath}
  \usepackage[protrusion=false,expansion=false]{microtype}
  \DisableLigatures[i]{encoding=T1}
  \usepackage{dsfont,esvect,bm,graphicx,subfigure}
}{
  \usepackage{amsmath,amsthm,amsfonts,dsfont,esvect,bm,graphicx,subfigure,enumitem,amssymb}
}
\allowdisplaybreaks
\usepackage[utf8]{inputenc}
\usepackage[dvipsnames]{xcolor}
\usepackage[english]{babel}
\usepackage{optidef}
\usepackage[hypertexnames=false]{hyperref}
\iftoggle{is_LNCS}{\urlstyle{rm}}{}
\iftoggle{is_LNCS}{\PassOptionsToPackage{hanginginout}{algorithm2e}}{}
\usepackage[linesnumbered,ruled,vlined]{algorithm2e}
\usepackage{mathtools}
\DeclarePairedDelimiter\multiset{\{\mkern-4.5mu\{}{\}\mkern-4.5mu\}}

\usepackage{aliascnt}

\iftoggle{is_LNCS}{
\let\LNCSproof\proof
\let\endLNCSproof\endproof
\renewenvironment{proof}[1][Proof]{%
  \def\proofname{#1}%
  \LNCSproof
}{\qed\endLNCSproof}
\csundef{claim}
\csundef{endclaim}
\spnewtheorem{claim}[theorem]{Claim}{\bfseries}{\itshape}
\spnewtheorem{observation}[theorem]{Observation}{\bfseries}{\itshape}
\spnewtheorem{assumption}[theorem]{Assumption}{\bfseries}{\itshape}
\spnewtheorem{fact}[theorem]{Fact}{\bfseries}{\itshape}
\spnewtheorem{model}{Model}{\bfseries}{\itshape}
\csundef{remark}
\csundef{endremark}
\spnewtheorem*{remark}{Remark}{\itshape}{\rmfamily}
}{

\newcommand{\newaliasedtheorem}[3]{%
	\newaliascnt{#1}{theorem}%
	\newtheorem{#1}[#1]{#2}%
	\aliascntresetthe{#1}%
}

\newaliasedtheorem{corollary}{Corollary}{Corollaries}
\newaliasedtheorem{lemma}{Lemma}{Lemmas}
\newaliasedtheorem{claim}{Claim}{Claims}
\newaliasedtheorem{observation}{Observation}{Observations}
\newaliasedtheorem{assumption}{Assumption}{Assumptions}
\newaliasedtheorem{conjecture}{Conjecture}{Conjectures}
\newaliasedtheorem{proposition}{Proposition}{Propositions}
\newaliasedtheorem{definition}{Definition}{Definitions}
\newaliasedtheorem{fact}{Fact}{Facts}
\newaliasedtheorem{example}{Example}{Examples}

}

\usepackage[anycase]{thmtools}

\DeclareMathOperator{\tr}{tr}
\DeclareMathOperator{\rt}{\textup{RT}}
\DeclareMathOperator{\cp}{\textup{CP}}
\DeclareMathOperator{\locc}{\mathcal{L}_{occ}}

\DeclareMathOperator{\rk}{rank}
\DeclareMathOperator{\sp1}{Span}

\DeclareMathOperator{\sign}{sign}
\DeclareMathOperator{\supp}{Supp}

\DeclareMathOperator{\mo}{\mathcal{O}}
\DeclareMathOperator{\dso}{\mathds{1}}
\DeclareMathOperator{\sk}{\textup{key}}
\newcommand{\opt}{\mathsf{OPT}}
\newcommand{\optlin}{\mathsf{OPT}_{\rm linear}}

\newcommand{\R}{\mathbb{R}}
\newcommand{\Q}{\mathbb{Q}}
\newcommand{\Z}{\mathbb{Z}}

\newcommand{\abs}[1]{\left\lvert#1\right\rvert}
\newcommand{\norm}[1]{\left\lVert#1\right\rVert_2}
\newcommand{\nnorm}[1]{\left\lVert#1\right\rVert_*}
\newcommand{\znorm}[1]{\left\lVert#1\right\rVert_0}
\newcommand{\onorm}[1]{\left\lVert#1\right\rVert_1}
\newcommand{\fnorm}[1]{\left\lVert#1\right\rVert_F}

\newcommand{\infnorm}[1]{\lVert#1\rVert_{\infty}}

\def\md{{\mathcal D}}
\def\mf{{\mathcal F}}
\def\ml{{\mathcal L}}
\def\mv{{\mathcal V}}

\def\mk{\mathbb{K}}

\SetKwInOut{Input}{Input}
\SetKwInOut{Output}{Output}
\SetKwFor{ForEach}{for each}{do}{end for}
\SetKwFor{For}{for}{do}{end for}
\SetKwIF{If}{ElseIf}{Else}{if}{then}{else if}{else}{end if}
\SetKw{Return}{return}
\SetKw{Continue}{Continue}

\newcommand{\LinearOracle}{\textup{\textsc{LinearOracle}}}
\newcommand{\FixHardVariables}{\textup{\textsc{FixHardVariables}}}
\newcommand{\SelectBranch}{\textup{\textsc{SelectBranch}}}
\newcommand{\ExactFinalize}{\textup{\textsc{ExactFinalize}}}
\newcommand{\KeepBest}{\textup{\textsc{KeepBest}}}
\newcommand{\LeafInit}{\textup{\textsc{LeafInit}}}
\newcommand{\Partition}{\textup{\textsc{Partition}}}

\newcommand{\LNCSmath}[2]{\iftoggle{is_LNCS}{#1}{#2}}

\usepackage{color}
\iftoggle{is_LNCS}{\csundef{note}\csundef{endnote}}{}

\iftoggle{is_LNCS}{}{\usepackage[margin=1in]{geometry}}

\iftoggle{is_LNCS}{
  \usepackage{letltxmacro}
  \SetAlFnt{\small}
  \SetAlCapFnt{\small}
  \SetAlCapNameFnt{\small}
  \SetAlgoNlRelativeSize{-1}
  \SetInd{0.2em}{0.4em}
  \SetAlgoSkip{}
  \LetLtxMacro{\LNCSalgorithm}{\algorithm}
  \let\endLNCSalgorithm\endalgorithm
  \renewenvironment{algorithm}[1][htbp]{%
    \ifstrequal{#1}{H}{\LNCSalgorithm[!htb]}{\LNCSalgorithm[#1]}%
  }{\endLNCSalgorithm}
  \SetKwInput{Input}{Input}
  \SetKwInput{Output}{Output}
  \let\LNCSHeaderInput\Input
  \renewcommand{\Input}[1]{\def\LNCSInputText{#1}}
  \renewcommand{\Output}[1]{%
    \LNCSHeaderInput{\LNCSInputText\quad\KwSty{Output:} #1}%
  }
  \AtBeginDocument{\RestyleAlgo{plain}\SetAlgoNoLine\SetAlgoNoEnd}
}{}

\makeatletter
\iftoggle{is_LNCS}{
  \newsavebox{\LNCSAlgorithmFrameBox}
  \let\LNCSOriginalMakeTheAlgo\algocf@makethealgo
  \renewcommand{\algocf@makethealgo}{%
    \sbox{\LNCSAlgorithmFrameBox}{\LNCSOriginalMakeTheAlgo}%
    \hbox{%
      \rlap{%
        \raisebox{\dimexpr\ht\LNCSAlgorithmFrameBox+1.5pt\relax}
          [0pt][0pt]{\rule{\wd\LNCSAlgorithmFrameBox}{0.4pt}}%
      }%
      \rlap{%
        \raisebox{\dimexpr-\dp\LNCSAlgorithmFrameBox-1.9pt\relax}
          [0pt][0pt]{\rule{\wd\LNCSAlgorithmFrameBox}{0.4pt}}%
      }%
      \usebox{\LNCSAlgorithmFrameBox}%
    }%
  }
}{}
\makeatother

%% file: sections/intro.tex
\section{Introduction}
\label{sec:intro}
Principal component analysis (PCA) finds a few orthogonal directions, called \emph{components}, that capture as much of the variance of a dataset as possible.
Sparse PCA (SPCA), in addition, limits the number of features used by each component, making the components easier to interpret.
With several components, sparsity and orthogonality interact through overlapping supports.
Enforcing orthogonality between two components may disrupt their orthogonality with other components, while standard orthogonalization may violate the sparsity constraints.

Already for one component, SPCA is NP-hard and admits no fully polynomial-time approximation scheme (FPTAS) unless $\mathrm P=\mathrm{NP}$~\cite{Mag17}, and it is hard to approximate within any constant factor under stronger complexity assumptions~\cite{Mag17,ChaPapRub16}.
This motivates structural assumptions on the covariance matrix, such as a fixed rank.
Fixed rank is also relevant for approximating SPCA on general covariance matrices.
Indeed, for a single component, applying an algorithm for low-rank matrices to a low-rank approximation of the covariance matrix gives approximation guarantees that depend on the decay of the spectrum of the covariance matrix~\cite{PapDimKor13}.

For a covariance matrix of fixed rank, polynomial-time algorithms with optimality or approximation guarantees are known for one component, and for several components with disjoint supports or with a common support.
Asteris, Papailiopoulos, and Karystinos~\cite{AstPapKar14} solve single-component SPCA exactly in polynomial time for fixed rank.
For components with disjoint supports, orthogonality holds automatically.
In this case, Asteris, Papailiopoulos, Kyrillidis, and Dimakis~\cite{AstPapKyrDim15arXiv} give a $(1-\epsilon)$-approximation algorithm for every $\epsilon\in(0,1)$, whose running time is polynomial in the input size and in $1/\epsilon$ for fixed rank and number of components.
Del Pia~\cite{dP23aMPA} solves this case exactly in polynomial time for fixed rank and number of components.
For fixed rank, the same paper also solves exactly row-sparse PCA, where all components are supported on a common set of at most $k$ rows, for a given integer $k$.
Without a rank assumption, Cory-Wright and Pauphilet~\cite{CorPau26} study the general model with overlapping supports through semidefinite, Lagrangian, and combinatorial relaxations, and obtain computational methods with instance-specific optimality guarantees.

The general model with individual sparsity constants and arbitrary support overlaps remains open even for fixed rank.
Indeed, restricting the rank simplifies the objective but does not remove the support-dependent orthogonality constraints.
We ask whether fixing the rank and the number of components makes this model tractable.
We show that the answer is negative for exact optimization and positive for approximation.

\paragraph{Notation.}
For a positive integer $m$, let $[m]:=\{1,2, \ldots,m\}$.
Denote the \emph{support} of a vector $x$ by $\supp(x):=\{i:x_i\ne0\}$, and let $\znorm{x}:=|\supp(x)|$.
Denote by $\mathbf{e}_i$ the $i$-th standard basis vector.
We denote by $\dso_J$ the indicator vector of a set $J$ and by $\dso_{\{P\}}$ the indicator of a condition $P$.
For a matrix $X$, $X_i$ denotes its $i$-th column and $X_{J,T}$ the submatrix with row indices $J$ and column indices $T$.
We denote by $\norm{\cdot}$ the 2-norm for vectors and the spectral norm for matrices, and by $\fnorm{\cdot}$ the Frobenius norm.
We denote the trace by $\tr(\cdot)$, the identity matrix of order $m$ by $I_m$, and a zero matrix by $O$.

\paragraph{The problem.}
Let $R \in \Q^{n\times r}$, let $d\le n$ be the number of components, and let $s=(s_1,\ldots,s_d)\in [n]^d$.
The integers $s_i$ are the \emph{sparsity constants}.
We study the problem
\begin{equation}
	\label{eqn:SPCA}
	\tag{SPCA}
	\opt(R; s):=
	\max_{X\in \R^{n\times d}} \fnorm{R^\top X}^2 \quad \text{s.t.} \quad X^\top X = I_d, \quad \znorm{X_i} \le s_i, \quad i\in [d].
\end{equation}
Thus, $s_i$ is the maximum number of non-zero entries of the $i$-th component $X_i$.
We denote by $\mf_s$ the feasible region of \ref{eqn:SPCA}.
Since $\fnorm{R^\top X}^2=\tr(X^\top C X)$ with $C:=RR^\top$, the matrix $C$ plays the role of the covariance matrix, and its rank is at most $r$.

\paragraph{Main results.}
Our first result shows that \ref{eqn:SPCA} is NP-hard even for $r=d=2$:
\begin{restatable}{theorem}{npHardnessTheorem}
	\label{thm:NPhard}
	Given $R\in \Z^{n\times 2}$, $k\in [n]$, and $\tau_0\in \Z$, it is NP-hard to decide whether $\opt(R; (k, k)) \ge \tau_0$.
\end{restatable}
Theorem~\ref{thm:NPhard} cannot be strengthened by lowering $r$ or $d$.
For $d=1$ and fixed $r$, \ref{eqn:SPCA} can be solved exactly in polynomial time~\cite{AstPapKar14,dP23aMPA}.
For $r=1$, \ref{eqn:SPCA} can be solved in polynomial time for every $d$ (see Observation~\ref{obs:rank_one} in Appendix~\ref{sec:appendix}).

We prove Theorem~\ref{thm:NPhard} by a reduction from $\Partition$, which shows how overlapping supports create hardness.
Let $m$ be the number of integers in the given instance of $\Partition$.
In the constructed instance of \ref{eqn:SPCA}, one component must have equal entries on $m+1$ rows and be orthogonal to a vector that encodes these integers.
Hence the entries of this vector in these rows sum to zero, and these rows give a partition.

Our main result is an FPTAS.
\begin{restatable}{theorem}{fptasTheorem}
	\label{thm:FPTAS}
	For fixed $r$ and $d$, there is an algorithm that, given $R\in\Q^{n\times r}$, $s\in[n]^d$, and $\epsilon\in\Q\cap(0,1)$, returns $X\in\mf_s$ with $\fnorm{R^\top X}^2 \ge (1-\epsilon) \opt(R;s)$, in time polynomial in the encoding size of $(R,s,\epsilon)$ and in $1/\epsilon$.
\end{restatable}
Since this reduction is from $\Partition$, Theorems~\ref{thm:NPhard} and~\ref{thm:FPTAS} give the same picture as for the knapsack problem: for fixed $r$ and $d$, \ref{eqn:SPCA} is NP-hard in the weak sense, and it admits an FPTAS.

The matrix returned by the algorithm of Theorem~\ref{thm:FPTAS} has exactly orthonormal columns.
All running times are in the bit model.
The input is rational, and the entries of the output are real algebraic numbers.
The algorithm outputs them exactly, in a standard finite encoding by an integer polynomial and an isolating interval (Appendix~\ref{app:complexity}).
For fixed $r$ and $d$, its bit complexity is at most $n\,\epsilon^{-\mo(rd+d^3)}\log^{\mo(d^2)}(2n/\epsilon)$ times a polynomial in the encoding size of $(R,s,\epsilon)$.
For fixed $r$, $d$, and $\epsilon$, the number of steps of the algorithm is nearly linear in $n$, that is, linear in $n$ up to a polylogarithmic factor (see Section~\ref{sec:guarantee}).
Theorems~\ref{thm:main_approximation} and~\ref{thm:main_alg_complexity} in Section~\ref{sec:guarantee} give the precise approximation guarantee and bit complexity.

\paragraph{The linear problem.}
Inspired by the auxiliary-variable and net approach of Asteris et al.~\cite[Lemma~9.2]{AstPapKyrDim15arXiv}, we use Frobenius-norm duality to reduce \ref{eqn:SPCA} to maximizing a linear function over $\mf_s$.
Specifically, by the Cauchy--Schwarz inequality,
\begin{equation*}
	\fnorm{R^\top X}=\max\{\tr(W^\top R^\top X): W\in\R^{r\times d},\, \fnorm{W}\le 1\}.
\end{equation*}
Hence, up to a factor $1-\epsilon$, \ref{eqn:SPCA} reduces to maximizing $(rd/\epsilon)^{\mo(rd)}$ linear functions $\tr(A^\top X)$ over $\mf_s$, with $A=RW$ for $W$ in a grid.
For $A\in\Q^{n\times d}$, let
\begin{equation}
	\label{eqn:sparse_linear}
	\optlin(A;s):=\max_{X\in \mf_s} \tr(A^\top X).
\end{equation}
For disjoint supports, the auxiliary problem of Asteris et al.~\cite{AstPapKyrDim15arXiv} maximizes a sum of squared projections and reduces to bipartite matching. 
Our reduction instead uses the linear objective in \eqref{eqn:sparse_linear}.
Problem~\eqref{eqn:sparse_linear} is a sparse version of the orthogonal Procrustes problem.
Without the sparsity constraints, an optimal solution is obtained from a singular value decomposition of $A$ (see~\cite[Theorem~4]{ZouHasTib06}), and Procrustes reformulations are used in algorithms for orthogonal sparse PCA~\cite{BenSunBabPal16}.
With overlapping supports, the linear problem already contains the interaction between sparsity and orthogonality.
Our next result shows that, unlike the classical problem, \eqref{eqn:sparse_linear} is NP-hard even for $d=2$.
\begin{restatable}{theorem}{linearHardnessTheorem}
	\label{thm:NP-hard_linear}
	Given $A\in \Z^{n\times 2}$, $k\in [n]$, and $\tau_0\in \Z$, it is NP-hard to decide whether $\optlin(A;(k, k)) \ge \tau_0$.
\end{restatable}
The proof of Theorem~\ref{thm:NP-hard_linear} is similar to that of Theorem~\ref{thm:NPhard}.
In the exact algorithms of Del Pia~\cite{dP23aMPA}, an auxiliary matrix with $r$ rows plays a role similar to that of $W$, and the space of these matrices is partitioned exactly by a hyperplane arrangement.
Here, by Theorem~\ref{thm:NP-hard_linear}, the problem for a fixed $W$ is already NP-hard, so the fundamental difficulty is not in enumeration of $W$, and we discretize this space directly.
Our main technical result is an FPTAS for the linear problem when $d$ is fixed.
\begin{restatable}{theorem}{linearFptasTheorem}
	\label{thm:linear_FPTAS}
	For fixed $d$, there is an algorithm that, given $A\in\Q^{n\times d}$, $s\in[n]^d$, and $t\in\Q\cap(0,1)$, returns $X\in\mf_s$ with $\tr(A^\top X) \ge (1-t) \optlin(A;s)$, in time polynomial in the encoding size of $(A,s,t)$ and in $1/t$.
\end{restatable}
As for Theorem~\ref{thm:FPTAS}, for fixed $d$ and $t$, the number of steps of this algorithm is linear in $n$ up to a polylogarithmic factor.
Theorem~\ref{thm:linear_FPTAS} holds for every $A\in\Q^{n\times d}$, so it gives an FPTAS for this sparse version of the Procrustes problem with a fixed number of columns, which may be of independent interest.
In our algorithm for \ref{eqn:SPCA}, fixing $r$ is needed only to bound the number of matrices $W$ in the grid.
Theorem~\ref{thm:FPTAS} follows from Theorem~\ref{thm:linear_FPTAS} and the reduction above.

\paragraph{Our techniques.}
The objective $\tr(A^\top X)$ and the sparsity constraints decompose over the rows of $X$, while the constraint $X^\top X=I_d$ couples all rows.
The algorithm of Theorem~\ref{thm:linear_FPTAS} handles this coupling with three ingredients.

(a) \textit{Approximate dynamic programming.}
For every row, we choose the set of columns that may be non-zero in that row, called its row type, and the values of its entries.
These choices are made by an approximate dynamic program (ADP) over a balanced binary tree whose leaves are the rows.
Each state records rounded upper bounds on the number of rows in which each column may be non-zero, and the Gram matrix $X^\top X$ rounded relative to its trace.
Since the tree has depth $\lceil\log_2 n\rceil$, the rounding errors accumulate over only logarithmically many levels.
For fixed $d$ and $t$, the number of states at each node is polylogarithmic in $n$.
The candidate solutions at the root are approximately orthogonal.

(b) \textit{Exact orthogonality by two-column updates.}
Orthogonalizing the candidate solutions at the root, for example by the Gram--Schmidt process, would in general violate the sparsity constraints.
Suppose that two columns $i$ and $l$ may both be non-zero in at least $d$ common rows.
In these rows, the other $d-2$ columns span a space of dimension at most $d-2$.
Hence the vectors supported on these rows that are orthogonal to all the other columns form a space of dimension at least two.
Changing columns $i$ and $l$ inside this space makes them orthogonal, does not change their inner products with the other columns, and does not decrease the objective (Proposition~\ref{prop:two_column_correctness}).
A sequence of such updates, followed by a normalization, turns an approximately orthogonal candidate solution into a point in $\mf_s$.

(c) \textit{Hard pairs.}
Pairs of columns with fewer than $d$ common rows cannot be repaired in this way, and we call them hard.
The number of variables involved in hard pairs is at most $d(d-1)^2$, independently of $n$.
We fix the values of these variables before running the ADP.
To this end, we cover the space of these variables by small boxes.
In each box, we compute an exact solution of the orthogonality constraints of the hard pairs, if one exists, with algorithms from real algebraic geometry~\cite{BasPolRoy06,BasLerRoy09}.

\paragraph{Organization.}
Section~\ref{sec:motivation} describes the algorithm for the linear problem, and Section~\ref{sec:guarantee} states the approximation guarantee, sketches its proof, and bounds the running time.
Appendix~\ref{sec:related_work} discusses further related work, Appendix~\ref{sec:pseudocode} contains the pseudocode, and Appendix~\ref{sec:appendix} contains all proofs.

%% file: sections/motivation.tex
\section{The algorithm for the linear problem}
\label{sec:motivation}
\label{sec:algorithm}
This section describes the algorithm $\LinearOracle$ of Theorem~\ref{thm:linear_FPTAS} for the linear problem~\eqref{eqn:sparse_linear}, with a fixed objective matrix $A\in\Q^{n\times d}$.
Our algorithm for \ref{eqn:SPCA} calls $\LinearOracle$ with accuracy $t=\epsilon/4$ and with $A=RW$ for every matrix $W$ in an $(\epsilon/4)$-net of $\{W\in\R^{r\times d}: \fnorm{W}\le 1\}$, and returns the best solution found (Algorithm~\ref{alg:main_alg} in Appendix~\ref{sec:pseudocode}).
Table~\ref{tab:parameters} lists the parameters of $\LinearOracle$ for an accuracy $t$, and Appendix~\ref{sec:pseudocode} gives the pseudocode of its subroutines.
\begin{table}[t]
	\caption{Parameters of $\LinearOracle$ for accuracy $t$, where $n$ is the number of rows and $d$ the number of columns of $A$.}
	\label{tab:parameters}
	\centering
	\renewcommand{\arraystretch}{1.15}
	\begin{tabular}{lll}
		\hline
		Parameter & Value & Role\\
		\hline
		$\delta$ & $t/(8(d+1))$ & error allowed for $\fnorm{X^\top X-I_d}$ at the root\\
		$D_d$ & $d(d-1)^2$ & upper bound on the number $D$ of hard variables\\
		$a$ & $\delta/(256(d+1)(D_d+1))$ & upper bound on the side length of the boxes\\
		$J_p$ & $\lceil 2/a\rceil$ & number of intervals covering $[-1,1]$\\
		$M$ & $\lceil 256d(d+1)/\delta\rceil$ & inverse of relative mesh size in $\mv$\\
		$J_0$ & $\lceil \log_2 (256dn(d+1)/\delta)\rceil$ & determines the smallest non-zero magnitude $2^{-J_0}$ in $\mv$\\
		$H_{\rm tree}$ & $\lceil\log_2 n\rceil$ & height of the ADP tree\\
		$\tau$ & $\delta/(128d^2)$ & slack in the sparsity capacity usages\\
		$P_d$ & $d\binom d2$ & upper bound on the number of kept rows in the analysis\\
		$K_0$ & $\lceil 16(P_d+1)/\tau\rceil$ & largest capacity cap that is not rounded\\
		$\alpha$ & $\tau/(8(H_{\rm tree}+1))$ & the geometric part of $U_c$ has ratio $1+\alpha$\\
		$\theta$ & $\delta/(128d^2(H_{\rm tree}+1))$ & relative width of the Gram matrix bins\\
		$\kappa_G$ & $\delta/(128n(H_{\rm tree}+1))$ & trace threshold of the small-trace bin\\
		\hline
	\end{tabular}
\end{table}

\subsection{Row types and column pairs}
\label{sec:motivation_row_type}

A \emph{candidate solution} is a matrix in $\R^{n\times d}$ constructed by our algorithm, which may violate the constraint $X^\top X=I_d$.
We decompose a candidate solution according to the set of columns that may be non-zero in each row.

\begin{definition}
	A \emph{row type} $T$ is a subset of $[d]$, indicating which entries of a row may be non-zero.
\end{definition}
Our algorithm fixes, row by row, which entries of a candidate solution may be non-zero, by assigning a row type to each row.
The same row type can be assigned to several rows.
There are only $2^d$ row types, a number that does not depend on $n$.
For a candidate solution $X$ whose $j$-th row is assigned a row type, we denote this row type by $\rt_j(X)$.
For a candidate solution $X$ with all its rows assigned row types, we classify the pairs of columns by the number of assigned row types that contain them.
\begin{definition}
	Let $X$ be a candidate solution with all its rows assigned row types, and let $\cp_d:= \{\{i, l\}: 1\le i < l \le d\}$ be the set of \emph{column pairs}.
	For every column pair $e \in \cp_d$, let $m_e(X):=\abs{ \{j\in [n]: e\subseteq \rt_j(X)\} }$ be the number of rows whose assigned row type contains $e$.
	The \emph{hard pairs} are the pairs in $H(X):=\{e\in \cp_d: m_e(X) < d\}$, and the \emph{simple pairs} are the pairs in $S(X):= \cp_d \setminus H(X)$.
\end{definition}
Our algorithm treats the rows whose assigned row types contain a hard pair separately, so we collect their row types in a list.
\begin{definition}
\label{def:listhardrowtype}
For a candidate solution $X$ with all its rows assigned row types, the \emph{list of hard row types} $\ml(X)$ is the multiset of the assigned row types of $X$ that contain a hard pair.
Let $b(X):=\abs{\ml(X)}$.
The elements of $\ml(X)$ have labels $1,\ldots,b(X)$, one for each row whose assigned row type contains a hard pair, and we write $\ml(X) = \multiset{T^{1}, T^{2}, \ldots, T^{b(X)}}$. 
\end{definition}
The rows whose assigned row types are in $\ml(X)$ are the \emph{hard rows}.
The \emph{hard variables} of $X$ are the entries $(j,i)$ such that $i\in e\subseteq\rt_j(X)$ for some $e\in H(X)$, and $D(X)$ denotes their number.
Every hard row contains a hard pair, and every hard pair is contained in the assigned row types of at most $d-1$ rows.
Hence $b(X)\le(d-1)\abs{H(X)}$ and $D(X)\le 2(d-1)\abs{H(X)}\le d(d-1)^2$ (Lemma~\ref{lem:number_hard_row} in Appendix~\ref{sec:appendix}).

The two-column update of Section~\ref{sec:motivation_two_column} makes a simple pair $\{i,l\}$ orthogonal using $d$ rows whose assigned row types contain $\{i,l\}$, and a hard pair does not have that many rows.
We therefore fix the values of the hard variables in advance.
Our algorithm determines the hard pairs, the simple pairs, and the list of hard row types before the values of $X$.
It first enumerates the sets $H$ and $S$ and the list $\ml$, and then fixes the values of the hard variables.
After that, it assigns the row types in $\ml$ to $b$ rows of the candidate solution, assigns row types without hard pairs to the remaining rows, and finally assigns values to the entries.
Therefore, in the rest of the paper, we drop the dependence on $X$ from the notation of this section.

\subsection{Two-column updates for exact orthogonality}
\label{sec:motivation_two_column}
Let $X$ be a candidate solution, and let $e = \{i, l\}\in S$ be a simple pair.
By modifying at most $2d$ entries of the columns $X_i$ and $X_l$, we obtain a matrix $\bar X$ such that
(a) $\bar X_i^\top \bar X_l = 0$, (b) $\bar X_i^\top X_p = X_i^\top X_p$ and $\bar X_l^\top X_p = X_l^\top X_p$ for every $p\in [d]\setminus\{i, l\}$, and (c) $\tr(A^\top \bar X) \ge \tr(A^\top X)$.
The modified entries lie in $d$ rows whose assigned row types contain $e$.
As explained in Section~\ref{sec:intro}, the vectors supported on these rows that are orthogonal to all other columns form a space of dimension at least two.
We change columns $i$ and $l$ only inside this space, which gives (b).
To obtain (a) and (c), we choose the new entries by solving a problem of the form
\begin{equation}
	\label{eqn:rotation_PCA}
	\max\left\{\tr(\Gamma^\top Z): Z\in \R^{h\times k},\ Z^\top Z = \Lambda\right\},
\end{equation}
where $\Gamma\in \R^{h\times k}$ with $k\le h$, and $\Lambda\in \R^{k\times k}$ is positive semidefinite.
By von Neumann's trace inequality~\cite{Mir75}, the optimal value of~\eqref{eqn:rotation_PCA} is $\tr(\sqrt{\Gamma\Lambda\Gamma^\top})$, and an optimal solution can be obtained from a singular value decomposition of $\Gamma\sqrt{\Lambda}$ (Proposition~\ref{prop:rotation_PCA} in Appendix~\ref{sec:appendix}).
Algorithm~\ref{alg:two_column} applies this update to all simple pairs, one at a time.

\begin{algorithm}[H]
	\caption{Repeated two-column updates}
	\label{alg:two_column}
	
	\Input{
		Set of simple pairs $S$, candidate solution $X$, objective matrix $A$, and row types $\{\rt_j\}_{j\in [n]}$.
	}
	\Output{
		$\bar X$ such that $\bar X^\top\bar X$ is diagonal and $\tr(A^\top \bar X) \ge \tr(A^\top X)$ under the assumptions of Proposition~\ref{prop:two_column_correctness}.
	}
	
	$Y\gets X$, $G\gets X^\top X$\;
	
	\ForEach{$e=\{i, l\}\in S$, in lexicographic order}{
		Choose any $d$ assigned rows containing $e$: $J_e\subseteq\{j:e\subseteq \rt_j\}$ with $|J_e| = d$\;
		
		$g\gets G_{il}$\;
		
		\If{$g\neq0$}{
			$D_e\gets |g|
			\begin{pmatrix}
				1 & -\operatorname{sgn}(g)\\
				-\operatorname{sgn}(g) & 1
			\end{pmatrix}$, 
			$Z\gets Y_{J_e,e}$,
			$B \gets Y_{J_e,[d]\setminus e}$\;
			
			Compute $N\in\R^{d\times 2}$ with $N^\top N=I_2$ and $B^\top N=O$\;
			
			$K\gets N^\top Z$, $Z_0\gets Z-NK$, $\Lambda_e\gets K^\top K+D_e$, $\Gamma_e\gets N^\top A_{J_e,e}$\;
			
			$K_{\rm new}\gets$ an optimal solution of \eqref{eqn:rotation_PCA} with $\Lambda= \Lambda_e$ and $\Gamma= \Gamma_e$, e.g., the one in Proposition~\ref{prop:rotation_PCA} or in Lemma~\ref{lem:runtime_K}\;
			
			$Y_{J_e,e} \gets Z_0+NK_{\rm new}$, $E_e\gets (\mathbf{e}_i, \mathbf{e}_l) \in \R^{d\times 2}$, $G\gets G+E_eD_eE_e^\top$\;
		}
	}
	
	\Return{$Y$}\;
\end{algorithm}
The next proposition shows that Algorithm~\ref{alg:two_column} is correct.
\begin{restatable}{proposition}{twoColumnCorrectnessProposition}
	\label{prop:two_column_correctness}
	Let $H:=\cp_d\setminus S$.
	Let $X$ satisfy $X_i^\top X_l = 0$ for every $\{i, l\} \in H$, and let every $e\in S$ satisfy $\abs{\{j: e\subseteq \rt_j\}}\ge d$.
	Let $\bar X$ be the output of Algorithm~\ref{alg:two_column} with input $(S, X, A, \{\rt_j\}_{j\in [n]})$.
	Then $\bar X^\top \bar X$ is diagonal with $(\bar X^\top \bar X)_{ii}=(X^\top X)_{ii}+\sum_{l\ne i}\abs{(X^\top X)_{il}}$ for every $i\in[d]$.
	Moreover, $\bar X_{jp}=X_{jp}$ for every $j\in[n]$ and $p\notin\rt_j$, and $\tr(A^\top \bar X) \ge \tr(A^\top X)$.
\end{restatable}

\paragraph{Proof idea of Proposition~\ref{prop:two_column_correctness}.}
Consider the iteration of a simple pair $e=\{i,l\}$ with $g\ne0$, and assume that $g=Y_i^\top Y_l$.
Since $B^\top$ has $d-2$ rows, its kernel has dimension at least $2$, so the matrix $N$ exists.
The iteration replaces the component $NK$ of $Z$ by $NK_{\rm new}$ and keeps $Z_0$.
Since $B^\top N=O$, the matrix $B^\top Z$ does not change.
Hence the inner products of columns $i$ and $l$ with the other columns do not change.
Since $N^\top N=I_2$ and $N^\top Z_0=O$, the matrix $Z^\top Z$ changes from $Z_0^\top Z_0+K^\top K$ to $Z_0^\top Z_0+\Lambda_e=Z_0^\top Z_0+K^\top K+D_e$.
By the choice of $D_e$, columns $i$ and $l$ become orthogonal, and their squared norms increase by $\abs{g}$.
For the objective, note that $\Lambda_e\succeq K^\top K$.
Thus, by Proposition~\ref{prop:rotation_PCA} and Weyl's monotonicity theorem, the optimal value of~\eqref{eqn:rotation_PCA} with $\Lambda=\Lambda_e$ is at least the optimal value with $\Lambda=K^\top K$.
Since $K$ is feasible for the latter problem, we obtain $\tr(\Gamma_e^\top K_{\rm new})\ge\tr(\Gamma_e^\top K)$, and the objective does not decrease.
Finally, the entry $(i,l)$ of the Gram matrix changes only in the iteration of $\{i,l\}$, where it becomes zero.
The entries of the hard pairs are zero and never change, so $\bar X^\top\bar X$ is diagonal.
The full proof is in Appendix~\ref{app:proofs_section2}. \qed

\subsection{Fixing the hard variables}
\label{sec:fix_hard_variables}
Fix a set $H$ of hard pairs and a list of hard row types $\ml=\multiset{T^1,\ldots,T^b}$.
For each $r\in[b]$, let $U_r := \cup_{e\in H, e\subseteq T^r} e$ be the set of indices in $T^r$ that belong to a hard pair contained in $T^r$.
The subroutine $\FixHardVariables$ (Algorithm~\ref{alg:fix_hard_variables} in Appendix~\ref{sec:pseudocode}) computes values $p_{ri}$, for $r\in[b]$ and $i\in U_r$, that satisfy the orthogonality constraints of the hard pairs, and we write $p_r:=(p_{ri})_{i\in U_r}$.
The ADP decides later which rows receive them.
The number of these variables is $D:= \sum_{r = 1}^b |U_r|$.
Since every $T^r$ contains a pair of $H$, and every pair of $H$ is contained in at most $d-1$ of the sets $T^1,\ldots,T^b$ (Algorithm~\ref{alg:linear_oracle}), the argument of Lemma~\ref{lem:number_hard_row} gives $D\le 2(d-1)|H|\le D_d$.
A solution of the orthogonality equations in $[-1,1]^D$, such as the zero solution, can be far from the hard variables of a \emph{target solution}, that is, an optimal solution of~\eqref{eqn:sparse_linear} that the algorithm aims to approximate.
We therefore cover $[-1,1]$ by the $J_p$ closed intervals $\Omega_\ell:=[-1+2(\ell-1)/J_p,\,-1+2\ell/J_p]$, $\ell\in[J_p]$, each of length $2/J_p\le a$, where $a$ and $J_p$ are given in Table~\ref{tab:parameters}.
For each box $Q=\Omega_{\ell_1}\times\cdots\times \Omega_{\ell_D}$, we decide whether $Q$ contains a tuple $p$ satisfying
\begin{equation}
	\label{eqn:hard_constraints}
	\sum_{r:\{i,k\}\subseteq T^r}p_{ri}p_{rk}=0, \quad \forall\,\{i,k\}\in H,
	\quad
	\sum_{r=1}^b\sum_{i\in U_r}p_{ri}^2\le d.
\end{equation}
Every $X\in\mf_s$ satisfies $\tr(X^\top X) = d$, so its hard variables satisfy the last inequality.
The bounded algebraic sampling algorithm of Basu, Leroy, and Roy~\cite{BasLerRoy09} solves this semialgebraic feasibility problem exactly (see Lemma~\ref{lem:hard_feas_complexity}).
If $p^*$ denotes the hard variables of a target solution, then $p^*$ lies in some box $Q^*$, and $\FixHardVariables$ returns a solution $p^{Q^*}$ in this box, with $\norm{p^{Q^*}-p^*} \le \sqrt{D} a$.

\subsection{Approximate dynamic programming}
\label{sec:motivation_ADP}
\label{sec:ADP}
Fix a set $H$ of hard pairs, the set $S:=\cp_d\setminus H$ of simple pairs, a list $\ml = \multiset{T^1,\ldots,T^b}$ of hard row types, and a vector $p$ returned by $\FixHardVariables$.
We call such a choice of $(H,\ml,p)$ a \emph{branch} of $\LinearOracle$.
For each branch, the subroutine $\SelectBranch$ (Algorithm~\ref{alg:select_branch} in Appendix~\ref{sec:pseudocode}) assigns row types and variable values to the rows of a candidate solution through an approximate dynamic programming (ADP) tree.
It keeps the values $p$ of the hard variables.
Its output consists of candidate solutions that satisfy the sparsity constraints and the orthogonality constraints of the hard pairs, and whose error $\fnorm{X^\top X - I_d}$ is small.

The ADP tree is a complete binary tree of height $H_{\rm tree}:=\lceil\log_2 n\rceil$ with $N_{\rm leaf}:=2^{H_{\rm tree}}$ leaves.
Leaves $1,\ldots,n$ correspond to the rows.
The remaining leaves are dummy leaves, which correspond to no row and contribute zero (Algorithm~\ref{alg:leaf_init} in Appendix~\ref{sec:pseudocode}).
A \emph{partial solution} at a node $u$ is a matrix $X_u$ whose rows in the subtree rooted at $u$ are assigned row types and values, and whose remaining rows are zero.
Let $G_u:=X_u^\top X_u$ and $v_u:= \tr(A^\top X_u)$.
The \emph{sparsity capacity usage} of column $i$ in $X_u$ is the number $c_{u,i}$ of rows of $X_u$ whose assigned row type contains $i$.
Thus $c_{u,i}$ counts the rows in which column $i$ is allowed to be non-zero, and it can exceed the number of non-zero entries assigned to column $i$.
The \emph{key} $\sk(X_u)$ of $X_u$ consists of the following terms:
\begin{itemize}
	\item $\locc$: the set of labels $r\in[b]$ such that the hard row type $T^r\in\ml$ is assigned to a row in the subtree of $u$.
	\item $\{\mu_e: e\in S\}$: $\mu_e := \min\{m_{u,e}, d\}$, where $m_{u,e}$ is the number of rows in the subtree of $u$ whose assigned row type contains the simple pair $e$.
	\item $\hat c$: a $d$-vector of upper bounds on the sparsity capacity usages $c_{u,i}$, defined below.
	\item $\hat G$: a bin label of the Gram matrix $G_u$, defined below.
\end{itemize}
At node $u$, we store a \emph{dictionary} $\md_u$ that maps $\sk(X_u)$ to the pair $(v_u, G_u)$.
For every key, the dictionary keeps the pair with the largest objective value found, and ties are broken by a fixed enumeration rule (Algorithm~\ref{alg:keep_best} in Appendix~\ref{sec:pseudocode}).
A \emph{state} at $u$ is an entry of $\md_u$, that is, a key together with its pair.
There are $\prod_{i = 1}^d (s_i + 1)$ possible vectors of sparsity capacity usages, and infinitely many possible Gram matrices $G_u$.
Inspired by Bertsimas and Demir~\cite{BerDem02}, we round both to reduce the number of states.

\paragraph{Capacity caps.} We limit the number of possible sparsity capacity usages by rounding them up.
We choose constants $K_0 > 0$ and $\alpha > 0$, whose values are given in Table~\ref{tab:parameters}.
We define a finite set $U_c$ as follows:
\begin{equation}
	\label{eqn:capacity_use}
	U_c := 
	\begin{cases} 
		\{0, 1, \ldots, n\}, & \text{if } K_0 \ge n, \\
		\{0, 1, \ldots, K_0\} \cup \{K_0(1 + \alpha)^l : l \in [l_{\max}]\}, & \text{otherwise,}
	\end{cases}
\end{equation}
where $l_{\max}$ is the smallest positive integer such that $K_0(1 + \alpha)^{l_{\max}} > n$.
At a leaf $u$ and for $i\in [d]$, the \emph{capacity cap} of column $i$ is
\begin{equation}
	\label{eqn:capacity_cap}
	\hat c_{u,i} = R_c(c_{u,i}) := \min \left\{ w\in U_c : w \ge c_{u,i} \right\}.
\end{equation}
At the merge of two nodes, we add the two caps and round the sum up with $R_c$.
We reject the merge if the sum, or its rounded value, is larger than $s_i$ for some $i\in[d]$.
As a result, for every node $u$, we have $c_{u, i} \le \hat c_{u, i}$, but in general $\hat c_{u, i}\ne R_c(c_{u, i})$.

\paragraph{Gram matrix bins.} We first use $\tr(G_u)$ to determine a scale, and then discretize every entry of $G_u$ at the scale.
More precisely, we choose constants $\kappa_G > 0$ and $\theta > 0$ as in Table~\ref{tab:parameters}, and define the following intervals: 
\begin{align*}
	I_{-1}& := [0, \kappa_G) \cap [0, 4d), 
	\quad I_0:= [\kappa_G, (1+\theta) \kappa_G) \cap [0, 4d), \quad  \ldots\\
	I_k& := [(1+\theta)^{k} \kappa_G , (1+\theta)^{k+1} \kappa_G) \cap [0, 4d), \quad \ldots
\end{align*}
Only finitely many of these intervals are nonempty.
For each partial solution $X_u$ with $\tr(G_u) < 4d$, we define the \emph{label mapping} $\mathrm{lab}(G_u)$ as follows.
If $\tr(G_u) \in I_{-1}$, then $\mathrm{lab}(G_u):=(-1, O)$.
If $\tr(G_u) \in I_k$ for some $k\ge 0$, then
\begin{equation*}
\mathrm{lab}(G_u):=\left(k, \left(\left\lfloor \frac{(G_u)_{il}}{ \theta \cdot \kappa_G (1+\theta)^{k}} \right\rfloor\right)_{1\le i, l \le d} \right).
\end{equation*}
The \emph{bin label} of $X_u$ is $\hat G(X_u):=\mathrm{lab}(G_u)$. 
Adding rows does not decrease the trace of the Gram matrix.
Hence partial solutions with trace at least $4d$ cannot be extended to approximately orthogonal candidate solutions in our algorithm, and we discard them.
The dictionaries store the exact Gram matrices, and the bin labels are used only to identify states.
For a fixed $k$, the number of possible labels for each Gram matrix entry is $\mo(1/\theta)$, as detailed in the workload analysis in Appendix~\ref{app:complexity}.

\paragraph{Leaves.}
For a row assigned to some row type $T$, we call the variables indexed by $T$ that are not hard variables \emph{free variables}.
Their values are chosen from the finite set of rational numbers
\begin{equation}
	\label{eqn:free_variable_set}
	\mv := \{0,\pm1\}
	\cup
	\LNCSmath{\{\pm2^{-k} (1+\ell/M): 1\le k\le J_0, 0\le\ell<M\}}{\left\{\pm2^{-k} \left(1+\frac{\ell}{M}\right):	1\le k\le J_0, 0\le\ell<M \right\}},
\end{equation}
where $M$ and $J_0$ are given in Table~\ref{tab:parameters}.
The subintervals of $[2^{-k},2^{-k+1}]$ defined by $\mv$ have length $2^{-k}/M$, so the spacing of $\mv$ is relative to the magnitude of its elements.
As a result, each entry $z$ of a target solution with $\abs{z}\ge 2^{-J_0}$ can be approximated by some $\hat z\in\mv$ with error at most $\abs{z}/M$.
Every other entry can be rounded to $0$, with error at most $2^{-J_0}$.
For a row type $T$, let $U_H(T) := \cup_{e\in H, e\subseteq T} e$.
We call a row type $T$ \emph{ordinary} if it contains no hard pair, that is, if $U_H(T)=\emptyset$.
At a leaf $j\le n$, the ADP considers three choices for row $j$ (Algorithm~\ref{alg:leaf_init} in Appendix~\ref{sec:pseudocode}):
(a) The row is unused.
(b) The row is assigned a nonempty ordinary row type $T$, and the variables indexed by $T$ take values in $\mv$.
(c) The row is assigned a hard row type $T^r\in\ml$.
Its hard variables, indexed by $U_r=U_H(T^r)$, take the values $p_r$, and its free variables, indexed by $T^r\setminus U_r$, take values in $\mv$.
Since their values $p_r$ may differ, two copies of the same row type in $\ml$ with different labels are distinct choices.

\paragraph{Merging and acceptance.}
At each internal node, $\SelectBranch$ combines every pair of states of the two children whose sets $\locc$ are disjoint and whose summed capacity caps do not exceed $s$.
It adds the Gram matrices and the objective values, and it takes the union of the sets $\locc$.
It also adds the counters $\mu_e$ up to $d$, rounds the summed caps with $R_c$, and updates the dictionary.
At the root, a state is \emph{accepted} if $\locc=[b]$, $\mu_e=d$ for every $e\in S$, $v\ge 0$, and $\fnorm{G-I_d}\le\delta$.
To reconstruct the candidate solution of an accepted state, every dictionary entry stores its leaf assignment or back-pointers to the two child states from which it was constructed.

\subsection{The linear oracle}
\label{sec:linear_oracle}
Algorithm~\ref{alg:linear_oracle} enumerates all branches.
For every accepted root state of $\SelectBranch$, it reconstructs the candidate solution $Y$, applies $\ExactFinalize$ (Algorithm~\ref{alg:exact_finalize} in Appendix~\ref{sec:pseudocode}), and keeps the best matrix found.

\begin{algorithm}[H]
	\caption{$\LinearOracle$}\label{alg:linear_oracle}
	
	\Input{
    Rational $A \in \mathbb{Q}^{n \times d}$, 
    sparsity constants $s$, $t\in \Q \cap (0, 1)$.
    }
    
	\Output{
    $X \in \mathcal{F}_s$ such that $\tr(A^\top X) \ge (1-t) \optlin(A;s)$.
    }
	
	$X_{\rm best} \gets \left(\begin{smallmatrix}I_d\\O\end{smallmatrix}\right)$\;
	
	\If{$A = O$}{\Return $X_{\rm best}$\;}
	
	\If{$d = 1$}{
		$x_{\rm best}\gets$ the restriction of $A$ to $s_1$ entries of largest absolute value, normalized\;
		\textbf{return} $x_{\rm best}$\;
	}

	\ForEach{$H \subseteq \cp_d$}{
		$S \gets \cp_d \setminus H$\;
		\For{$b = 0, 1, \dots, \min\{n, (d-1)|H|\}$ }
		{
			\ForEach{multiset $\ml=\multiset{T^1, T^2, \ldots, T^b}$ of subsets of $[d]$}
			{
				\If{every $T^i$ contains at least one $e\in H$, and every $e\in H$ is contained in at most $d-1$ of the sets $T^1,\ldots,T^b$}
				{
					$D \gets \sum_{i = 1}^b \abs{\cup_{e\in H, e\subseteq T^i} e}$\;
					\ForEach{$p \in \FixHardVariables(H, \ml, D, t)$}{
						$\mathcal{R} \gets \SelectBranch(A, s, H, \ml, p, t)$\;
						\ForEach{accepted root state $q \in \mathcal{R}$}{
							Reconstruct its candidate solution $Y$ and assigned row types $\rt_j$\;
							$X \gets \ExactFinalize(A, Y, (\rt_j), S)$\;
							\If{$\tr(A^\top X) > \tr(A^\top X_{\rm best})$}{
								$X_{\rm best} \leftarrow X$\;
							}
						}
					}
				}
			}
		}
	}
	\Return $X_{\mathrm{best}}$\;
\end{algorithm}

The subroutine $\ExactFinalize$ first applies Algorithm~\ref{alg:two_column} to $Y$.
Since $\locc=[b]$ at the root and the sets $\locc$ of the two children are disjoint at every merge, each hard row type in $\ml$ is assigned to exactly one row of $Y$.
Moreover, ordinary row types contain no hard pair.
Hence \eqref{eqn:hard_constraints} gives $Y_i^\top Y_l = 0$ for every $\{i, l\}\in H$.
Since $\mu_e=d$ for every $e\in S$ at the root, Proposition~\ref{prop:two_column_correctness} applies, and the resulting matrix $\widetilde Y$ has orthogonal columns $\widetilde Y_i$.
$\ExactFinalize$ then divides each $\widetilde Y_i$ by its norm, and flips the sign of every $\widetilde Y_i$ with $A_i^\top\widetilde Y_i<0$.
The result is in $\mf_s$, and the sign flips do not decrease the objective.

%% file: sections/guarantees.tex
\section{Approximation guarantee and running time}
\label{sec:guarantee}
We now state the approximation guarantee of Algorithm~\ref{alg:main_alg} and sketch its proof.
The full proof is in Appendix~\ref{app:approximation_guarantee}.
Then we bound the running time.

\begin{restatable}{theorem}{approximationTheorem}
	\label{thm:main_approximation}
	Let $R\in\Q^{n\times r}$, $s\in[n]^d$, and $\epsilon\in\Q\cap(0,1)$.
	Then Algorithm~\ref{alg:main_alg} returns $X\in\mf_s$ satisfying $\fnorm{R^\top X}^2 \ge (1-\epsilon)\opt(R;s)$.
\end{restatable}

\paragraph{Proof idea of Theorem~\ref{thm:main_approximation}.}
The grid of matrices $W$ in Algorithm~\ref{alg:main_alg} contains a matrix within Frobenius distance $\epsilon/4$ of $R^\top X^*/\fnorm{R^\top X^*}$, where $X^*$ is an optimal solution of \ref{eqn:SPCA}.
Hence it suffices to show that $\LinearOracle$ returns $X\in\mf_s$ with $\tr(A^\top X)\ge(1-t)\optlin(A;s)$.
By Proposition~\ref{prop:two_column_correctness} and the acceptance tests at the root, every output of $\LinearOracle$ is in $\mf_s$.
For the objective value, let $X^*$ now be an optimal solution of~\eqref{eqn:sparse_linear}, and consider the branch whose hard pairs $H$ and list $\ml$ are those of $X^*$.
We construct a chain of matrices
\begin{equation*}
	X^* \longrightarrow Z \longrightarrow Z_0 \longrightarrow Y \longrightarrow \widetilde Y \longrightarrow X,
\end{equation*}
in which each step loses only a small part of the objective value or of the orthogonality.

\textit{Slack.}
The matrix $Z$ is obtained from $X^*$ by deleting rows, that is, by setting them to zero.
We keep the at most $P_d$ rows that contain hard pairs or that we select to cover each simple pair $d$ times.
For every column $i$ with $s_i>K_0$, we then delete rows of small weight until at most $(1-4\tau)s_i$ rows of $Z$ have a row type that contains $i$.
The weight of a row is the sum of its share of the objective value and its share of $\fnorm{X^*}^2$.
Since $s_i>K_0$ is large compared with $P_d/\tau$, the deleted rows have total weight $\mo(d\tau)$.
Thus $Z$ loses at most a fraction $\mo(d\tau)$ of the objective value, and $\fnorm{Z^\top Z-I_d}=\mo(d^2\tau)$.
The slack $4\tau s_i$ absorbs the rounding of the capacity caps.
Indeed, each merge multiplies a cap by at most $1+\alpha$, and over the $H_{\rm tree}$ levels this gives a factor at most $e^{\tau/8}$, which is smaller than $1/(1-4\tau)$.
For $s_i\le K_0$, the caps are never rounded.

\textit{Discretization.}
The matrix $Z_0$ is obtained from $Z$ by two changes.
The hard variables of $Z$ are replaced by the exact solution $p$ that $\FixHardVariables$ returns for the box containing them, and $p$ is within distance $\sqrt{D}a$ of these values.
Each free variable is rounded to $\mv$, with relative error at most $1/M$, or with absolute error at most $2^{-J_0}$ for very small entries.
The difference $\tr(A^\top(Z_0-Z))$ only involves entries of $A$ in the support of $X^*$.
By Lemma~\ref{lem:fnorm_A_in_supp}, these entries have Frobenius norm at most $d\optlin(A;s)$.
We obtain $\tr(A^\top Z_0)\ge(1-\delta/4)\optlin(A;s)$ and $\fnorm{Z_0^\top Z_0-I_d}\le\delta/4$ (Proposition~\ref{prop:finite_witness}).
Moreover, the columns of $Z_0$ corresponding to hard pairs are exactly orthogonal, and every simple pair is contained in the row types of at least $d$ rows of $Z_0$.

\textit{Approximate dynamic programming.}
The rows of $Z_0$ give a sequence of states from the leaves to the root.
At every node, $\KeepBest$ may replace a state by another state with the same key and a larger objective value.
The two Gram matrices then lie in the same bin, so they differ by at most $d\theta\tr(G_u)+2\kappa_G$ in Frobenius norm.
The trace-relative bins keep these errors proportional to the traces, which sum to less than $2d$ over the at most $N_{\rm leaf}$ states of a level.
Summing over the $H_{\rm tree}+1=\mo(\log n)$ levels, the total change of the Gram matrix is less than $\delta/8$.
Hence the root has an accepted candidate $Y$ with $\tr(A^\top Y)\ge\tr(A^\top Z_0)$ and $\fnorm{Y^\top Y-I_d}<\delta$ (Proposition~\ref{prop:shadow_state_survival}).
Since the occupied labels of the two children are disjoint at every merge, each hard row type is used exactly once in $Y$, so the columns of $Y$ corresponding to hard pairs are exactly orthogonal.

\textit{Exact orthogonality.}
Algorithm~\ref{alg:two_column} turns $Y$ into a matrix $\widetilde Y$ with orthogonal columns and $\tr(A^\top\widetilde Y)\ge\tr(A^\top Y)$.
By Proposition~\ref{prop:two_column_correctness}, the squared norm of column $i$ of $\widetilde Y$ is $(Y^\top Y)_{ii}+\sum_{l\ne i}\abs{(Y^\top Y)_{il}}$, which is at most $1+(1+\sqrt{d-1})\delta$.
Normalizing the columns and flipping their signs as in $\ExactFinalize$ gives $X\in\mf_s$ and keeps at least a factor $1-d\delta$ of the objective.
Since $\delta=t/(8(d+1))$, we obtain $\tr(A^\top X)\ge(1-t)\optlin(A;s)$ (Proposition~\ref{prop:linear_oracle_guarantee}). \qed

The next theorem bounds the bit complexity of Algorithm~\ref{alg:main_alg}.
Here, real algebraic numbers are represented exactly, as described in Appendix~\ref{app:complexity}.
\begin{restatable}{theorem}{mainAlgComplexityTheorem}
	\label{thm:main_alg_complexity}
	Let $R\in\Q^{n\times r}$, $s\in[n]^d$, and $\epsilon\in\Q\cap(0,1)$.
	Let $B_{\rm SPCA}$ be the encoding size of $(R, s, \epsilon)$, and let $U_{\epsilon}:= 1 + \log_2(n+1) + \log_2(r + d + 1) + \log_2(1/\epsilon)$.
	Then there is an implementation of Algorithm~\ref{alg:main_alg} whose bit complexity $T_{\rm SPCA}$ satisfies 
	\begin{equation*}
		T_{\rm SPCA} \le F_{\rm SPCA}(r, d) n \epsilon^{-(rd + d^3 - d^2 + 4d + 2 + c)} (B_{\rm SPCA} + 1)^c U_{\epsilon}^{d^2 + 5d + 4 +4c},
	\end{equation*}
	where $F_{\rm SPCA}(r,d)$ depends only on $r$ and $d$, and $c$ is an absolute constant.
\end{restatable}
Since $\log_2(n+1)$ and $\log_2(1/\epsilon)$ are at most $B_{\rm SPCA}$, this bound is polynomial in $B_{\rm SPCA}$ and in $1/\epsilon$ for fixed $r$ and $d$.
Together with Theorem~\ref{thm:main_approximation}, Theorem~\ref{thm:main_alg_complexity} proves Theorem~\ref{thm:FPTAS}.
The proof of Theorem~\ref{thm:main_alg_complexity} is in Appendix~\ref{app:complexity}.

For fixed $r$, $d$, and $\epsilon$, the number of steps of Algorithm~\ref{alg:main_alg} is linear in $n$ up to a polylogarithmic factor.
This dependence on $n$ comes from the ADP tree, the trace-relative Gram matrix bins, and the slack in the capacity caps, whereas a straightforward DP with exact capacity counts and a uniform Gram matrix grid can have $\mo(n^{d+d(d+1)/2})$ states per row.
The additional factors in the bit complexity bound account for the accuracy-dependent enumeration and discretization, as well as the cost of exact algebraic sampling, field operations, and comparisons.
In particular, the degrees of the algebraic field extensions are bounded in terms of $d$, and the bit complexity of these operations is polynomial in the input encoding size.
Further details are given in Appendix~\ref{app:complexity}.

%% file: sections/acknowledgement.tex
\iftoggle{is_LNCS}{%
\begin{credits}
\subsubsection*{\ackname}
AI tools (Claude, developed by Anthropic, and ChatGPT, developed by OpenAI) were used for proofreading and suggesting improvements to the exposition.
We take full responsibility for the content of the paper.
We agree that reviewers may upload this submission to generative AI services.
\end{credits}
}{%
\bigskip
\begin{small}\sloppy
\paragraph{Funding.}
A. Del Pia is partially supported by AFOSR Grant \mbox{FA9550-23-1-0433} and ONR Grant \mbox{N00014-25-1-2490}.
Any opinions, findings, conclusions, or recommendations expressed in this material are those of the authors and do not necessarily reflect the views of the Air Force Office of Scientific Research or the Office of Naval Research.

\paragraph{AI methodology.}
AI tools (Claude, developed by Anthropic, and ChatGPT, developed by OpenAI) were used for proofreading and suggesting improvements to the exposition.
The authors take full responsibility for the content of the paper.
\end{small}
}

%% file: sections/relatedwork.tex
\section{Additional related work}
\label{sec:related_work}
In this section, we discuss additional related work.

Algorithms for single-component SPCA have been studied extensively.
Existing approaches include greedy and iterative algorithms \cite{YuaZha13,HagPhaZhu16,LiXie25}, regression-based formulations \cite{ZouHasTib06}, and mixed-integer and convex optimization \cite{BerCorPau22,DeyMazWan22,LiXie25}. 
Chan, Papailiopoulos, and Rubinstein~\cite{ChaPapRub16} give a polynomial-time algorithm with approximation factor $\max\{1/\sqrt{k},n^{-1/3}\}$, where $k$ is the sparsity constant and $n$ is the dimension. 
Other approximation algorithms exploit low-rank structure \cite{PapDimKor13}, semidefinite relaxations \cite{ChoDriWooZho20,dPZho25xx}, and block-diagonal approximations~\cite{dPZhoZhu25ICLR}.

Several methods compute multiple sparse components of a covariance matrix $C$.
A common approach is \emph{deflation}~\cite{Mac09}.
It computes a single sparse component $x$ of $C$, replaces $C$ by a deflated matrix, such as $(I - x x^\top) C (I - x x^\top)$, and iterates.
Deflation can be combined with any algorithm for single-component SPCA.
However, the resulting components need not be orthogonal, and so they may not be feasible for \ref{eqn:SPCA}.
In contrast, Bertsimas and Kitane~\cite{BerKit23} compute all components at once, with orthogonality enforced, when the components share the same support or have disjoint supports.
Their method approximates the problem by binary linear optimization, and they report better solutions than deflation.
Cheng and Hu~\cite{CheHu26} enforce sparsity and orthogonality together through Gram--Schmidt orthogonalization.
They combine this approach with branch-and-bound to compute near-optimal solutions, and they speed it up by a decomposition based on block-diagonal approximations of the covariance matrix.
For row-sparse PCA, where all components share a common support, Dey, Molinaro, and Wang~\cite{DeyMolWan22c} propose convex relaxations that provably approximate the feasible region, and convex integer programs that give dual bounds.

Another line of research studies statistical guarantees for SPCA, including estimation and support recovery under spiked covariance models \cite{AmiWai08,DesMon16}. 
In particular, Vu, Cho, Lei, and Rohe~\cite{VuChoLeiRoh13} propose a Fantope relaxation with statistical guarantees for sparse principal subspace estimation.

%% file: sections/algorithm.tex
\section{Pseudocode}
\label{sec:pseudocode}
This appendix gives the pseudocode of our algorithm for \ref{eqn:SPCA} and of the subroutines of $\LinearOracle$ (Algorithm~\ref{alg:linear_oracle}), which are described in Section~\ref{sec:motivation}.
Algorithm~\ref{alg:main_alg} is our algorithm for \ref{eqn:SPCA}.
Algorithm~\ref{alg:fix_hard_variables} computes the values of the hard variables (Section~\ref{sec:fix_hard_variables}).
Algorithm~\ref{alg:keep_best} updates a dictionary with the data of a new partial solution, and Algorithm~\ref{alg:leaf_init} initializes the dictionary of a leaf of the ADP tree (Section~\ref{sec:motivation_ADP}).
Algorithm~\ref{alg:select_branch} runs the ADP for one branch of $\LinearOracle$.
It omits the back-pointers used to reconstruct the candidate solutions of accepted states.
For a state $q$, we write $\locc(q)$, $\mu_e(q)$, $\hat c(q)$, $G(q)$, and $v(q)$ for its components.
Algorithm~\ref{alg:exact_finalize} turns an accepted candidate solution into a point of $\mf_s$ (Section~\ref{sec:linear_oracle}).

\begin{algorithm}[H]
	
	\caption{$(1-\epsilon)$-algorithm for \ref{eqn:SPCA}}
	\label{alg:main_alg}
	\Input{
    $R \in \mathbb{Q}^{n \times r}$, 
    sparsity constants $s\in [n]^d$, 
    $\epsilon \in \Q \cap (0, 1)$.
    }
	\Output{
    $X \in \mathcal{F}_s$ such that $\fnorm{R^\top X}^2 \ge (1-\epsilon) \opt(R;s)$.
    }
	
	$X_{\mathrm{best}} \gets \left(\begin{smallmatrix}I_d\\O\end{smallmatrix}\right)$\;
	\If{$R = O$}{
		\Return $X_{\rm best}$\;
	}
	$J_W \gets \lceil 4rd/\epsilon \rceil$; $t \gets \epsilon/4$\;
	\ForEach{$W \in J_W^{-1} \Z^{r \times d}$ with $\fnorm{W} \le 1$}{
		$A \gets RW$\;
		$X \gets \LinearOracle(A, s, t)$\;
		\If{$\fnorm{R^\top X}^2 > \fnorm{R^\top X_{\rm best}}^2$}{
			$X_{\rm best} \gets X$\;
		}
	}
	\Return $X_{\rm best}$\;
\end{algorithm}

\begin{algorithm}[H]
 	\caption{$\FixHardVariables$}
 	\label{alg:fix_hard_variables}
 	
 	\Input{
 		Set of hard pairs $H$, 
        list of hard row types $\mathcal L=\multiset{T^1, T^2, \ldots, T^b}$,
 		number of hard variables $D$, 
        parameter $t\in \Q \cap (0, 1)$.
 	}
 	\Output{
 		A finite list $\mathcal P$ of value assignments for the hard variables.
 	}
 	$\mathcal P\gets\emptyset$\;
 	\If{$D=0$}{
 		\Return $\{\emptyset\}$\;
 	}
 	Set $a$ and $J_p$ as in Table~\ref{tab:parameters}\;
 	Cover $[-1,1]$ by $J_p$ closed intervals $\Omega_1,\ldots,\Omega_{J_p}$ of length at most $a$, as in Section~\ref{sec:fix_hard_variables}\;
 	\ForEach{box
 		$Q=\Omega_{\ell_1}\times\cdots\times \Omega_{\ell_D} \subseteq[-1,1]^D$}{
 		Decide whether the quadratic system given by $p\in Q$ and \eqref{eqn:hard_constraints} is feasible\;
 		\If{the system is feasible}{
 			Find one solution $p^Q$, add $p^Q$ to $\mathcal{P}$\;
 		}
 	}
 	
 	\Return $\mathcal P$\;
 \end{algorithm}

\begin{algorithm}[H]
	\caption{$\KeepBest$}
	\label{alg:keep_best}
	
	\Input{
		A dictionary $\md$, 
        the data $\locc,(\mu_e),\hat c, G, v$ of a candidate, 
        sparsity constants $s$,
        and parameter $t\in \Q \cap (0, 1)$
	}
	\Output{
		The updated dictionary $\md$.
	}
	\If{$\hat c_i>s_i$ for some $i\in[d]$ \textbf{or} $\tr(G) \ge 4d$}{
		\Return $\md$\;
	}
	$\hat G\gets \mathrm{lab}(G)$ as described in Section~\ref{sec:motivation_ADP}, $\sk \gets (\locc,(\mu_e),\hat c, \hat G)$\;
	\If{$\sk\notin\md$}{
		$\md[\sk]\gets(v,G)$\;
	}
	\ElseIf{$v> \text{the objective value recorded in } \md[\sk]$}{
		$\md[\sk]\gets(v,G)$\;
	}
	
	\Return $\md$\;
\end{algorithm}

\begin{algorithm}[H]
	\caption{$\LeafInit$}
	\label{alg:leaf_init}
	
	\Input{
		Leaf index $j$, 
        objective matrix $A$ with rows $a_j$, 
        sparsity constant vector $s$,
		set of hard pairs $H$, set of simple pairs $S$,
		list of hard row types $\ml=\multiset{T^1,\ldots,T^b}$,
		fixed hard-variable values $p$,
		and parameter $t\in \Q \cap (0, 1)$.
	}
	\Output{
		The initialized dictionary $\md_j$ for leaf $j$.
	}
	
	$\md_j\gets\emptyset$\;
	Form the data $q_{\rm unused} \gets (\emptyset,0,0,O,0)$ of an unused row\;
	$\md_j \gets \KeepBest(\md_j, q_{\rm unused}, s, t)$\;
	\If{$j\le n$}{
		\ForEach{nonempty $T\subseteq[d]$ with $U_H(T)=\emptyset$}{
			\ForEach{$z\in\mathcal V^{|T|}$}{
				$y_T \gets z$, $y_{[d]\setminus T} \gets 0$, $\locc\gets\emptyset$, $\hat c\gets \dso_{T}$, $\mu_e \gets \dso_{\{e\subseteq T\}}$ for every $e\in S$\; 
				$G\gets y y^\top$, $v\gets a_j y$\;
				Form the candidate data $q \gets (\locc,\mu,\hat c, G, v)$\;
				$\md_j \gets \KeepBest(\md_j, q, s, t)$\;
			}
		}
		
		\For{$r=1,\ldots,b$}{
			$U_r\gets U_H(T^r)$, $F_r\gets T^r\setminus U_r$\;
			\ForEach{$z\in\mathcal V^{|F_r|}$}{
				$y_{U_r} \gets p_r$, $y_{F_r} \gets z$, $y_{[d]\setminus T^r} \gets 0$, $\locc \gets \{r\}$, $\hat c \gets \dso_{T^r}$, $\mu_e \gets \dso_{\{e\subseteq T^r\}}$ for every $e\in S$\;
				$G\gets y y^\top$, $v\gets a_j y$\;
				Form the candidate data $q \gets (\locc,\mu,\hat c, G, v)$\;
				$\md_j \gets \KeepBest(\md_j, q, s, t)$\;
			}
		}
	}
	
	\Return $\md_j$\;
\end{algorithm}

\begin{algorithm}[H]
	\caption{$\SelectBranch$}
	\label{alg:select_branch}
	
	\Input{
		Objective matrix $A$, sparsity constant vector $s$,
		set of hard pairs $H$,
		list of hard row types
		$\ml=\multiset{T^1,\ldots,T^b}$,
		fixed hard-variable values $p$,
		and parameter $t\in \Q \cap (0, 1)$.
	}
	\Output{
		A set $\mathcal R$ of accepted root states.
	}
	
	Set $\delta$ as in Table~\ref{tab:parameters}, $S\gets \cp_d\setminus H$\;
	 
	$\md_j\gets \LeafInit(j, A, s, H, S, \ml, p, t)$ for every $j\in [N_{\rm leaf}]$\;

	\For{each internal node $u$, from bottom to top}{
		Let $u_L$ and $u_R$ be the two children of $u$\;
		$\md_u\gets\emptyset$\;
		\ForEach{$q_L\in\md_{u_L}$ and $q_R\in\md_{u_R}$}{
			\If{$\locc(q_L)\cap\locc(q_R) = \emptyset$}{
					$c_i^{\rm sum} \gets \hat c_i(q_L)+\hat c_i(q_R)$ for every $i\in [d]$\;
					\If{$c_i^{\rm sum} \le s_i$,  $\forall i\in[d]$}{
						$\hat c_i\gets R_c(c_i^{\rm sum})$ according to \eqref{eqn:capacity_cap} for every $i\in [d]$\;		
                        
						$\locc \gets \locc(q_L)\cup\locc(q_R)$, 
						$\mu_e \gets \min\{d,\mu_e(q_L)+\mu_e(q_R)\}$ for every $e\in S$\;
                        
						$v\gets v(q_L)+v(q_R)$, $G\gets G(q_L)+G(q_R)$\;
						
						Form the data $q \gets (\locc,\mu,\hat c,G,v)$ of the parent candidate\;
                        
						$\md_u \gets \KeepBest(\md_u, q, s, t)$\;
					}
			}
		}
	}
	
	$\mathcal R\gets\emptyset$, $\md_{\rm root} \gets$ the dictionary at the root\;
	
	\ForEach{state $q$ in $\md_{\rm root}$}{
		\If{$\locc(q) = [b]$ \textbf{and} $\mu_e(q) = d$, $\forall e\in S$ \textbf{and} $v(q) \ge 0$ \textbf{and} $\fnorm{G(q)-I_d}^2 \le \delta^2$ }{
			$\mathcal R\gets\mathcal R\cup\{q\}$\;
		}		
	}
	
	\Return $\mathcal R$\;
\end{algorithm}

\begin{algorithm}[H]
	\caption{$\ExactFinalize$}
	\label{alg:exact_finalize}
	
	\Input{
		Objective matrix $A$ with columns $A_i$, candidate solution $Y$ of an accepted root state of $\SelectBranch$, 
        assigned row types $(\rt_j)_{j=1}^n$, 
        and set of simple pairs $S$.
	}
	\Output{
		A matrix in $\mathcal F_s$.
	}
	
	$G^0\gets Y^\top Y$,
	$\widetilde Y \gets$ Output of Algorithm~\ref{alg:two_column} with input $(S, Y, A,(\rt_j)_{j=1}^n)$\;
	
	Divide each $\widetilde Y_i$ by $\norm{\widetilde Y_i}= \sqrt{\onorm{G_i^0}}$, where $G_i^0$ is the $i$-th column of $G^0$\;
	
	For each $i\in [d]$, flip the sign of $\widetilde Y_i$ if $A_i^\top \widetilde Y_i < 0$\;
	
	\Return{$\widetilde Y$}\;
\end{algorithm}

%% file: sections/appendix.tex
\section{Deferred proofs}
\label{sec:appendix}

\subsection{Hardness results}
In this section, we prove Theorems~\ref{thm:NPhard} and~\ref{thm:NP-hard_linear}.
\npHardnessTheorem*

\begin{proof}[Proof of Theorem~\ref{thm:NPhard}]
	We reduce $\Partition$ to the decision version of \ref{eqn:SPCA}.
	An instance of $\Partition$ is a multiset of positive integers $a_1, a_2, \ldots, a_m$, and the question is whether it can be split into two subsets with equal sums.
	
	\textit{Construction.}
	Let $M:=\sum_{i = 1}^m a_i$, and define the vector $b \in \Z^{2m+1}$ as follows.
	The first $m$ entries of $b$ are $2a_1, 2 a_2, \ldots, 2a_m$, the $(m+1)$-th and $(m+2)$-th entries of $b$ are both $-M$, and the remaining $m-1$ entries are zero.
	Let $1_{2m+1}$ be the all-one vector in $\R^{2m+1}$, and note that $1_{2m+1}^\top b = 2M - M - M = 0$.
	Let $B:= 8m \norm{b}^2$, $n:=2m+2$, and
	\begin{align*}
		R := \begin{pmatrix}
			0 & B \\
			B b & 1_{2m+1}
		\end{pmatrix}
		\in \Z^{n\times 2},
		\qquad
		R R^\top = 
		\begin{pmatrix}
			B^2 & B 1_{2m+1}^\top\\
			B 1_{2m+1} & B^2 b b^\top + 1_{2m+1} 1_{2m+1}^\top
		\end{pmatrix}.
	\end{align*}
	Finally, let $k := m+2$ and $\tau_0:= B^2 (\norm{b}^2 + 1) + m + 1$.
	The instance $(R,k,\tau_0)$ can be computed in time polynomial in the encoding size of $a_1,\ldots,a_m$.
	
	Let $r_1$ and $r_2$ be the two columns of $R$.
	Since $1_{2m+1}^\top b = 0$, the columns $r_1$ and $r_2$ are orthogonal.
	Moreover, $\norm{r_1}^2=B^2\norm{b}^2$ and $\norm{r_2}^2=B^2 + 2m + 1$, and therefore
	\begin{align*}
		\tr(RR^\top) = \norm{r_1}^2 + \norm{r_2}^2 = B^2 \norm{b}^2 + B^2 + 2m + 1
		= \tau_0 + m.
	\end{align*}
	We show that $\opt(R; (k, k)) \ge \tau_0$ if and only if the $\Partition$ instance is a yes-instance.
	
	\textit{From a partition to a solution.}
	Suppose that $J\subseteq [m]$ satisfies $\sum_{j\in J} a_j = M / 2$.
	Since the $a_j$ are positive, $1\le |J|\le m-1$.
	Since $b$ has $m-1$ zero entries, there is a set $F$ of $m - |J|$ indices of zero entries of $b$.
	Let $I := J \cup \{m + 1\} \cup F$.
	Then $|I| = m+1$ and $\sum_{i\in I} b_i = 2(M/2) - M = 0$.
	Let
	\begin{align*}
		x := \frac{1}{\sqrt{B^2 + m + 1}} \begin{pmatrix}
			B\\
			\dso_{I}
		\end{pmatrix},
		\qquad
		y:= \frac{1}{\norm{b}} \begin{pmatrix}
			0\\
			b
		\end{pmatrix},
	\end{align*}
	and $X:= (x, y)$.
	Both vectors have unit norm, and $\znorm{x} = \znorm{y} = m+2 = k$.
	Moreover, $x^\top y = \dso_I^\top b/(\norm{b}\sqrt{B^2+m+1}) = 0$.
	Hence $X\in\mf_{(k,k)}$.
	Moreover, $r_1^\top x = 0$, $r_2^\top x = \sqrt{B^2+m+1}$, $r_1^\top y = B\norm{b}$, and $r_2^\top y = 0$.
	Therefore,
	\begin{align*}
		\fnorm{R^\top X}^2 
		= (B^2 + m + 1) + B^2 \norm{b}^2 = \tau_0.
	\end{align*}
	
	\textit{From a solution to a partition.}
	Suppose that $\opt(R; (k, k)) \ge \tau_0$.
	Let $X^*= (x^*, y^*)$ be an optimal solution, and let $P:= x^* (x^*)^\top + y^* (y^*)^\top$ be the orthogonal projector onto $\sp1(\{x^*, y^*\})$.
	Let $H:= r_1 r_1^\top /\norm{r_1}^2 + r_2 r_2^\top /\norm{r_2}^2$ be the orthogonal projector onto $\sp1(\{r_1, r_2\})$.
	
	(a) We first show that the entries of $P$ and $H$ are close to each other.
	Since $\tr(P RR^\top) = \opt(R; (k, k)) \ge \tau_0$ and $\tr(RR^\top) = \tau_0 + m$, we have $\tr((I - P) RR^\top) \le m$.
	Since $I-P$ is positive semidefinite and $\norm{r_1}^2, \norm{r_2}^2 \ge B^2$, we obtain
	\begin{align*}
		\tr((I - P) H) = \frac{r_1^\top (I-P) r_1}{\norm{r_1}^2} + \frac{r_2^\top (I-P) r_2}{\norm{r_2}^2}
		\le \frac{1}{B^2} \tr((I - P)RR^\top) \le \frac{m}{B^2}.
	\end{align*}
	Since $\fnorm{P}^2 = \fnorm{H}^2 = \tr(H) = 2$, we have $\fnorm{P - H}^2 = 4 - 2\tr(PH) = 2\tr((I-P) H) \le 2m/B^2$.
	Since $B = 8m \norm{b}^2$ and $\sqrt{2m}<2m$, every $i,j\in[n]$ satisfy
	\begin{align}
		\label{eqn:P_close_H}
		\abs{P_{ij} - H_{ij}} \le \fnorm{P - H} \le \frac{\sqrt{2m}}{B} < \frac{1}{4\norm{b}^2}.
	\end{align}
	
	(b) Next, we determine the structure of the supports of $x^*$ and $y^*$.
	Let $2\le i, j\le m+3$.
	Then $b_{i-1}$ and $b_{j-1}$ are non-zero integers, and $B^2 > 8\norm{b}^2$.
	Hence
	\begin{align*}
		|H_{ij}|  = \abs{\frac{b_{i-1} b_{j-1}}{\norm{b}^2} + \frac{1}{B^2 + 2m + 1}}
		\ge \frac{1}{\norm{b}^2}  - \frac{1}{B^2} 
		> \frac{7}{8\norm{b}^2},
	\end{align*}
	and \eqref{eqn:P_close_H} gives $P_{ij}\ne 0$.
	Moreover, since $B^2 > 2m + 1$, we have $H_{11} = B^2 / (B^2 + 2m + 1) > 1/2$.
	Since $\norm{b}^2\ge 2$, \eqref{eqn:P_close_H} gives $P_{11} = (x_1^*)^2 + (y_1^*)^2 > 1/4$.
	In particular, $x^*_1\ne0$ or $y^*_1\ne0$.
	Both sparsity constants equal $k$, so after swapping the two columns of $X^*$ if necessary, we may assume that $x^*_1\ne 0$.
	After flipping the sign of $x^*$ if necessary, we may also assume that $x^*_1> 0$.
	
	We show that $y^*_1 = 0$. 
	Suppose not.
	Since $x^*_1\ne0$, $n = 2m+2$, and $\znorm{x^*}\le k = m + 2$, there exists an index $i \in \{2,\ldots,m+3\}$ such that $x^*_i = 0$.
	Since $y^*_1\ne 0$ and $\znorm{y^*} \le k$, there exists an index $j \in \{2,\ldots,m+3\}$ such that $y^*_j = 0$.
	Then $P_{ij} = x_i^* x_j^* + y_i^* y_j^* = 0$, a contradiction.
	Therefore, $y^*_1 = 0$, and so $P_{11} = (x_1^*)^2$.
	We write
	\begin{align*}
		x^* = \begin{pmatrix}
			\sqrt{P_{11}} \\
			z
		\end{pmatrix},
		\qquad
		y^* = \begin{pmatrix}
			0\\
			v
		\end{pmatrix},
	\end{align*}
	where $z,v\in\R^{2m+1}$ satisfy $|\supp(z)| \le m+1$, $z^\top v = 0$, $\norm{z}^2 = 1-P_{11}$, and $\norm{v}=1$.
	
	(c) Finally, we extract a partition from $\supp(z)$.
	We first show that $z \ne 0$.
	Suppose not.
	Then $x^*=\mathbf e_1$, thus $(x^*)^\top RR^\top x^* = B^2$.
	Since $r_1$ and $r_2$ are orthogonal, the two non-zero eigenvalues of $RR^\top$ are $\norm{r_1}^2$ and $\norm{r_2}^2$.
	Since $\norm{b}^2\ge 2$ and $B^2>2m+1$, the largest eigenvalue of $RR^\top$ is $\norm{r_1}^2 = B^2\norm{b}^2$.
	Hence
	\begin{align*}
		\tr(P RR^\top) \le B^2 + B^2 \norm{b}^2 < \tau_0,
	\end{align*}
	a contradiction. 
	
	Let $\alpha$ be the inner product of $z / \norm{z}$ and $b / \norm{b}$.
	Since $z/\norm{z}$ and $v$ are orthonormal, the squared norm of the projection of $b/\norm{b}$ onto $\sp1(\{z, v\})$ is $\alpha^2 + (b^\top v)^2/\norm{b}^2\le 1$.
	Hence $(b^\top v)^2 \le \norm{b}^2 (1 - \alpha^2)$.
	Similarly, since $b/\norm{b}$ and $1_{2m+1}/\sqrt{2m+1}$ are orthonormal, we have $(v^\top 1_{2m+1})^2 \le (2m+1)(1 - (v^\top b)^2/\norm{b}^2)$.
	Using $B^2\norm{b}^2 > 2m+1$, we obtain
	\begin{align*}
		(y^*)^\top RR^\top y^*
		&= B^2 (v^\top b)^2 + (v^\top 1_{2m+1})^2\\
		&\le 2m + 1 + (B^2 \norm{b}^2 - 2m - 1) \frac{(v^\top b)^2}{\norm{b}^2}\\
		& \le 2m + 1 + (B^2 \norm{b}^2 - 2m - 1) (1 - \alpha^2)\\
		& = B^2 \norm{b}^2 - (B^2 \norm{b}^2 - 2m - 1) \alpha^2.
	\end{align*}
	Moreover, since $(b^\top z)^2 = \alpha^2\norm{b}^2\norm{z}^2 = \alpha^2\norm{b}^2(1-P_{11})$, we have
	\begin{align*}
		(x^*)^\top RR^\top x^* &= (r_1^\top x^*)^2 + (r_2^\top x^*)^2 \\
		&= B^2 \norm{b}^2 (1 - P_{11}) \alpha^2 + (B\sqrt{P_{11}} + 1_{2m+1}^\top z)^2.
	\end{align*}
	Adding the two bounds, we obtain
	\begin{align}
		\label{eqn:hardness_upper}
		\opt(R; (k, k)) \le B^2 \norm{b}^2 + (B \sqrt{P_{11}} + 1_{2m+1}^\top z)^2 - (B^2 \norm{b}^2 P_{11} - 2m-1) \alpha^2.
	\end{align}
	Since $B = 8m \norm{b}^2$ and $P_{11} > 1/4$, we have $B^2 \norm{b}^2 P_{11} - 2m-1 > 0$. 
	Let $I':=\supp(z)$.
	By the Cauchy--Schwarz inequality and since $\abs{I'}\le m + 1$, we have
	\begin{align}
		\label{eqn:hardness_CS}
		(B \sqrt{P_{11}} + 1_{2m+1}^\top z)^2 \le (B^2 + \abs{I'})\norm{x^*}^2 \le B^2 + m + 1.
	\end{align}
	Together with \eqref{eqn:hardness_upper} and $\opt(R;(k,k))\ge\tau_0$, this gives
	\begin{align*}
		\tau_0 \le \opt(R; (k, k)) \le \tau_0 - (B^2 \norm{b}^2 P_{11} - 2m-1) \alpha^2.
	\end{align*}
	Therefore $\alpha = 0$, that is, $z^\top b = 0$, and both inequalities in \eqref{eqn:hardness_CS} hold with equality.
	Equality in the second inequality gives $\abs{I'} = m+1$.
	Equality in the Cauchy--Schwarz inequality shows that $(\sqrt{P_{11}}, z_{I'})$ is a multiple of $(B,1,\ldots,1)$, so $z_{I'}$ is a non-zero multiple of the all-one vector.
	Since $z^\top b = 0$, we obtain $\sum_{i \in I'} b_i = 0$.
	
	The set $I'$ cannot contain both $m+1$ and $m+2$.
	Otherwise, the entries $2a_1, \ldots, 2a_m$ indexed by $I'$ would have to sum to $2M$, so $[m]\subseteq I'$ and $|I'|\ge m+2$, a contradiction.
	The set $I'$ also contains at least one of $m+1$ and $m+2$.
	Otherwise, since $|I'|=m+1$ and $b$ has only $m-1$ zero entries, we have $I'\cap[m]\ne\emptyset$, and thus $\sum_{i \in I'} b_i \ge \min_{i\in [m]} 2a_i > 0$, a contradiction.
	Thus $I'$ contains exactly one of $m+1$ and $m+2$, and $\bar I:=I'\cap[m]$ satisfies 
	\begin{align*}
		\sum_{i \in \bar I} 2a_i = M,
	\end{align*}
	which gives the desired partition.
\end{proof}

The next observation shows that Theorem~\ref{thm:NPhard} cannot be extended to the case $r=1$.

\begin{observation}
	\label{obs:rank_one}
	Let $r=1$ and $R=\rho\in\Q^{n}$, and let $K:=\min\{n,\sum_{i=1}^d s_i\}$.
	Then $\opt(\rho;s)$ is the sum of the $K$ largest values among $\rho_1^2,\ldots,\rho_n^2$, and an optimal solution can be computed exactly in polynomial time.
\end{observation}

\begin{proof}
	For a set $P\subseteq[n]$, denote by $\rho_P$ the vector that agrees with $\rho$ on $P$ and is zero elsewhere.
	Let $X\in\mf_s$, and let $U$ be the union of the supports of the columns of $X$.
	Since the columns of $X$ are orthonormal, $\fnorm{\rho^\top X}^2=\norm{XX^\top\rho}^2$ is the squared norm of the projection of $\rho$ onto the column space of $X$.
	This space is contained in $\{x\in\R^n:\supp(x)\subseteq U\}$, so $\fnorm{\rho^\top X}^2\le\norm{\rho_U}^2$.
	Since $|U|\le K$, the value $\norm{\rho_U}^2$ is at most the sum of the $K$ largest values $\rho_j^2$.
	
	Conversely, let $U^*$ be a set of $K$ indices with the largest values $\rho_j^2$.
	Since $d\le n$ and $s_i\ge1$ for all $i\in[d]$, we have $d\le K\le\sum_{i=1}^d s_i$.
	Hence we can partition $U^*$ into nonempty sets $P_1,\ldots,P_d$ with $|P_i|\le s_i$ for all $i\in[d]$.
	For every $i\in[d]$, let $X_i:=\rho_{P_i}/\norm{\rho_{P_i}}$ if $\rho_{P_i}\ne0$, and let $X_i:=\mathbf e_j$ for some $j\in P_i$ otherwise.
	The columns of $X$ have disjoint supports, so $X\in\mf_s$.
	Moreover, $\fnorm{\rho^\top X}^2=\sum_{i=1}^d(\rho^\top X_i)^2=\sum_{i=1}^d\norm{\rho_{P_i}}^2=\norm{\rho_{U^*}}^2$.
	The set $U^*$ and the matrix $X$ can be computed in polynomial time, where each column of $X$ is represented as a rational vector divided by the square root of a rational number.
\end{proof}

Next, we prove Theorem~\ref{thm:NP-hard_linear}.
As in the proof of Theorem~\ref{thm:NPhard}, we reduce from $\Partition$.

\linearHardnessTheorem*

\begin{proof}[Proof of Theorem~\ref{thm:NP-hard_linear}]
	We reduce $\Partition$ to the decision version of the linear problem~\eqref{eqn:sparse_linear}.
	Let $a_1, a_2, \ldots, a_m$ be an instance of $\Partition$, and let $M:= \sum_{i = 1}^m a_i$.
	After doubling all $a_i$, we may assume that $M$ is even.
	
	\textit{Construction.}
	Let $q:= \lceil \sqrt{m+2}\,\rceil$, $k:= q^2$, $n:=m+k+1$, $Q:=\sum_{i=1}^m a_i^2 + (M/2)^2$, and $L_0 := Q + 1$.
	We define $A = (A_1, A_2) \in \Z^{n\times 2}$ as follows.
	The first $m + k$ entries of $A_1$ are one, and the last entry of $A_1$ is zero.
	The first $m$ entries of $A_2$ are $2a_1, 2a_2, \ldots, 2a_m$, the $(m+1)$-th entry of $A_2$ is $-M$, the last entry of $A_2$ is $Q-1$, and the remaining $k-1$ entries of $A_2$ are zero.
	Finally, let $\tau_0:= q + L_0$.
	The instance $(A,k,\tau_0)$ can be computed in time polynomial in the encoding size of $a_1,\ldots,a_m$.
	
	We have $\norm{A_2}^2 = 4Q + (Q-1)^2 = (Q+1)^2 = L_0^2$.
	Let $(x, y)\in\mf_{(k,k)}$.
	Since the last entry of $A_1$ is zero and $\znorm{x}\le k$, the Cauchy--Schwarz inequality gives
	\begin{align}
		\label{eqn:CS_hardness}
		\begin{split}
			A_1^\top x &\le \norm{(A_1)_{\supp(x)}}\norm{x} = \sqrt{\abs{\supp(x)\setminus\{n\}}} \le \sqrt{k} = q,\\
			A_2^\top y &\le \norm{A_2} \norm{y} = L_0,
		\end{split}
	\end{align}
	where the vector $(A_1)_{\supp(x)}$ agrees with $A_1$ on $\supp(x)$ and is zero elsewhere.
	Hence $\optlin(A; (k, k))\le\tau_0$.
	We show that equality holds if and only if the $\Partition$ instance is a yes-instance.
	
	\textit{From a partition to a solution.}
	Suppose that $I\subseteq[m]$ satisfies $\sum_{i\in I} a_i = M / 2$.
	Let $F$ be a set of $k - 1 - |I|$ indices in $\{m+2,\ldots,m+k\}$, which index zero entries of $A_2$.
	Let $S:= I \cup \{m + 1\} \cup F$, so that $\abs{S} = k = q^2$ and $S\subseteq[m+k]$.
	Let $x:= \dso_S / q$, $y := A_2 / L_0$, and $X:= (x, y)$.
	Both vectors have unit norm, $\znorm{x} = k$, and $\znorm{y} = m+2 \le k$.
	Moreover, $x^\top y = (\sum_{i\in I}2a_i - M)/(qL_0) = 0$.
	Hence $X\in\mf_{(k, k)}$, and
	\begin{align*}
		\tr(A^\top X) = A_1^\top x + A_2^\top y = \frac{k}{q} + \frac{\norm{A_2}^2}{L_0} = q + L_0 = \tau_0.
	\end{align*}
	
	\textit{From a solution to a partition.}
	Suppose that $\optlin(A; (k, k)) \ge \tau_0$, and let $X^*= (x^*, y^*)\in\mf_{(k, k)}$ satisfy $\tr(A^\top X^*) \ge \tau_0$.
	Then all inequalities in \eqref{eqn:CS_hardness} hold with equality.
	Thus $S:= \supp(x^*)$ has cardinality $k$, $n\notin S$, $x^* = \dso_S / q$, and $y^* = A_2 / L_0$.
	Since $(x^*)^\top y^*=0$, we have $\sum_{i \in S} (A_2)_i = 0$.
	Since $|S|=k$, $S\subseteq[m+k]$, and $A_2$ has exactly $k-1$ zero entries in $[m+k]$, we have $S\cap[m+1]\ne\emptyset$.
	If $m+1\notin S$, then $\sum_{i \in S} (A_2)_i\ge\min_{i\in[m]} 2a_i>0$, a contradiction.
	Hence $m+1\in S$, and $I:= S\cap [m]$ satisfies
	\begin{align*}
		0 = \sum_{i \in S} (A_2)_i = \sum_{i\in I} 2a_i - M.
	\end{align*}
	Thus $\sum_{i\in I} a_i = M/2$, which gives the desired partition.
\end{proof}

\subsection{Proofs for Section~\ref{sec:motivation}}
\label{app:proofs_section2}
In this section, we prove the bounds on the hard rows and hard variables, the solution of~\eqref{eqn:rotation_PCA}, and Proposition~\ref{prop:two_column_correctness}.

\begin{lemma}
	\label{lem:number_hard_row}
	Let $X$ be a candidate solution with all its rows assigned row types.
	Then
	\begin{align*}
		b(X) &\le \sum_{e\in H(X)} m_e(X) \le (d-1) \abs{H(X)}, \\
		D(X) &\le 2\sum_{e\in H(X)} m_e(X) \le 2(d-1) \abs{H(X)}.
	\end{align*}
\end{lemma}

\begin{proof}
	Every hard row contains a hard pair, so it is counted in $m_e(X)$ for some $e\in H(X)$.
	This gives $b(X) \le \sum_{e\in H(X)} m_e(X)$.
	Each occurrence of a hard pair in the assigned row type of a row gives at most two hard variables in that row.
	This gives $D(X) \le 2\sum_{e\in H(X)} m_e(X)$.
	Since $m_e(X)\le d-1$ for every $e\in H(X)$, the remaining two inequalities hold.
\end{proof}

\begin{proposition}
	\label{prop:rotation_PCA}
	Let $\Gamma\in \R^{h\times k}$ with $k\le h$, and let $\Lambda\in \R^{k\times k}$ be positive semidefinite.
	Then the optimal value of~\eqref{eqn:rotation_PCA} is $\tr(\sqrt{\Gamma\Lambda\Gamma^\top})$.
	Moreover, an optimal solution is $Z^* = U V^\top \sqrt{\Lambda}$, where $\Gamma\sqrt{\Lambda}=U\Sigma V^\top$ with $U\in\R^{h\times k}$, $U^\top U=I_k$, $\Sigma\in\R^{k\times k}$ diagonal and non-negative, and $V\in\R^{k\times k}$ orthogonal, i.e., a thin singular value decomposition including zero singular values.
\end{proposition}

\begin{proof}[Proof of Proposition~\ref{prop:rotation_PCA}]
	We first show that every $Z\in\R^{h\times k}$ with $Z^\top Z=\Lambda$ can be written as $Z=\Psi \sqrt{\Lambda}$, where $\Psi\in \R^{h\times k}$ satisfies $\Psi^\top \Psi = I_k$.
	Indeed, let $\rho:= \rk(Z) \le \min\{k, h\} = k$.
	By the singular value decomposition, there are orthogonal matrices $P_1\in \R^{h\times h}$ and $P_2\in \R^{k\times k}$ such that
	\begin{align*}
		Z = P_1 
		\begin{pmatrix}
			\sigma_1 & & & \\
			& \ddots & & \\
			& & \sigma_\rho & \\
			& & & O
		\end{pmatrix}
		P_2
		\quad
		\xRightarrow{Z^\top Z = \Lambda}
		\quad
		\Lambda = P_2^\top 
		\begin{pmatrix}
			\sigma_1^2 & & & \\
			& \ddots & & \\
			& & \sigma_\rho^2 & \\
			& & & O
		\end{pmatrix}
		P_2,
	\end{align*}
	which implies that $Z = P_1 \begin{pmatrix}
		I_k\\
		O
	\end{pmatrix}
	P_2 \sqrt{\Lambda}$.
	Conversely, every $\Psi$ with $\Psi^\top\Psi=I_k$ gives a feasible $Z=\Psi\sqrt \Lambda$.
	Therefore,
	\begin{align*}
		\max_{Z^\top Z = \Lambda} \tr(\Gamma^\top Z) 
		&= \max_{\Psi^\top \Psi = I_k} \tr(\Gamma^\top \Psi \sqrt{\Lambda})
		= \max_{\Psi^\top \Psi = I_k} \tr(\sqrt{\Lambda} \Gamma^\top \Psi)\\
		&= \nnorm{\sqrt{\Lambda} \Gamma^\top}
		= \tr(\sqrt{\Gamma\Lambda\Gamma^\top}),
	\end{align*}
	where $\nnorm{\cdot}$ is the nuclear norm.
	The third equality follows from von Neumann's trace inequality~\cite{Mir75}, applied after padding $\sqrt{\Lambda} \Gamma^\top$ and $\Psi$ with zeros to square matrices, which gives $\tr(\sqrt{\Lambda} \Gamma^\top \Psi) \le \nnorm{\sqrt{\Lambda} \Gamma^\top} \norm{\Psi} \le \nnorm{\sqrt{\Lambda} \Gamma^\top}$, and from the fact that $\Psi = UV^\top$ attains equality, where $\Gamma\sqrt \Lambda=U\Sigma V^\top$ is the thin singular value decomposition in the statement.
	Indeed, $\tr(\sqrt{\Lambda} \Gamma^\top UV^\top)=\tr(V\Sigma U^\top UV^\top)=\tr(\Sigma)=\nnorm{\Gamma\sqrt \Lambda}$.
	Hence $Z^*=UV^\top\sqrt \Lambda$ is an optimal solution.
\end{proof}

Next, we show Proposition~\ref{prop:two_column_correctness}, which states the correctness of Algorithm~\ref{alg:two_column}.

\twoColumnCorrectnessProposition*

\begin{proof}[Proof of Proposition~\ref{prop:two_column_correctness}]
	Fix $e = \{i, l\}\in S$, and let $\hat Y$ and $\bar Y$ be the values of $Y$ in Algorithm~\ref{alg:two_column} immediately before and after the iteration of the for-loop corresponding to $e$.
	
	(1) In the first part of the proof, we assume that $G=\hat Y^\top\hat Y$ at the beginning of the iteration, so that $g=\hat Y_i^\top\hat Y_l$.
	We show that $\bar Y^\top \bar Y=\hat Y^\top\hat Y+E_eD_eE_e^\top$.
	In particular, $\bar Y_i^\top \bar Y_l = 0$, the entries $\bar Y_i^\top \bar Y_i$ and $\bar Y_l^\top \bar Y_l$ increase by $\abs{g}$, and no other entry of the Gram matrix changes.
	Since line~10 updates $G$ by the same matrix $E_eD_eE_e^\top$, the equality $G=Y^\top Y$ is preserved, and by induction it holds at the beginning of every iteration.
	We also show that $\bar Y$ differs from $\hat Y$ only in entries $(j,p)$ with $j\in J_e$ and $p\in e$, and that $\tr(A^\top \bar Y)\ge \tr(A^\top \hat Y)$.
	If $g=0$, then $\bar Y=\hat Y$ and there is nothing to show.
	
	Assume $g\ne0$, and assume without loss of generality that $e = \{1, 2\}$ and that $J_e$ consists of the first $d$ rows.
	Then, with $Z$ and $B$ as in Algorithm~\ref{alg:two_column}, the matrix $Y$ has the block form
	\begin{align*}
		Y = \begin{pmatrix}
			Z & B \\
			Y_{[n]\backslash J_e, e} & Y_{[n]\backslash J_e, [d]\backslash e}
		\end{pmatrix},
	\end{align*}
	and therefore
	\begin{align*}
		Y^\top Y = \begin{pmatrix}
			Z^\top Z + Y_{[n]\backslash J_e, e}^\top Y_{[n]\backslash J_e, e} &
			Z^\top B + Y_{[n]\backslash J_e, e}^\top Y_{[n]\backslash J_e, [d]\backslash e}\\
			B^\top Z + Y_{[n]\backslash J_e, [d]\backslash e}^\top Y_{[n]\backslash J_e, e} &
			B^\top B + Y_{[n]\backslash J_e, [d]\backslash e}^\top Y_{[n]\backslash J_e, [d]\backslash e}
		\end{pmatrix}.
	\end{align*}
	The iteration changes only the block $Z$.
	Hence, to compute $\bar Y^\top \bar Y$, it suffices to compute the new values of $Z^\top Z$ and $B^\top Z$.
	Let $\hat Z:= \hat Y_{J_e, e}$ and $\bar Z:= \hat Z - NN^\top \hat Z + NK_{\rm new}$ be the initial and updated values of $Z$ in the iteration.
	
	We first show that, on line~7 of Algorithm~\ref{alg:two_column}, the matrix $N$ exists, that is, $\ker(B^\top)$ has dimension at least $2$.
	Indeed, $B^\top\in\R^{(d-2)\times d}$, so $\rk(B^\top)\le d-2$, and thus $\ker(B^\top)\subseteq\R^d$ has dimension at least $d- (d-2) = 2$.
	This also covers the case $d=2$, where $B$ has no columns and $\ker(B^\top)=\R^2$.
	
	Next, we compute the new value of $Z^\top Z$.
	As in Algorithm~\ref{alg:two_column}, let $K:=N^\top\hat Z$ and $Z_0:= \hat Z - NK$.
	Since $N^\top N=I_2$, we have $N^\top Z_0=O$, and therefore
	\begin{align*}
		\bar Z^\top \bar Z 
		& = (Z_0 + NK_{\rm new})^\top (Z_0 + NK_{\rm new}) 
		= Z_0^\top Z_0 + K_{\rm new}^\top N^\top N K_{\rm new}\\
		& = Z_0^\top Z_0 + K_{\rm new}^\top K_{\rm new} 
		= Z_0^\top Z_0 + \Lambda_e\\
		& = Z_0^\top Z_0 + K^\top K + D_e 
		= \hat Z^\top \hat Z + D_e,
	\end{align*}
	where the last equality follows from $\hat Z = Z_0 + NK$ and $N^\top Z_0=O$.
	Since the rows outside $J_e$ do not change, we obtain $\bar Y_i^\top\bar Y_l=\hat Y_i^\top\hat Y_l+(D_e)_{12}=g-\abs{g}\operatorname{sgn}(g)=0$, and the diagonal entries $\bar Y_i^\top\bar Y_i$ and $\bar Y_l^\top\bar Y_l$ increase by $(D_e)_{11}=(D_e)_{22}=\abs{g}$.
	
	Then, we compute the new value of $B^\top Z$.
	By the choice of $N$, we have $B^\top N = O$, and therefore
	\begin{align*}
		B^\top \bar Z = B^\top (Z_0 + NK_{\rm new}) = B^\top Z_0 = B^\top \hat Z.
	\end{align*}
	Therefore, except for the entries indexed by $e\times e$, the Gram matrix does not change, and $\bar Y^\top \bar Y=\hat Y^\top\hat Y+E_eD_eE_e^\top$.
	Moreover, only the entries of $Y_{J_e,e}$ change.
	
	Finally, we show that $\tr(A^\top \bar Y)\ge\tr(A^\top \hat Y)$.
	Since only $Z$ changes, it suffices to compare the values of $\tr(A_{J_e, e}^\top Z)$.
	We claim that
	\begin{align*}
		\tr(A_{J_e, e}^\top N K_{\rm new}) \ge \tr(A_{J_e, e}^\top N K).
	\end{align*}
	Indeed, by Proposition~\ref{prop:rotation_PCA},
	\begin{align*}
		\tr(A_{J_e, e}^\top N K_{\rm new}) &= \tr(\sqrt{N^\top A_{J_e, e} (K^\top K + D_e) A_{J_e, e}^\top N})\\
		& \ge \tr(\sqrt{N^\top A_{J_e, e} K^\top K A_{J_e, e}^\top N}) \\
		& \ge \tr(A_{J_e, e}^\top N K),
	\end{align*}
	where the equality holds by Proposition~\ref{prop:rotation_PCA} with $\Gamma=\Gamma_e=N^\top A_{J_e,e}$ and $\Lambda=\Lambda_e$.
	The proposition applies because $D_e$ has eigenvalues $0$ and $2\abs{g}$, and thus $\Lambda_e=K^\top K+D_e\succeq0$.
	The first inequality follows from
	\begin{align*}
		N^\top A_{J_e, e} (K^\top K + D_e) A_{J_e, e}^\top N \succeq 
		N^\top A_{J_e, e} K^\top K A_{J_e, e}^\top N
	\end{align*}
	and from Weyl's monotonicity theorem: if $X\succeq Y\succeq 0$, then $\lambda_k(X)\ge\lambda_k(Y)$ for every $k$, and hence $\tr(\sqrt X)\ge\tr(\sqrt Y)$.
	The second inequality follows from Proposition~\ref{prop:rotation_PCA} with $\Lambda=K^\top K$, since $K$ is feasible for \eqref{eqn:rotation_PCA} with this $\Lambda$.
	Therefore,
	\begin{align*}
		\tr(A_{J_e, e}^\top \bar Z) = \tr(A_{J_e, e}^\top (Z_0 + N K_{\rm new}))
		\ge \tr(A_{J_e, e}^\top (Z_0 + N K))
		= \tr(A_{J_e, e}^\top \hat Z).
	\end{align*}
	
	(2) In the second part of the proof, we prove the statement.
	Initially $G=X^\top X$, so by part (1), $G=Y^\top Y$ at the beginning of every iteration.
	By part (1), the entry $(i,l)$ of the Gram matrix changes only in the iteration of $\{i,l\}$, where it becomes zero.
	Hence the value $g$ used in the iteration of $\{i,l\}$ equals $(X^\top X)_{il}$, and this iteration adds $\abs{(X^\top X)_{il}}$ to the diagonal entries $(i,i)$ and $(l,l)$.
	At the end of Algorithm~\ref{alg:two_column}, the entry $(i,l)$ is zero for every $\{i, l\}\in S$.
	It is also zero for every $\{i, l\} \in H$, since $(X^\top X)_{il}=0$ and such entries never change.
	Thus $\bar X^\top \bar X$ is diagonal, and $(\bar X^\top \bar X)_{ii}=(X^\top X)_{ii}+\sum_{l\ne i}\abs{(X^\top X)_{il}}$, where the terms with $\{i,l\}\in H$ are zero.
	Each iteration changes only entries $(j,p)$ with $j\in J_e$ and $p\in e\subseteq\rt_j$, so $\bar X_{jp}=X_{jp}$ whenever $p\notin\rt_j$.
	Finally, the objective value does not decrease in any iteration, so $\tr(A^\top\bar X) \ge \tr(A^\top X)$.
\end{proof}

\subsection{Proof of Theorem~\ref{thm:main_approximation}}
\label{app:approximation_guarantee}
In this section, we prove Theorem~\ref{thm:main_approximation}.
We first prove the approximation guarantee of $\LinearOracle$ (Algorithm~\ref{alg:linear_oracle}).
Afterwards, we use it to prove the guarantee for Algorithm~\ref{alg:main_alg}, our main approximation algorithm for \ref{eqn:SPCA}.

We first show that every output of $\LinearOracle$ is in $\mf_s$.
This is clear for the initial matrix $X_{\rm best}$ and for the output in the case $d=1$.
Consider an accepted root state of $\SelectBranch$, with candidate solution $Y$ and assigned row types $(\rt_j)_{j\in[n]}$.
At the root, $\locc=[b]$, and the occupied labels of the two children are disjoint at every merge.
Hence every hard row type $T^r\in\ml$ is assigned to exactly one row, whose hard variables take the values $p_r$.
Moreover, ordinary row types contain no pair of $H$.
Therefore, for every $\{i,l\}\in H$, the only rows that contribute to $Y_i^\top Y_l$ are hard rows whose row type contains $\{i,l\}$, and \eqref{eqn:hard_constraints} gives $Y_i^\top Y_l=\sum_{r:\{i,l\}\subseteq T^r}p_{ri}p_{rl}=0$.
At the root, we also have $\mu_e=d$ for every $e\in S$, so every simple pair is contained in at least $d$ assigned row types.
Thus, Proposition~\ref{prop:two_column_correctness} applies to $Y$.
By this proposition, the output $\widetilde Y$ of Algorithm~\ref{alg:two_column} has orthogonal columns, and $\widetilde Y_{ji}=Y_{ji}=0$ whenever $i\notin\rt_j$, since $\LeafInit$ sets the entries outside the row type to zero.
The number of rows whose assigned row type contains $i$ is at most $\hat c_i\le s_i$, so $\znorm{\widetilde Y_i}\le s_i$.
Moreover, $\norm{\widetilde Y_i}^2\ge G^0_{ii}\ge1-\delta>0$ by the root filter of $\SelectBranch$, so the normalization in $\ExactFinalize$ (Algorithm~\ref{alg:exact_finalize}) is well defined.
The normalization and the sign flips change neither the supports nor the orthogonality of the columns.
Therefore, the output of $\ExactFinalize$, and hence of $\LinearOracle$, is in $\mf_s$.
It remains to show that the output of $\LinearOracle$ has a sufficiently large objective value.

\paragraph{High-level idea.}
Let $X^*$ be an optimal solution of~\eqref{eqn:sparse_linear}.
We show that there exists a matrix $Z_0$ that is close to $X^*$ and whose branch of $\LinearOracle$ yields an accepted root candidate in $\SelectBranch$ (Algorithm~\ref{alg:select_branch}).
The output of $\LinearOracle$ is at least as good as the matrix that $\ExactFinalize$ obtains from this candidate, and this gives the approximation guarantee.
We use the chain of matrices $X^* \to Z \to Z_0 \to Y \to \widetilde Y \to X$ described in the proof idea of Theorem~\ref{thm:main_approximation} in Section~\ref{sec:guarantee}.

\paragraph{Notation.}
If $A = O$, every matrix in $\mf_s$ is optimal, and $\LinearOracle$ returns $X_{\rm best}\in\mf_s$.
If $d=1$, $\LinearOracle$ returns an optimal solution before it enumerates any branch.
Thus, we assume that $A\ne O$ and $d\ge2$.
Then $\optlin(A;s)>0$.
Indeed, if $A_{ji}\ne0$, let $X_i:=\sign(A_{ji})\mathbf e_j$, and let the other columns of $X$ be distinct standard basis vectors $\mathbf e_k$ with $k\ne j$, multiplied by $\pm1$ so that $A_l^\top X_l\ge0$.
Then $X\in\mf_s$ and $\tr(A^\top X)\ge|A_{ji}|>0$.
For the $j$-th row $x^*_j$ of $X^*$, we define its \emph{objective contribution} $q_j:=a_j (x^*_j)^\top$, where $a_j$ is the $j$-th row of $A$. 
Flipping the sign of a row $x^*_j$ changes neither $(X^*)^\top X^*$ nor the supports, and it replaces $q_j$ by $-q_j$.
Hence the optimality of $X^*$ gives $q_j\ge 0$ for every $j\in[n]$.

We next define the analogs for $X^*$ of the objects in Section~\ref{sec:motivation_row_type}.
For every row $x^*_j$ of $X^*$, let $\rt_j^*:=\supp(x^*_j)$ be its row type.
For a pair $e\in\cp_d$, let $m_e^*:= \abs{\{j:e\subseteq \rt_j^*\}}$.
Let $H^*:= \{e:m_e^*<d\}$ and $S^*:=\cp_d\setminus H^*$ be the sets of hard and simple pairs, respectively.

The rows whose row types contain at least one pair in $H^*$ give the list of hard row types $\ml^* := \multiset{(T^*)^1, (T^*)^2, \ldots, (T^*)^{b^*}}$.  
For a hard row type $(T^*)^r$, the variables corresponding to $U_{H^*}((T^*)^r) = \cup_{e\in H^*, e\subseteq (T^*)^r} e$ are the hard variables, which are fixed by $\FixHardVariables$.
By Lemma~\ref{lem:number_hard_row}, $b^*\le\min\{n,(d-1)|H^*|\}$, and $D^* := \sum_{r=1}^{b^*}|U_{H^*}((T^*)^r)| \le d(d-1)^2$.
Moreover, every element of $\ml^*$ contains a pair of $H^*$, and every $e\in H^*$ is contained in $m_e^*\le d-1$ elements of $\ml^*$.
Hence $\LinearOracle$ enumerates the branch with $H=H^*$ and $\ml=\ml^*$.
We label the hard rows of $X^*$ so that the row with label $r$ has the row type $T^r$ of the multiset $\ml$ enumerated by $\LinearOracle$, and all labels below refer to this labeling.

Before constructing $Z_0$, we need a technical lemma.

\begin{lemma}
	\label{lem:fnorm_A_in_supp}
	Let $X^*\in\mf_s$, let $\mathcal I:= \{(j, i): X^*_{ji} \ne 0\}$, and let $\bar A \in \R^{n\times d}$ be the matrix with $\bar A_{ji} = A_{ji}$ for $(j, i)\in \mathcal I$ and $\bar A_{ji} = 0$ otherwise.
	Then, $\fnorm{\bar A} \le d\optlin(A;s)$.
\end{lemma}

\begin{proof}
	Fix a column $i\in [d]$, and let $S_i^*:= \supp(X^*_i)$.
	Then $\norm{\bar A_i}^2=\sum_{j\in S_i^*} A_{ji}^2$.
	If $\abs{S_i^*} \le n - d +1$, let $Q_i := S_i^*$.
	Otherwise, let $Q_i\subseteq S_i^*$ be a set of $n-d+1$ indices $j$ with the largest values $A_{ji}^2$.
	Since $|S_i^*|\le n$ and $n\ge d$, we have
	\begin{align}
		\label{eqn:upper_bound_Ai}
		\sum_{j\in Q_i} A_{ji}^2 
		\ge \min\left\{\frac{n - d + 1}{|S_i^*|}, 1\right\} \sum_{j\in S_i^*} A_{ji}^2
		\ge \frac{n - d + 1}{n} \sum_{j\in S_i^*} A_{ji}^2
		\ge \frac{1}{d} \norm{\bar A_i}^2.
	\end{align}
	
	We next show that $\optlin(A;s)\ge \norm{\bar A_i} / \sqrt{d}$.
	We may assume that $\bar A_i\ne 0$.
	Let $\tilde a\in\R^n$ agree with $A_i$ on $Q_i$ and be zero elsewhere.
	By \eqref{eqn:upper_bound_Ai}, we have $\norm{\tilde a}\ge\norm{\bar A_i}/\sqrt d>0$.
	Let $X_i:=\tilde a/\norm{\tilde a}$.
	Since $|Q_i| \le n - d + 1$, there are at least $d-1$ indices outside $Q_i$.
	For $l\ne i$, let $X_l$ be $\pm\mathbf e_j$ for distinct indices $j\notin Q_i$, with the sign chosen so that $A_l^\top X_l \ge 0$.
	Then $X^\top X = I_d$, $\znorm{X_i}\le|Q_i|\le|S_i^*|\le s_i$, and $\znorm{X_l}=1\le s_l$ for $l\ne i$.
	Hence $X\in\mf_s$, and
	\begin{align*}
		\optlin(A;s) \ge \tr(A^\top X) = \norm{\tilde a} + \sum_{l \ne i}  A_l^\top X_l \ge \norm{\tilde a} \ge \frac{1}{\sqrt{d}} \norm{\bar A_i}.
	\end{align*}
	Squaring and summing over $i\in [d]$, we obtain
	\begin{align*}
		d^2 \optlin(A;s)^2 \ge \sum_{i=1}^d \norm{\bar A_i}^2 = \fnorm{\bar A}^2.
	\end{align*}
	Since $\optlin(A;s)\ge 0$, this gives $\fnorm{\bar A} \le d\optlin(A;s)$.
\end{proof}

In the next proposition, we show that there exists a matrix $Z_0$ that is close to $X^*$ and passes all filters in $\SelectBranch$, unless a better candidate replaces it.

\begin{proposition}
	\label{prop:finite_witness}
	For $H=H^*$ and $\ml=\ml^*$ in $\LinearOracle$, there exist a vector $p$ returned by $\FixHardVariables$ and a matrix $Z_0$ whose rows are among the enumeration of $\SelectBranch$ for the branch $(H^*,\ml^*,p)$ such that
	\begin{align}
		\tr(A^\top Z_0)	&\ge \left(1-\frac{\delta}{4}\right)\optlin(A;s), \label{eqn:Z0_objective}\\
		\fnorm{Z_0^\top Z_0-I_d} &\le \frac{\delta}{4}. 	\label{eqn:Z0_gram}
	\end{align}
	Moreover:
	(i) $(Z_0)_i^\top (Z_0)_l = 0$ for every $\{i, l\}\in H^*$.
	(ii) Every labeled hard row type in $\ml^*$ is assigned exactly once to a row of $Z_0$.
	(iii) Every simple pair $e\in S^*$ is contained in the assigned row types of at least $d$ rows of $Z_0$.
	(iv) At every node of the ADP tree, the capacity caps computed along the rows of $Z_0$, and the sums of the caps of the two children, are at most $s$ componentwise.
\end{proposition}

\begin{proof}
	We first construct a matrix $Z$ from $X^*$.
	We keep every hard row of $X^*$ and, for each simple pair $e\in S^*$, any $d$ rows whose row types contain $e$.
	The number of kept rows is at most $(d-1)\abs{H^*} + d\abs{S^*}\le P_d = d\binom d2$.
	
	Next, we delete a small number of the remaining rows.
	We define the \emph{weight} of each row $j\in [n]$ by $w_j:= q_j / \tr(A^\top X^*) + \norm{x^*_j}^2 / d$.
	Since $\sum_{j\in [n]} q_j = \tr(A^\top X^*)$ and $\fnorm{X^*}^2 = d$, we have $\sum_{j\in [n]} w_j=2$.
	Using the parameter $\tau$ in Table~\ref{tab:parameters}, we delete rows of small weight, column by column.
	For each column $i$, we proceed as follows.
	\begin{itemize}
		\item If $s_i \le K_0 = \lceil 16(P_d + 1) / \tau \rceil$, we delete no row for this column.
		\item Otherwise, we define $t_i := \lfloor (1-4\tau) s_i \rfloor$, and compute the current sparsity capacity usage of this column $c_i$.
		If $c_i > t_i$, we delete the $c_i - t_i$ lowest-weight rows $j$ that are currently undeleted and unkept and satisfy $i\in \rt_j^*$ (or equivalently, $j\in \supp(X_i^*)$).
	\end{itemize}
	Deleting a row means setting the entire row to zero, not just its entry in column $i$, and replacing its assigned row type by $\emptyset$.
	We denote the resulting matrix by $Z$.
	
	We claim that the cumulative weight of the removed rows never exceeds $10d\tau$.
	Indeed, consider a column $i$ with $s_i > K_0$ and $c_i > t_i$.
	By the definitions of $K_0$ and $t_i$, we have $P_d + 1 < \tau s_i / 16$, and $t_i \ge (1-4\tau) s_i - 1$.
	Since $\tau = \delta / (128 d^2) < 1/32$, we have
	\begin{align*}
		c_i - P_d \ge t_i + 1 - P_d > (1-4\tau) s_i - P_d > \left(1 - 4\tau -\frac{\tau}{16}\right) s_i > 0.
	\end{align*}
	Let $C_i$ be the set of rows $j$ that are currently undeleted and unkept and satisfy $i\in\rt_j^*$.
	Then $|C_i|\ge c_i-P_d$, and $c_i-t_i< c_i-P_d$ because $t_i\ge(1-4\tau)s_i-1>\tau s_i/16-1>P_d$.
	Hence $C_i$ contains enough rows to delete.
	The weights are non-negative, and we delete the $c_i-t_i$ rows of $C_i$ of lowest weight.
	Therefore, the deleted weight is at most
	\begin{align*}
		\frac{c_i-t_i}{|C_i|}\sum_{j\in C_i}w_j
		\le 2\cdot\frac{c_i - t_i}{c_i - P_d} \le 2\cdot\frac{s_i - t_i}{s_i - P_d} \le 2\cdot\frac{4\tau s_i + 1}{s_i - P_d} < 10\tau,
	\end{align*}
	where the second inequality follows from $t_i > P_d$ and $c_i \le s_i$, the third inequality follows from $t_i\ge (1-4\tau) s_i - 1$, and the last inequality is equivalent to $1+5\tau P_d<\tau s_i$, which follows from $\tau s_i>16(P_d+1)$.
	Since there are at most $d$ columns, the removed rows have a total weight of at most $10d\tau$. 
	
	Since all hard rows and the selected $d$ rows for each simple pair in $S^*$ are kept, the deletion affects neither the hard constraints \eqref{eqn:hard_constraints} nor the two-column updates of Algorithm~\ref{alg:two_column}.
	Since the deleted weight is at most $10d\tau$, the deleted rows satisfy $\sum_j q_j\le10d\tau\tr(A^\top X^*)$ and $\sum_j\norm{x_j^*}^2\le10d^2\tau$.
	Moreover, $Z^\top Z-I_d=-\sum_j (x_j^*)^\top x_j^*$, where both sums range over the deleted rows.
	Hence $Z$ satisfies
	\begin{align*}
		\tr(A^\top Z) \ge (1-10d\tau) \optlin(A;s), \qquad
		\fnorm{Z^\top Z-I_d}\le10d^2\tau.
	\end{align*}
	Since $\tau = \delta / (128d^2)$, we obtain
	\begin{align}
		\label{eqn:diff_Z}
		\tr(A^\top Z) \ge \left(1 - \frac{5\delta}{64d}\right) \optlin(A;s), \qquad
		\fnorm{Z^\top Z-I_d}\le \frac{5\delta}{64}.
	\end{align}
	
	Next, let $p^*$ be the values of the hard variables in $Z$.
	Then $p^*\in[-1,1]^{D^*}$, so $p^*$ belongs to one of the boxes $Q^*$ enumerated by $\FixHardVariables$.
	Moreover, $\sum_{r,i}(p^*_{ri})^2\le\fnorm{X^*}^2=d$, and $p^*$ satisfies the orthogonality constraints of the hard pairs in \eqref{eqn:hard_constraints} exactly.
	Indeed, for $\{i,k\}\in H^*$, every row $j$ with $X^*_{ji}X^*_{jk}\ne0$ satisfies $\{i,k\}\subseteq\rt^*_j$, so it is a kept hard row and both entries are hard variables.
	Hence $\FixHardVariables$ returns some $p\in Q^*$ that satisfies \eqref{eqn:hard_constraints}.
	Let $Z^H$ be the matrix obtained from $Z$ by replacing only the hard variables by the values in $p$, while leaving every free variable unchanged.
	Since every side of the box $Q^*$ has length at most $a$, we have
	\begin{align}
		\label{eqn:diff_ZH_Z}
		\fnorm{Z^H - Z} = \norm{p - p^*} \le \sqrt{D^*}a \le \sqrt{D_d} a,
	\end{align}
	where $D_d = d(d-1)^2$ is given in Table~\ref{tab:parameters}.
	Let $\kappa_v:=2^{-J_0}$.
	By the construction of $\mv$, for every free variable $z\in[-1,1]$, we can choose $\hat z\in\mv$ such that
	\begin{align*}
		\abs{z - \hat z} \le 
		\begin{cases}
			\abs{z} / M, & \text{if } |z| \ge \kappa_v, \\
			\kappa_v, & \text{otherwise.}
		\end{cases}
	\end{align*}
	Indeed, if $|z| = 1$, we choose $\hat z = z$.
	If $\kappa_v\le|z| < 1$, we choose $k_z \in [J_0]$ such that $2^{-k_z} \le |z| < 2^{-k_z + 1}$, let $\ell_z:= \lfloor M(2^{k_z} |z| - 1) \rfloor$, and let $\hat z:= \sign(z)2^{-k_z} (1 + \ell_z / M)$.
	Then $\abs{z - \hat z}\le 2^{-k_z} / M\le |z| / M$.
	Otherwise, we choose $\hat z = 0$.
	We round every free variable of $Z^H$ in this way to the finite set $\mv$ defined in \eqref{eqn:free_variable_set}, and we denote the resulting matrix by $Z_0$.
	
	We now show \eqref{eqn:Z0_gram}.
	Let $F$ be the set of indices of the free variables, let $F_{\rm large}:=\{(j, i) \in F: \abs{Z^H_{ji}} \ge \kappa_v\}$, and let $F_{\rm small}:= F\setminus F_{\rm large}$.
	By the triangle inequality, and since $\sum_{(j,i)\in F}(Z^H_{ji})^2\le\fnorm{X^*}^2=d$ and $|F|\le nd$, we have
	\begin{align*}
		\fnorm{Z_0 - Z^H} 
		\le \frac{1}{M} \cdot \sqrt{\sum_{(j, i)\in F_{\rm large}} (Z^H_{ji})^2} 
		+ \kappa_v \cdot \sqrt{\abs{F_{\rm small}}}
		\le \frac{\sqrt{d}}{M} + \kappa_v \sqrt{nd}.
	\end{align*}
	Together with \eqref{eqn:diff_ZH_Z} and the triangle inequality, this gives
	\begin{align*}
		\fnorm{Z_0 - Z} \le \fnorm{Z_0 - Z^H} + \fnorm{Z^H - Z} 
		\le \frac{\sqrt{d}}{M} + \kappa_v \sqrt{nd} + \sqrt{D_d} a.
	\end{align*}
	Recall the choices of $a$, $M$, and $J_0$ in Table~\ref{tab:parameters}:
	\begin{align*}
		a = \frac{\delta}{256 (d+1) (D_d + 1)}, \quad
		M = \left\lceil \frac{256d(d+1)}{\delta} \right\rceil, \quad
		J_0 = \left\lceil \log_2\left(\frac{256dn(d+1)}{\delta}\right) \right\rceil.
	\end{align*}
	These choices give
	\begin{align*}
		\max\left\{\frac{\sqrt{d}}{M}, \kappa_v \sqrt{nd}, \sqrt{D_d} a \right\} \le \frac{\delta}{256 (d+1)},
	\end{align*}
	and therefore $\fnorm{Z_0 - Z} < \delta / (64(d+1))$.
	Since $\fnorm{Z}\le\fnorm{X^*}=\sqrt d$, we obtain
	\begin{align*}
		\fnorm{Z_0^\top Z_0 - Z^\top Z} 
		& = \fnorm{(Z_0 - Z)^\top (Z_0 - Z) + (Z_0 - Z)^\top Z + Z^\top (Z_0 - Z)}\\
		&\le \fnorm{Z_0 - Z}^2 + 2 \fnorm{Z} \fnorm{Z_0 - Z}\\
		& <  \left( \frac{\delta}{64(d+1)}\right)^2 + \frac{\delta}{32 \sqrt{d+1}} < \frac{\delta}{16}.
	\end{align*}
	Together with \eqref{eqn:diff_Z}, this gives
	\begin{align*}
		\fnorm{Z_0^\top Z_0 - I_d} \le \fnorm{Z_0^\top Z_0 - Z^\top Z} + \fnorm{Z^\top Z - I_d}
		< \frac{\delta}{16} + \frac{5\delta}{64} < \frac{\delta}{4}.
	\end{align*}
	
	Then, we show \eqref{eqn:Z0_objective}.
	The matrix $Z_0-Z$ is supported in $\mathcal I$.
	Indeed, the hard variables of a hard row $j$ lie in $\rt_j^*=\supp(x^*_j)$, the free variables lie in $\rt_j^*$, and deleted rows vanish in both matrices.
	Hence $\tr(A^\top(Z_0-Z))=\tr(\bar A^\top(Z_0-Z))$, and by the Cauchy--Schwarz inequality and Lemma~\ref{lem:fnorm_A_in_supp},
	\begin{align*}
		\abs{\tr(A^\top (Z_0 - Z))}\le d\optlin(A;s)\fnorm{Z_0 - Z}. 
	\end{align*}
	Together with \eqref{eqn:diff_Z}, this gives
	\begin{align*}
		\tr(A^\top Z_0) 
		&= \tr(A^\top Z) + \tr(A^\top (Z_0 - Z))\\
		&\ge \tr(A^\top Z) - d\optlin(A;s) \cdot \frac{\delta}{64(d+1)}\\
		& \ge \left(1 - \frac{5\delta}{64d} - \frac{\delta}{64}\right) \optlin(A;s)
		\ge \left(1 - \frac{\delta}{4}\right) \optlin(A;s).
	\end{align*}
	
	Finally, we show (i)--(iv).
	Properties (i)--(iii) hold by the construction of $Z_0$.
	It remains to show (iv).
	The transformations from $Z$ to $Z^{H}$ and from $Z^H$ to $Z_0$ change only variable values, not the assigned row types.
	Hence the sparsity capacity usage $c_i(Z_0)$ of column $i$ equals $c_i(Z)$.
	For a node $u$ of the ADP tree, let $c_{u,i}$ be the number of rows of $Z_0$ in the subtree of $u$ whose assigned row type contains $i$, and let $\hat c_{u,i}$ be the cap computed along these rows.
	At a leaf, $\hat c_{u,i}=c_{u,i}\in\{0,1\}$, and at an internal node $u$ with children $u_L$ and $u_R$, $\hat c_{u,i}=R_c(\hat c_{u_L,i}+\hat c_{u_R,i})\ge\hat c_{u_L,i}+\hat c_{u_R,i}$.
	Therefore, it suffices to show that $\hat c_{u,i}\le s_i$ for every node $u$.
	
	If $s_i\le K_0 = \lceil 16(P_d + 1) / \tau \rceil$, then $R_c(c)=c$ for every integer $c\le K_0$, so no rounding occurs as long as the caps are at most $s_i$.
	By induction on the height of $u$, we obtain $\hat c_{u,i}=c_{u,i}\le c_i(Z_0)\le s_i$.
	
	If $s_i > K_0$, then $c_i(Z_0) \le t_i \le (1-4\tau) s_i$.
	Every sum of two caps is either an integer or larger than $K_0$.
	For every such sum $c\le n$, we have $R_c(c)\le(1+\alpha)c$.
	Indeed, if $c\le K_0$, then $R_c(c)=c$.
	Otherwise, $R_c(c)=K_0(1+\alpha)^l$ for the smallest $l\ge1$ with $K_0(1+\alpha)^l\ge c$, and this $l$ satisfies $K_0(1+\alpha)^{l-1}<c$.
	By induction on the height $h(u)$ of $u$, we obtain
	\begin{align*}
		\hat c_{u,i}\le(1+\alpha)^{h(u)}c_{u,i}\le(1+\alpha)^{H_{\rm tree}}c_i(Z_0)\le e^{\alpha H_{\rm tree}}(1-4\tau)s_i.
	\end{align*}
	Since $\alpha = \tau / (8(H_{\rm tree}+1))$ and $1-4\tau\le e^{-4\tau}$, the right-hand side is at most $e^{\tau/8-4\tau}s_i < s_i$, as desired.
	In the induction step, the induction hypothesis and this bound give $\hat c_{u_L,i}+\hat c_{u_R,i}<s_i\le n$, so the merge is not rejected and the bound $R_c(c)\le(1+\alpha)c$ applies.
\end{proof}

$\KeepBest$ may replace the partial solutions given by the rows of $Z_0$ by other partial solutions with the same key and a larger objective value.
In Proposition~\ref{prop:shadow_state_survival}, we show that the resulting root candidate $Y$ is still nearly orthogonal.
We first need a technical lemma about the Gram matrix bins.

\begin{lemma}
	\label{lem:bin_diameter}
	Let $G,G' \succeq 0$ satisfy $\tr(G),\tr(G') < 4d$ and $\mathrm{lab}(G)=\mathrm{lab}(G')$.
	If $\mathrm{lab}(G)=\mathrm{lab}(G')=(-1,O)$, then $\fnorm{G - G'} < 2 \kappa_G$.
	Otherwise, let $[u, (1+\theta)u)$, with $u=\kappa_G(1+\theta)^k$ for some integer $k\ge0$, be the trace interval of the common label.
	Then $\fnorm{G - G'}\le d \theta u \le d \theta \tr(G)$ and $\tr(G') \le (1+\theta) \tr(G)$.
\end{lemma}

\begin{proof}
	Recall the Gram matrix bins introduced in Section~\ref{sec:motivation_ADP}.
	Let $M$ be a positive semidefinite matrix with vector of eigenvalues $\lambda$.
	Then $\fnorm{M} = \norm{\lambda} \le \onorm{\lambda} = \tr(M)$.
	If both $G$ and $G'$ belong to the small-trace bin $(-1, O)$, then the triangle inequality gives
	\begin{align*}
		\fnorm{G-G'} \le \fnorm{G} + \fnorm{G'} \le \tr(G)+\tr(G') < 2 \kappa_G.
	\end{align*}
	Now assume that the two matrices belong to the same bin with trace interval $[u, (1+\theta)u)$.  
	Since they have the same bin label, we have $\abs{G_{il} - G'_{il}} < \theta u$ for all $i,l\in[d]$.
	Thus $\fnorm{G - G'} \le d \infnorm{G - G'} \le d \theta u \le d\theta \tr(G)$, where $\infnorm{\cdot}$ denotes the largest absolute value of an entry.
	Finally, since both traces lie in $[u,(1+\theta)u)$, we have $\tr(G') < (1+\theta) u \le (1+\theta)\tr(G)$.
\end{proof}

\begin{proposition}
	\label{prop:shadow_state_survival}
	Consider the branch of $\LinearOracle$ that contains the matrix $Z_0$ of Proposition~\ref{prop:finite_witness}.  
	Then $\SelectBranch$ returns a root state whose candidate solution $Y$ satisfies
	\begin{align}
		\tr(A^\top Y)&\ge \tr(A^\top Z_0), \label{eqn:Y_objective}\\
		\fnorm{Y^\top Y-Z_0^\top Z_0} & < \frac{\delta}{8}. \label{eqn:Y_gram_difference}
	\end{align}
	Moreover:
	(i) For every $i\in[d]$, at most $s_i$ rows of $Y$ have an assigned row type that contains $i$.
	(ii) $Y_i^\top Y_l = 0$ for every $\{i, l\}\in H^*$.
	(iii) Every simple pair $e\in S^*$ is contained in the assigned row types of at least $d$ rows of $Y$.
\end{proposition}

\begin{proof}
	We follow the states of $Z_0$ through the ADP tree.
	At every leaf, we start from the state given by the corresponding row of $Z_0$, with its assigned row type and variable values.
	If $\KeepBest$ replaces this state by another state with the same key, we follow the state retained in the dictionary instead.
	At every internal node, we follow the merge of the two followed child states.
	The followed states have the same occupancy sets $\locc$, capped counters $\mu_e$, and capacity caps $\hat c$ as the states of $Z_0$, because these quantities are part of the key.
	Their partial solutions, Gram matrices, and Gram matrix bin labels may differ from those of $Z_0$.
	By Proposition~\ref{prop:finite_witness}~(ii) and~(iv), no followed state is rejected by the occupancy test (line~7) or the capacity test (line~9) of $\SelectBranch$, or by the capacity test in $\KeepBest$.
	
	We next show that the trace test in $\KeepBest$ rejects no followed state.
	Follow the states up to and including the first level, if any, at which one of them fails this test. 
	At each reached node $u$, let $q$ be the followed state immediately before this test. 
	If $q$ passes, let $\bar q$ be the representative finally retained for its key. 
	Since $\mathrm{lab}(G_u(q)) = \mathrm{lab}(G_u(\bar q))$, Lemma~\ref{lem:bin_diameter} gives $\fnorm{G_u(q) - G_u(\bar q)} \le 2\kappa_G$ if $\tr(G_u(q)) < \kappa_G$, and $\fnorm{G_u(q) - G_u(\bar q)} \le d\theta \tr(G_u(q))$ otherwise.
	Moreover, Lemma~\ref{lem:bin_diameter} gives
	\begin{align}
		\label{eqn:shadow_trace_update}
		\tr(G_u(\bar q)) \le (1+\theta) \tr(G_u(q)) + \kappa_G.
	\end{align}
	For $\ell=0, 1, \ldots, H_{\rm tree}$, let $E_\ell$ be the sum of the traces of the actual Gram matrices of the followed states immediately before $\KeepBest$ at level $\ell$ of the ADP tree, where the leaves are at level $0$.
	Then $E_0=\tr(Z_0^\top Z_0)$. 
	By Proposition~\ref{prop:finite_witness}, $\fnorm{Z_0^\top Z_0 - I_d}\le \delta/4$, and thus
	\begin{align*}
		E_0 = d+\tr(Z_0^\top Z_0-I_d) \le d + \sqrt{d} \fnorm{Z_0^\top Z_0 - I_d}
		\le	d+ \frac{\sqrt d \delta}{4}.
	\end{align*}
	
	A merge adds the actual Gram matrices of the two child states.
	Since there are at most $N_{\rm leaf}$ followed states at every level, \eqref{eqn:shadow_trace_update} gives
	\begin{align}
		\label{eqn:shadow_trace_recursion}
		E_{\ell+1} \le (1+\theta) E_\ell + N_{\rm leaf}\kappa_G.
	\end{align}
	Applying \eqref{eqn:shadow_trace_recursion} at most $H_{\rm tree}+1$ times, we obtain, for every level $\ell$,
	\begin{align*}
		E_\ell & \le (1+\theta)^{H_{\rm tree}+1} \left(E_0 + (H_{\rm tree}+1) N_{\rm leaf} \kappa_G\right)\\
		& \le (1+\theta)^{H_{\rm tree}+1} \left(d + \frac{\sqrt d \delta}{4} + (H_{\rm tree}+1) N_{\rm leaf} \kappa_G\right)\\
		& \le \exp\{\theta(H_{\rm tree}+1)\} \cdot \left(d + \frac{\sqrt d \delta}{4} + (H_{\rm tree}+1) N_{\rm leaf} \kappa_G\right)\\
		& = \exp\left\{\frac{\delta}{128d^2}\right\} \cdot \left(d + \frac{\sqrt d \delta}{4} + N_{\rm leaf} \cdot \frac{\delta}{128n}\right)\\
		& \le \frac{128}{127} \cdot \frac{3d}{2} < 2d,
	\end{align*}
	where the equality follows from the definitions $\theta = \delta / (128d^2(H_{\rm tree}+1))$ and $\kappa_G = \delta / (128 n (H_{\rm tree}+1))$.
	The inequalities in the last line follow from $\delta < 1$, from $N_{\rm leaf} < 2n$, and from $\exp\{x\} \le 1/(1-x)$ for $x \in [0, 1)$.
	Hence $\tr(G_u(q)) < 2d$ for every reached node $u$.
	Thus no followed state fails the trace test in $\KeepBest$, and the followed states reach the root.
	
	We now show \eqref{eqn:Y_objective} and (i)--(iii).
	Let $q_u$ be the merged candidate at a node $u$ before $\KeepBest$, and let $\bar q_u$ be the representative retained for its key.
	Since $\KeepBest$ replaces a representative only by one with a larger objective value, $v(\bar q_u)\ge v(q_u)$.
	Moreover, $v(q_u)=v(\bar q_{u_L})+v(\bar q_{u_R})$ for the children $u_L$ and $u_R$ of $u$.
	By induction on the height, $v(\bar q_u)$ is at least the objective value of the rows of $Z_0$ in the subtree of $u$, which gives \eqref{eqn:Y_objective} at the root.
	For (i), the number of rows of $Y$ whose assigned row type contains $i$ is at most $\hat c_i\le s_i$.
	For (ii), the root key has $\locc=[b]$, and the occupied labels are disjoint at every merge, so each hard row type $T^r$ is used exactly once, with hard variables equal to $p_r$.
	Since ordinary row types contain no pair of $H^*$, we obtain $Y_i^\top Y_l=\sum_{r:\{i,l\}\subseteq T^r}p_{ri}p_{rl}=0$ for every $\{i,l\}\in H^*$ by \eqref{eqn:hard_constraints}.
	For (iii), the root key has $\mu_e=d$ for every $e\in S^*$, as the key of $Z_0$ does.
	
	We now show \eqref{eqn:Y_gram_difference}.
	Since the Gram matrices are added at every merge, we have $Y^\top Y=Z_0^\top Z_0+\sum_u\big(G_u(\bar q_u)-G_u(q_u)\big)$, where the sum ranges over all nodes $u$.
	For every followed state at a node $u$, Lemma~\ref{lem:bin_diameter} gives
	\begin{align*}
		\fnorm{G_u(q_u)-G_u(\bar q_u)} \le d\theta\tr(G_u(q_u)) + 2\kappa_G.
	\end{align*}
	Since $E_\ell<2d$ and $N_{\rm leaf}<2n$, the sum over the followed states at level $\ell$ is at most $d\theta E_\ell+2N_{\rm leaf}\kappa_G \le 3\delta / (64(H_{\rm tree}+1))$.
	Summing over the $H_{\rm tree}+1$ levels, we obtain
	\begin{align*}
		\fnorm{Y^\top Y - Z_0^\top Z_0} \le (H_{\rm tree} + 1) \cdot \frac{3\delta}{64 (H_{\rm tree}+1)} < \frac{\delta}{8},
	\end{align*}
	which gives us \eqref{eqn:Y_gram_difference}.
	
	Finally, \eqref{eqn:Z0_gram} and \eqref{eqn:Y_gram_difference} give $\fnorm{Y^\top Y - I_d} < \delta$.
	Hence the root state $q$ of $Y$ passes the filter $\fnorm{G(q) - I_d} \le \delta$ in $\SelectBranch$.
	Moreover, since $A\ne O$, we have $\optlin(A;s) > 0$.
	Hence
	\begin{align*}
		v(q)=\tr(A^\top Y) \ge \tr(A^\top Z_0) \ge (1 - \delta / 4) \optlin(A;s) > 0.
	\end{align*}
	Therefore, $q$ also passes the filter $v(q) \ge 0$, and $\SelectBranch$ returns $q$.
\end{proof}

We are now ready to prove the approximation guarantee of $\LinearOracle$.

\begin{proposition}
	\label{prop:linear_oracle_guarantee}
	Let $A\in\Q^{n\times d}$, $s\in[n]^d$, and $t\in\Q\cap(0,1)$.
	Then $\LinearOracle$ returns $X\in\mf_s$ satisfying
	\begin{align*}
		\tr(A^\top X)\ge(1-t)\optlin(A;s).
	\end{align*}
\end{proposition}

\begin{proof}[Proof of Proposition~\ref{prop:linear_oracle_guarantee}]
	If $A=O$, then every feasible solution is optimal.
	If $d=1$, then the normalized restriction of $A_1$ to $s_1$ entries of largest absolute value is optimal.
	Hence we assume that $d\ge 2$ and $A\ne O$.
	
	Let $Y$ be the root candidate of Proposition~\ref{prop:shadow_state_survival}.
	By Propositions~\ref{prop:finite_witness} and~\ref{prop:shadow_state_survival}, we have $\tr(A^\top Y) \ge (1 - \delta / 4) \optlin(A;s) > 0$ and $\fnorm{Y^\top Y - I_d} < \delta$.
	
	Let $\widetilde Y$ be the output of Algorithm~\ref{alg:two_column} in $\ExactFinalize$.
	By Proposition~\ref{prop:shadow_state_survival}~(ii) and~(iii), Proposition~\ref{prop:two_column_correctness} applies to $Y$.
	By Proposition~\ref{prop:two_column_correctness}, we have $\tr(A^\top \widetilde Y)\ge \tr(A^\top Y)$.
	Moreover, $\widetilde Y^\top \widetilde Y$ is diagonal, with diagonal entries $(Y^\top Y)_{ii} + \sum_{k\ne i} |(Y^\top Y)_{ik}|$ for $i\in [d]$.
	Since $\fnorm{Y^\top Y-I_d}<\delta$, by the Cauchy--Schwarz inequality these entries are at most
	\begin{align*}
		(1+\delta) + \sum_{k\ne i} |(Y^\top Y)_{ik}| \le (1+\delta) + \sqrt{d-1} \delta.
	\end{align*}
	Let $\rho:=\sqrt{1 + (1 + \sqrt{d-1}) \delta}$, so that $\norm{\widetilde Y_i}\le\rho$ for every $i\in[d]$.
	After the normalization and the sign flips in $\ExactFinalize$, the returned matrix $X$ satisfies $A_i^\top X_i=\abs{A_i^\top\widetilde Y_i}/\norm{\widetilde Y_i}\ge A_i^\top\widetilde Y_i/\rho$ for every $i\in[d]$.
	Summing over $i$, we obtain
	\begin{align*}
		\tr(A^\top X) \ge \frac{\tr(A^\top \widetilde Y)}{\rho}
		\ge \frac{1}{\sqrt{1 + (1 + \sqrt{d-1}) \delta}} \cdot \left(1 - \frac{\delta}{4}\right) \optlin(A;s).
	\end{align*}
	Since $\LinearOracle$ returns the best matrix it finds, it remains to show that $1 - t \le (1 - \delta / 4) / \sqrt{1 + (1 + \sqrt{d-1}) \delta}$. 
	Since $(1 + x)^{-1/2}\ge 1 - x/2$ for every $x\ge 0$ and $\delta = t / (8(d+1))$, we obtain
	\begin{align*}
		\frac{1}{\sqrt{1 + (1 + \sqrt{d-1}) \delta}} \cdot \left(1 - \frac{\delta}{4}\right) 
		&\ge \left(1 - \frac{(1 + \sqrt{d-1}) \delta}{2}\right) \cdot \left(1 - \frac{\delta}{4}\right) \\
		& \ge (1 - d\delta) \cdot \left(1 - \frac{\delta}{4}\right)\\
		& \ge \left(1 - \frac{t}{8}\right) \left(1 - \frac{t}{32d}\right) \\
		& \ge 1 - \left(\frac{1}{8} + \frac{1}{32d}\right)t
		\ge 1 - t.
	\end{align*}
\end{proof}

Next, we prove the approximation guarantee of Algorithm~\ref{alg:main_alg}.

\approximationTheorem*

\begin{proof}[Proof of Theorem~\ref{thm:main_approximation}]
	If $\opt(R;s)=0$, every matrix in $\mf_s$ is optimal.
	Thus, we assume that $\opt(R;s)>0$.
	Let $X^*$ be an optimal solution of \ref{eqn:SPCA}, and let $W^*:= R^\top X^* / \fnorm{R^\top X^*}$, so that $\fnorm{W^*} = 1$.
	Let $J_W$ and $t = \epsilon / 4$ be as in Algorithm~\ref{alg:main_alg}, and let $m:=rd$. 
	Let $W\in J_W^{-1}\Z^{r\times d}$ be defined by
	\begin{align*}
		W_{ij} := \frac{\sign(W_{ij}^*) \lfloor J_W \abs{W^*_{ij}} \rfloor}{J_W},
		\quad i\in [r], \ j\in [d].
	\end{align*}
	Since $\abs{W_{ij}}\le\abs{W^*_{ij}}$ and $\abs{W_{ij}-W^*_{ij}}\le 1/J_W$, we have $\fnorm{W}\le 1$ and $\fnorm{W - W^*}\le \sqrt{m} / J_W \le t$.
	Hence Algorithm~\ref{alg:main_alg} enumerates $W$.
	By the Cauchy--Schwarz inequality,
	\begin{align*}
		\optlin(RW;s)	&\ge \tr((RW)^\top X^*)
		= \tr(W^\top R^\top X^*)\\
		&\ge \tr((W^*)^\top R^\top X^*) - \fnorm{W - W^*} \fnorm{R^\top X^*}\\
		&\ge (1-t) \sqrt{\opt(R;s)}.
	\end{align*}
	
	Let $X_W$ be the output of $\LinearOracle$ for the input $(RW,s,t)$.
	Since $\fnorm{W}\le1$, the Cauchy--Schwarz inequality and Proposition~\ref{prop:linear_oracle_guarantee} give
	\begin{align*}
		\fnorm{R^\top X_W} 
		&\ge	\tr((RW)^\top X_W) 
		\ge \left(1 - \frac{\epsilon}{4}\right) \optlin(RW;s) \\
		&\ge \left(1 - \frac{\epsilon}{4}\right)^2 \sqrt{\opt(R;s)},
	\end{align*}
	and thus
	\begin{align*}
		\fnorm{R^\top X_W}^2 
		\ge \left(1 - \frac{\epsilon}{4}\right)^4 \opt(R;s)
		\ge \left(1 - \epsilon\right) \opt(R;s),
	\end{align*}
	where the last inequality follows from $(1-x)^4 \ge 1-4x$ for $x\in [0, 1]$.
	Since Algorithm~\ref{alg:main_alg} returns the best matrix it finds, its output satisfies the same bound.
\end{proof}

\subsection{Bit complexity}
\label{app:complexity}
In this section, we prove Theorem~\ref{thm:main_alg_complexity}, which we restate below.
Real algebraic numbers are represented exactly, as described after Definition~\ref{def:field_ext}.

\mainAlgComplexityTheorem*

We first bound the running time of one call to $\LinearOracle$, and then multiply this bound by the number of matrices $W$ enumerated by Algorithm~\ref{alg:main_alg}.
We first count the arithmetic operations performed by the ADP and by $\LinearOracle$, and then we bound the bit complexity of these operations.

\subsubsection{Workload analysis.}

Let $q := \binom{d}{2}$, which is an upper bound on $|H|$.
Recall that $D_d = d(d-1)^2$ is an upper bound on the number of hard variables in one branch of $\LinearOracle$.
Let $q_G := \binom{d+1}{2}$ be the number of entries in the lower triangular part of a $d\times d$ Gram matrix, and let $B_{\rm in}$ be the encoding size of the input $(A, s, t)$, including the dimensions.

\paragraph{Number of choices of $H$ and $\ml$.}
There are at most $2^q$ sets $H$ of hard pairs.
By Lemma~\ref{lem:number_hard_row}, the number of hard rows satisfies $b\le b_d:=(d-1)q$.
For each $b$, there are at most $2^{d b_d}$ lists $\ml = \multiset{T^1, T^2, \ldots, T^b}$.
Therefore, the total number of choices in lines~7--11 of $\LinearOracle$ is at most
\begin{align*}
	E_d := (b_d + 1) 2^{q + d b_d}.
\end{align*}

\paragraph{Number of boxes $Q$.}
Since $a = \delta / (256(d+1)(D_d + 1))$ and $\delta = t/ (8(d+1))$ in Table~\ref{tab:parameters}, the number of boxes is at most 
\begin{align}
	\label{eqn:upper_bound_num_boxes}
	J_p^{D_d} = \left\lceil \frac{4096 (d+1)^2 (D_d + 1)}{t} \right\rceil^{D_d} \le 
	f_{\rm branches}(d) t^{-D_d}
\end{align}
for some function $f_{\rm branches}$ that depends only on $d$.
For every box, the subroutine $\FixHardVariables$ decides whether a quadratic system is feasible and, if so, computes one solution.
This system has at most $D_d$ variables and a number of constraints that depends only on $d$.
Hence the cost of one such call is a factor that depends only on $d$ times a fixed polynomial in the bit length of the coefficients (see Lemma~\ref{lem:hard_feas_complexity}).

\paragraph{Number of choices of free variables at a leaf.}
$\SelectBranch$ first initializes the leaves.
The running time of this step is governed by the number of choices of free variables at a leaf.
Recall that $M = \lceil 256 d(d+1) / \delta \rceil$, and $J_0 = \lceil \log_2(256 d n (d+1)/\delta) \rceil = \mo(\log(n+1) + \log(d + 1) + \log(1/t))$. 
Let
\begin{align*}
	U:=1 + \log_2(n+1) + \log_2(d + 1) + \log_2(1/t).
\end{align*}
By the definition of $\mv$, we have
\begin{align*}
	\abs{\mv} \le 2MJ_0 + 3 \le \frac{4097d(d+1)^2}{t} J_0 + 3 \le f_{\rm free}(d) \cdot \frac{U}{t}
\end{align*}
for some function $f_{\rm free}$.
At a leaf of the ADP tree, there are at most $2^d$ ordinary row types and at most $b_d$ hard row types in $\ml$.
The row is either unused, or it has one of these row types and at most $d$ free variables with values in $\mathcal V$.
Hence the number of leaf candidates is at most
\begin{equation}
	\label{eqn:ub_leaf}
	L_{\rm leaf} \le 1 + (2^d + b_d) \abs{\mv}^d  \le f_{\rm leaf}(d)t^{-d} U^d
\end{equation}
for some function $f_{\rm leaf}$.

Next, we bound the number of states at the internal nodes.

\paragraph{Number of capacity caps.}
Recall that $K_0 = \lceil 16(P_d+1)/\tau \rceil$ and $\alpha = \tau / (8(H_{\rm tree}+1))$ in Table~\ref{tab:parameters}, with $P_d = dq$, $\tau = \delta / (128 d^2)$, and $H_{\rm tree} = \lceil \log_2(n) \rceil$.
Since $\log(1+x) \ge x / 2$ for $x\in [0,1]$, we have
\begin{align*}
	\log_{1+\alpha}(n) 
	\le \frac{\log(n+1)}{\log(1 +\alpha)} 
	\le \frac{2\log(n+1)}{\alpha}
	\le \frac{16384d^2(d+1) (H_{\rm tree}+1) U}{t}.
\end{align*}
Since $U_c$ consists of the integers $0,\ldots,K_0$ and a geometric sequence above $K_0$, we obtain
\begin{align*}
	\abs{U_c} \le K_0 + 1 + \log_{1 + \alpha}(n) + 1
	\le \frac{16385(P_d + 1 + (H_{\rm tree}+1)U)d^2 (d+1)}{t} + 2.
\end{align*}
Therefore, the number of vectors $\hat c$ of capacity caps is at most
\begin{align}
	\label{eqn:ub_sparsity}
	\abs{U_c}^d \le f_{\rm sparsity}(d) t^{-d} U^{2d}
\end{align} 
for some function $f_{\rm sparsity}(d)$.

\paragraph{Number of Gram matrix bins.}
Since Gram matrices are symmetric, a Gram matrix bin label only needs to store the lower triangular part of the rounded Gram matrix.
Recall that $\theta = \delta / (128 d^2 (H_{\rm tree}+1)) < 1$ and $\kappa_G = \delta / (128n (H_{\rm tree}+1))$. 
We first count the number $L_G$ of intervals $I_k$ with $k\ge 0$ that contain a trace smaller than $4d$.
We have
\begin{align*}
	L_G \le 1 + \frac{\log(4d) - \log(\kappa_G)}{\log(1+\theta)}
	\le 1 + \frac{2\log\left(\frac{4d}{\kappa_G}\right)}{\theta}.
\end{align*}
We claim that, within each trace interval $[(1+\theta)^k\kappa_G, (1+\theta)^{k+1}\kappa_G)$, each lower triangular entry has at most $7/\theta$ possible bins.
Indeed, since $G\succeq 0$, we have $\abs{G_{il}} \le \tr(G) < (1+\theta)^{k+1}\kappa_G$, and thus the number of possible bins is at most 
\begin{align*}
	2\left\lceil \frac{(1+\theta)^{k+1}\kappa_G}{\theta (1+\theta)^k \kappa_G} \right\rceil + 1
	&= 2\left\lceil \frac{1+\theta}{\theta} \right\rceil + 1
	\le 2 \left(\frac{1}{\theta} + 1 + 1\right) + 1\\
	&\le \frac{7}{\theta}
	= \frac{7168 (d+1) d^2 (H_{\rm tree}+1)}{t}.
\end{align*}
Therefore, the total number $M_G$ of Gram matrix bin labels satisfies
\begin{equation}
	\label{eqn:ub_gram_bins}
	M_G \le L_G \cdot \left(\frac{7}{\theta}\right)^{q_G} + 1
	\le f_{\rm bin}(d) t^{-q_G - 1} U^{q_G + 2}
\end{equation}
for some function $f_{\rm bin}$.

\paragraph{Number of states at one internal node.}

Recall that a dictionary key consists of a set $\locc$, a vector of capped counters $(\mu_e)_{e\in S}$, a vector $\hat c$ of capacity caps, and a Gram matrix bin label $\mathrm{lab}(G)$.
There are at most $2^b$ possible sets $\locc$ and at most $(d+1)^q$ possible vectors $(\mu_e)_{e\in S}$.
Moreover, there are at most $|U_c|^d$ possible vectors $\hat c$ and at most $M_G$ possible labels $\mathrm{lab}(G)$.
Thus, the number of states at one node is at most
\begin{equation}
	\label{eqn:number_ADP_states}
	K_{\rm state} \le 2^b(d+1)^q \abs{U_c}^d M_G.
\end{equation}
Since $b\le(d-1)q$, the first two factors depend only on $d$.
Substituting \eqref{eqn:ub_sparsity} and \eqref{eqn:ub_gram_bins} gives
\begin{align*}
	K_{\rm state} \le f_{\rm state}(d) t^{-(q_G + d + 1)} U^{q_G + 2 d + 2}
\end{align*}
for some function $f_{\rm state}$.

\paragraph{Number of state pairs at a merge.}
When two child nodes are merged, the number $K_{\rm merge}$ of state pairs satisfies
\begin{align}
	\label{eqn:merge_complexity}
	K_{\rm merge} \le K_{\rm state}^2 \le f_{\rm state}^2(d) t^{-2(q_G + d + 1)} U^{2q_G + 4 d + 4}.
\end{align}

\paragraph{ADP workload for a fixed $p$.}
Fix $H$, $b$, $\ml$, and a vector $p$ of values of the hard variables in $\LinearOracle$.
The ADP tree has $N_{\rm leaf}<2n$ leaves and $N_{\rm leaf}-1$ internal nodes.
At each leaf, $\LeafInit$ enumerates at most $L_{\rm leaf}$ candidates and processes each of them with $\KeepBest$.
Hence the leaves require at most $N_{\rm leaf} L_{\rm leaf}$ processing steps.
At each internal node, $\SelectBranch$ processes every pair of states of its two children, that is, at most $K_{\rm merge}$ pairs.
Hence the internal nodes require at most $(N_{\rm leaf}-1)K_{\rm merge}$ processing steps.
Finally, each of the at most $K_{\rm state}$ root states is checked against the acceptance conditions.
Therefore, for a fixed $p$, the total ADP workload is
\begin{align}
	\label{ub:branch_workload}
	\begin{split}
		W_{\rm ADP} &\le 1 + N_{\rm leaf} L_{\rm leaf} + (N_{\rm leaf}-1) K_{\rm merge} + K_{\rm state}\\
		&\le f_{\rm bw}(d) n t^{-(d^2+3d+2)} U^{d^2+5d+4}
	\end{split}
\end{align}
for some function $f_{\rm bw}$, where we used $2q_G=d^2+d$.

\paragraph{$\LinearOracle$ workload.}
For every box $Q$, $\FixHardVariables$ solves one quadratic system exactly and, if the system is feasible, returns one vector $p$.  
Therefore,
\begin{align*}
	W_{\LinearOracle} &\le E_d J_p^{D_d} (1 + N_{\rm leaf} L_{\rm leaf} + (N_{\rm leaf}-1) K_{\rm merge} + (4nd+1) K_{\rm state}),
\end{align*}
where the leading $1$ accounts for the call to $\FixHardVariables$ for the box.
The term $4nd K_{\rm state}$ is an upper bound that accounts, for each accepted root state, for reconstructing its candidate solution by backtracking to the $n$ leaves and for $\ExactFinalize$.
Substituting the estimates of $E_d$ and $J_p^{D_d}$, and absorbing all factors that depend only on $d$ into a function $f_{\rm lin}(d)$, we obtain
\begin{equation}
	\label{eqn:linear_oracle_raw_work}
	W_{\LinearOracle} \le f_{\rm lin}(d) n t^{-D_d-d^2-3d-2} U^{d^2+5d+4}.
\end{equation}

Next, we bound the bit complexity of each subroutine, and then of Algorithm~\ref{alg:main_alg}.

\subsubsection{Bit complexity analysis.}

To bound the bit complexity of Algorithm~\ref{alg:main_alg}, we need to address two issues.
The first is how to represent irrational algebraic numbers.
The second is the bit complexity of arithmetic operations on these representations.
Irrational numbers arise only in $\ExactFinalize$ and $\FixHardVariables$, which solve \eqref{eqn:rotation_PCA} and \eqref{eqn:hard_constraints}, respectively.

A standard way to represent a real root of a polynomial with integer coefficients is to store the polynomial together with an interval that contains exactly one of its roots (see, e.g.,~\cite[Section~2.2.2]{Bas14}).
For example, the polynomial $x^2 - 2$ and the \emph{root isolating interval} $(1,2)$ specify $\sqrt{2}$ exactly, because $\sqrt{2}$ is the only root of the polynomial in this interval.
To store the number $2 + 3\sqrt{2}$, we store $\sqrt{2}$ as above, together with its \emph{rational coordinate coefficients} $2$ and $3$.
We use the following standard notions from field theory, see for example~\cite[Chapter~V]{Lan02}.

\begin{definition}
	\label{def:field_ext}
	Let $\gamma\in\R$ be a root of a non-zero polynomial with integer coefficients. 
	We denote by $\mk:=\Q(\gamma)$ the smallest subfield of $\R$ that contains $\Q$ and $\gamma$.
	In other words, $\mk$ is the set of numbers obtained from rational numbers and $\gamma$ by addition, subtraction, multiplication, and division by non-zero numbers.
	Let $f\in\Q[x]$ be an irreducible polynomial with $f(\gamma) = 0$.
	We call $f$ a \emph{defining polynomial} of $\mk$.
	The \emph{degree} $[\mk:\Q]$ of the field extension is the degree of $f$.
\end{definition}

We represent $\mk$ by a defining polynomial $f$ and a rational interval that contains $\gamma$ and no other root of $f$, and we represent each element of $\mk$ by its rational coordinate coefficients.
For $m:= [\mk : \Q]$, each element of $\mk$ is stored as $\sum_{j = 0}^{m - 1} q_j \gamma^j$ with $q_j\in \Q$.
The \emph{representation size} of a finite set of elements of $\mk$ is the largest encoding size of the coefficients of $f$, of the endpoints of the interval, and of the rational coordinate coefficients of these elements.
We first bound the bit complexity of computations that satisfy the following assumption.

\begin{assumption}
	\label{ass:bit_complexity}
	All input entries belong to one field $\mk = \Q(\gamma) \subseteq \R$ with $[\mk:\Q]\le \Delta$.
	Let $B_{\rm enc}\ge 1$ be an upper bound on the representation size of the input entries.
	Every divisor is non-zero, and every square root is the non-negative root of a non-negative number. 
	The computation has at most $N_0$ input entries and performs at most $L$ additions, subtractions, multiplications, divisions, and equality or sign tests, and at most $\kappa$ square-root computations.
\end{assumption}

For example, the computation $x_1 x_2 + x_3$ with $x_1, x_2, x_3\in \mk$ has $N_0 = 3$, $L = 2$, and $\kappa = 0$.

\begin{lemma}
	\label{lem:bit_complexity}
	Consider a computation that satisfies Assumption~\ref{ass:bit_complexity} and keeps every scalar result.
	Let $T$ be its bit complexity, let $\mk_{\rm out}$ be a field that contains all scalar results, and let $B_{\rm out}$ be the representation size of the scalar results in $\mk_{\rm out}$.
	Then $\mk_{\rm out}$ can be chosen so that
	\begin{align*}
		[\mk_{\rm out} : \Q] &\le \Delta 2^{\kappa}, \quad
		T \le F(\Delta, N_0, L, \kappa) (B_{\rm enc} + 1)^c, \\
		B_{\rm out} &\le C(\Delta, N_0, L, \kappa) (B_{\rm enc}+1),
	\end{align*}
	for some parameter-dependent constants $F$ and $C$, and for an absolute constant $c > 0$.
\end{lemma}

\begin{proof}
	The computation consists of steps, each of which is an arithmetic operation, a comparison, or a square root.
	Before each step, let $\mk$ be the current field, which contains all kept values.
	Let $m:=[\mk:\Q]$, let $u$ be the number of kept values, and let $b_{\rm cur}$ be the representation size of the kept values.
	We first show that the bit complexity $T_{\rm step}$ of each step and the largest encoding size $b_{\rm new}$ after the step satisfy
	\begin{align}
		\label{eqn:step_complexity}
		T_{\rm step} \le F_{\rm step}(m, u) (b_{\rm cur} + 1)^c, \quad
		b_{\rm new} \le C_{\rm step}(m, u) (b_{\rm cur} + 1),
	\end{align}
	for some parameter-dependent constants $F_{\rm step}$ and $C_{\rm step}$, and for an absolute constant $c>0$.
	\begin{enumerate}
		\item[(i)] 
		\textit{Arithmetic and comparison in $\mk$.} 
		By~\cite[Section~2.2]{Ura24}, arithmetic operations and comparisons in $\mk$ have bit complexity $C(m)(b_{\rm cur} + 1)^c$, and the encoding sizes of the results are at most $C(m) (b_{\rm cur} + 1)$.
		An equality test has the same bounds, since it compares the coordinate coefficients reduced modulo the defining polynomial.
		
		\item[(ii)]
		\textit{Field extension after adjoining a root.} 
		If the square root of an element of $\mk$ does not belong to $\mk$, we extend $\mk$.
		The multiplication table and univariate representation procedures in \cite[Sections~12.2 and~12.4]{BasPolRoy06} compute a primitive element of the new field, of degree at most $2m$.
		If necessary, the factorization algorithm of Lenstra, Lenstra, and Lov{\'a}sz~\cite{LenLenLov82} computes its irreducible defining polynomial.
		The real-root isolation and sign determination procedures in \cite[Sections~10.2 and~10.3]{BasPolRoy06} identify the intended real root.
		The univariate representation then expresses the $u$ kept values in the new field.
		The bit complexity of these procedures is $F(m,u)(b_{\rm cur}+1)^c$, and the encoding sizes of the results are at most $C(m,u)(b_{\rm cur}+1)$.
	\end{enumerate}
	
	We now combine the steps.
	Arithmetic operations and comparisons stay in the current field, and adjoining a square root multiplies the degree of the field extension by at most two.
	Hence $[\mk_{\rm out} : \Q] \le \Delta 2^{\kappa}$.
	Let $J:=L + \kappa$, $m_{\max}:=\Delta 2^{\kappa}$, and $u_{\max}:= \max\{1, N_0 + J\}$.
	Then every step has $m\le m_{\max}$ and $u\le u_{\max}$.
	Let $F_{\max}$ be the maximum of $F_{\rm step}(m, u)$ over $m\in [m_{\max}]$ and $u\in [u_{\max}]$, and let $C_{\max}$ be one plus the maximum of $C_{\rm step}(m, u)$ over the same range.
	Let $b_j$ be the largest encoding size after $j$ steps.
	Then \eqref{eqn:step_complexity} gives
	\begin{align*}
		b_{j+1} + 1\le C_{\max} (b_j + 1) \le \cdots \le C_{\max}^{j + 1} (b_0 + 1) \le C_{\max}^{j + 1} (B_{\rm enc} + 1).
	\end{align*}
	Since there are at most $J$ steps, \eqref{eqn:step_complexity} gives the statement with $C(\Delta, N_0, L, \kappa):= C_{\max}^J$ and $F(\Delta, N_0, L, \kappa):= J F_{\max} C_{\max}^{cJ}$.
\end{proof}

Next, we analyze the running time of $\ExactFinalize$.
We start with the computation of $N$ in Algorithm~\ref{alg:two_column}.

\begin{lemma}
	\label{lem:runtime_N}
	Let $d\ge 2$, let $\mk\subseteq\R$ be an algebraic number field, and let $B\in \mk^{d\times (d-2)}$ be as in Algorithm~\ref{alg:two_column}.
	One can construct $N\in \R^{d\times 2}$ such that $N^\top N = I_2$ and $B^\top N = O$, with entries in an extension of $\mk$ obtained by adjoining at most two square roots, using $\mo(d^3)$ field operations and comparisons.
\end{lemma} 

\begin{proof}
	If $d = 2$, we set $N := I_2$.
	Assume that $d > 2$.
	In the proof of Proposition~\ref{prop:two_column_correctness}, we showed that $\ker(B^\top)$ has dimension at least $2$.
	Gaussian elimination computes two linearly independent vectors $w_1, w_2\in\ker(B^\top)\cap\mk^d$ with $\mo(d^3)$ field operations.
	Let $\hat w_2:= w_2 - (w_1^\top w_2 / w_1^\top w_1) w_1$, so that $\hat w_2\ne0$ and $\hat w_2^\top w_1 = 0$.
	Let $N_1 := w_1 / \sqrt{w_1^\top w_1}$, $N_2:= \hat w_2 / \sqrt{\hat w_2^\top \hat w_2}$, and $N := (N_1, N_2)$.
	Then $N^\top N = I_2$ and $B^\top N = O$, and the construction uses two square roots.
\end{proof}

Next, we consider the computation of $K_{\rm new}$.

\begin{lemma}
	\label{lem:runtime_K}
	Let $\mk\subseteq\R$ be an algebraic number field, and let $\Gamma, \Lambda\in \mk^{2\times 2}$ with $\Lambda\succeq 0$.
	An exact optimal solution to \eqref{eqn:rotation_PCA} can be computed using $\mo(1)$ field operations and comparisons and at most three square roots.
\end{lemma}

\begin{proof}
	In the proof of Proposition~\ref{prop:rotation_PCA}, we showed that every feasible solution of \eqref{eqn:rotation_PCA} has the form $\Psi \sqrt{\Lambda}$, where $\Psi$ is a $2\times 2$ orthogonal matrix.
	Hence it suffices to compute $\sqrt{\Lambda}$ and an optimal $\Psi$.
	If $\Lambda = O$, we set $K_{\rm new} := O$.
	Assume that $\Lambda\ne O$.
	Let $a:= \sqrt{\det(\Lambda)}$, $b:=\sqrt{\tr(\Lambda) + 2a} > 0$, and $P:= (\Lambda + a I_2) / b$.
	By the Cayley--Hamilton theorem, $\Lambda^2  - \tr(\Lambda) \Lambda + \det(\Lambda) I_2 = O$, and therefore
	\begin{align*}
		P^2 = \frac{\Lambda^2 + 2a\Lambda + a^2 I_2}{b^2} = \frac{\tr(\Lambda) \Lambda + 2a \Lambda}{b^2} = \Lambda.
	\end{align*}
	Since $a \ge 0$, we have $P\succeq 0$, and thus $\sqrt{\Lambda} = P$.
	
	Next, we determine $\Psi$.
	Every $2\times 2$ orthogonal matrix has one of the two forms $\Psi_+ = \begin{pmatrix}
		x & -y \\
		y & x
	\end{pmatrix}$ or 
	$\Psi_- = \begin{pmatrix}
		x & y \\
		y & -x
	\end{pmatrix}$, where $x^2+y^2=1$.
	We choose $\Psi$ to maximize $\tr(\Gamma^\top \Psi\sqrt{\Lambda}) = \tr(\sqrt{\Lambda}\Gamma^\top \Psi)$. 
	Write $M:=\sqrt{\Lambda}\Gamma^\top = (m_{ij})$, and define 
	\begin{align*}
		v_+ := \begin{pmatrix}
			m_{11} + m_{22}\\
			m_{12} - m_{21}
		\end{pmatrix}, \quad
		v_- := \begin{pmatrix}
			m_{11} - m_{22}\\
			m_{12} + m_{21}
		\end{pmatrix}.
	\end{align*}
	Since $\tr(M\Psi_+)=(x,y)v_+$ and $\tr(M\Psi_-)=(x,y)v_-$, by the Cauchy--Schwarz inequality the maximum of \eqref{eqn:rotation_PCA} is $\max\{\norm{v_+},\norm{v_-}\}$.
	If $\norm{v_+}^2 = \norm{v_-}^2 = 0$, we take $\Psi = I_2$.
	Otherwise, assume without loss of generality that $\norm{v_+}^2 \ge \norm{v_-}^2$.
	Then we set $\Psi=\Psi_+$ with $(x, y)^\top = v_+ / \norm{v_+}$.
	
	Finally, we set $K_{\rm new}:=\Psi\sqrt\Lambda$.
	The computation uses $\mo(1)$ field operations and comparisons, and at most three square roots.
\end{proof}

We now bound the degree of the field extension and the encoding sizes of the scalar results of $\ExactFinalize$.
\begin{lemma}
	\label{lem:exact_finalize_encoding}
	Consider the computation $\ExactFinalize$, with the notation of Assumption~\ref{ass:bit_complexity}. 
	Let $Y^0:=Y$ be the input candidate solution and $G^0:=Y^\top Y$.
	Let $B_{\rm tot-enc} \ge 1$ be an upper bound on the total encoding size of the coefficients of the defining polynomial of $\mk$, of the endpoints of its root isolating interval, and of the rational coordinate coefficients of the entries of $A$, $Y^0$, and $G^0$.
	Let $\mk_{\rm out}$ be a field that contains all scalar results of $\ExactFinalize$, and let $B_{\rm out}$ be the representation size of these scalar results in $\mk_{\rm out}$.
	Then $\mk_{\rm out}$ can be chosen so that
	\begin{align*}
		[\mk_{\rm out}: \Q]\le \Delta 2^{5\binom{d}{2} + d}, \quad
		B_{\rm out} \le C_{\ExactFinalize}(d, \Delta) (B_{\rm tot-enc} + 1),
	\end{align*}
	for some parameter-dependent constant $C_{\ExactFinalize}$.
\end{lemma}

\begin{proof}
	We first bound $[\mk_{\rm out}: \Q]$.
	By Lemmas~\ref{lem:runtime_N} and~\ref{lem:runtime_K}, each pair $e\in S$ in Algorithm~\ref{alg:two_column} uses at most five square roots. 
	The normalization of the columns uses $d$ further square roots.
	Hence Lemma~\ref{lem:bit_complexity} gives $[\mk_{\rm out}: \Q]\le \Delta 2^{5\binom{d}{2} + d}$.
	
	Every scalar value produced by the local computations of $\ExactFinalize$ is obtained through arithmetic operations or taking square roots. 
	In particular, each iteration of Algorithm~\ref{alg:two_column} involves only a number of scalar inputs, arithmetic operations, and square roots bounded in terms of $d$.
	We can defer the field representation update of any stored value until it is needed for a computation, using the embeddings between successive fields.
	Since the number of iterations is $\mo(d^2)$, Lemma~\ref{lem:bit_complexity} gives encoding sizes at most $C(d,\Delta)(B_{\rm tot-enc}+1)$ for all locally computed values in Algorithm~\ref{alg:two_column}.
	After the local updates, the representations of the remaining entries are updated into the final field $\mk_{\rm out}$.
	As each such update involves only a number of operations depending on $d$ and $\Delta$, the same encoding-size bound holds for all output entries, with a constant that does not depend on $n$.
	
	Finally, since $B_{\rm tot-enc}$ is the total encoding size of the inputs, the encoding sizes of $\norm{\widetilde Y_i} = \sqrt{\onorm{G_i^0}}$ and $A_i^\top \widetilde Y_i$ are at most the desired upper bound.
    Here $\widetilde Y$ denotes the matrix after Algorithm~\ref{alg:two_column} and before the normalization.
	Indeed, since $G_i^0$ has $d$ entries, Lemma~\ref{lem:bit_complexity} bounds the encoding size of this term by $C_{\rm 2-norm}(d, \Delta)(B_{\rm tot-enc} + 1)$ for some parameter-dependent constant $C_{\rm 2-norm}$.
    It remains to upper bound the encoding size of $A_i^\top \widetilde Y_i$.
    Let $I_i$ denote the set of rows in column $i$ modified by Algorithm~\ref{alg:two_column}.
	Then $|I_i|\le d(d-1)$, and
	\begin{align*}
		A_i^\top \widetilde Y_i = A_i^\top Y_i^0 + \sum_{j\in I_i} A_{ji} (\widetilde Y_{ji} - Y_{ji}^0).
	\end{align*}
	Since $B_{\rm tot-enc}$ is the total input encoding size, computing the first term on the right-hand side by clearing denominators and adding its at most $n$ terms gives an encoding size of at most $C_{\rm sum}(\Delta)(B_{\rm tot-enc}+1)$, for some constant $C_{\rm sum}$ that depends only on $\Delta$. 
	The sum on the right-hand side has only $\mo(d^2)$ terms, whose encoding sizes are bounded by Lemma~\ref{lem:bit_complexity} and the analysis above. 
	The normalization and the sign tests satisfy the same bound.
\end{proof}

\begin{proposition}
	\label{prop:exact_finalize_complexity}
	Let $B_{\rm tot-enc}$ be as in Lemma~\ref{lem:exact_finalize_encoding}.
	Then there is an implementation of $\ExactFinalize$ whose bit complexity $T_{\ExactFinalize}$ satisfies 
	\begin{align*}
		T_{\ExactFinalize} \le n F_{\ExactFinalize}(d, \Delta)(B_{\rm tot-enc} + 1)^c
	\end{align*}
	for some parameter-dependent constant $F_{\ExactFinalize}$ and an absolute constant $c$.
\end{proposition}

\begin{proof}
	By Lemmas~\ref{lem:runtime_N} and~\ref{lem:runtime_K}, the iterations of Algorithm~\ref{alg:two_column} use $\mo(d^5)$ field operations and comparisons and at most $5\binom{d}{2}$ square roots.
	Therefore, by Lemma~\ref{lem:bit_complexity}, their bit complexity $T_{\rm iterations}$ satisfies
	\begin{align*}
		T_{\rm iterations} \le F(d,\Delta) (B_{\rm tot-enc} + 1)^c.
	\end{align*}
	
	The remaining computations of $\ExactFinalize$ require $\mo(nd^2)$ scalar operations and updates of field representations.
	By Lemma~\ref{lem:exact_finalize_encoding}, their inputs have encoding sizes at most $C_{\ExactFinalize}(d,\Delta) (B_{\rm tot-enc} + 1)$, and the degree of the field extension is at most $\Delta 2^{5\binom{d}{2} + d}$.
	As in the proof of Lemma~\ref{lem:exact_finalize_encoding}, the partial sums of the $n$-term sums in these computations have encoding sizes at most $C(d,\Delta)(B_{\rm tot-enc}+1)$.
	Hence, by Lemma~\ref{lem:bit_complexity} applied to each addition, multiplication, and field update separately, each of these scalar computations has bit complexity $F(d,\Delta) (B_{\rm tot-enc} + 1)^c$.
	Absorbing $d^2$ into $F(d,\Delta)$, the remaining computations have bit complexity $n F(d,\Delta) (B_{\rm tot-enc} + 1)^c$.
	Adding the two bounds concludes the proof.
\end{proof}

Next, we bound the bit complexity of $\FixHardVariables$. 
The main step is to solve the system \eqref{eqn:hard_constraints} in each box $Q$.
We reduce this problem to finding a zero of a non-negative multivariate quartic polynomial, and then apply a bounded algebraic sampling algorithm.

\begin{proposition}
	\label{prop:bounded_algebraic_sampling}
	Let $k\ge 1$, and let $p\in \Z[x_1, x_2, \ldots, x_k]$ be a polynomial of degree at most $\nu$ with coefficients of encoding size at most $B_{\rm coef}\ge 1$.
	Assume that $p\ge 0$ on $\R^k$ and that $Z(p):= \{x\in \R^k: p(x) = 0\}$ is bounded.
	Then there is an algorithm that returns exact points of $Z(p)$ that meet every connected component of $Z(p)$.
	In particular, it returns no point if $Z(p) = \emptyset$.
	Let $T_{\rm samp}$ be its bit complexity, let $\Delta_{\rm samp}$ be the largest degree of the field extension of a returned point, and let $B_{\rm samp}$ be the largest representation size of a returned point.
	Then
	\begin{align*}
		T_{\rm samp} &\le F_{\rm samp}(k, \nu) (B_{\rm coef} + 1)^c, \quad
		\Delta_{\rm samp} \le (\nu + 2) (\nu + 1)^{k-1}, \\
		B_{\rm samp} &\le C_{\rm samp}(k, \nu) (B_{\rm coef} + 1),
	\end{align*}
	for some parameter-dependent constants $F_{\rm samp}$ and $C_{\rm samp}$, and an absolute constant $c$.
\end{proposition}

\begin{proof}
	We apply the bounded algebraic sampling algorithm \cite[Algorithm~12.15]{BasPolRoy06}, as recalled in \cite[Algorithm~2.8]{BasLerRoy09}.
	This algorithm deforms $p$ using an infinitesimal, computes finitely many critical points of the deformed polynomial, and takes their limits to obtain points of $Z(p)$. 
	These points meet every bounded connected component of $Z(p)$. 
	Since $Z(p)$ is bounded, they meet every connected component. 
	In particular, the returned list is empty when $Z(p)=\emptyset$.
	
	Each returned point is specified by a real univariate representation
	\begin{align*}
		(f,g_0,g_1,\ldots,g_k,\pi),
	\end{align*}
	where $\pi$ is a \emph{Thom encoding} that identifies a real root $\gamma$ of $f$, and the point is
	\begin{align*}
		x=\left(
		\frac{g_1(\gamma)}{g_0(\gamma)},\ldots,
		\frac{g_k(\gamma)}{g_0(\gamma)}
		\right),\qquad g_0(\gamma)\neq 0.
	\end{align*}
	
	We next bound the sizes of these representations.
	Let $N:= (\nu +2) (\nu + 1)^{k-1}$.
	By the analysis of \cite[Algorithm~2.8]{BasLerRoy09}, the degrees of $f,g_0,\ldots,g_k$ are at most $N$, and the bitsizes of their coefficients are bounded by a function of $k$ and $\nu$ that is linear in $B_{\rm coef}$.
	In particular, these encoding sizes are at most $C_0(k,\nu)(B_{\rm coef}+1)$ for some parameter-dependent constant $C_0$.
	By the complexity analysis of this algorithm in \cite[Chapter~12]{BasPolRoy06}, the number of arithmetic operations depends only on $k$ and $\nu$, and these operations are performed on integers whose bitsizes are bounded in the same way.
	Hence the algorithm has bit complexity at most $F_1(k,\nu)(B_{\rm coef}+1)^{c_1}$, for some parameter-dependent constant $F_1$ and an absolute constant $c_1$.
	
	Finally, we convert the real univariate representations to the representation of Definition~\ref{def:field_ext}.
	For the tuple $(f,g_0,\ldots,g_k,\pi)$, let $\gamma$ be the real root of $f$ specified by $\pi$.
	By polynomial factorization over $\Q$ \cite{LenLenLov82}, real-root isolation, and sign determination \cite[Section~10.2]{BasPolRoy06}, we can compute the primitive irreducible factor $m\in\Z[x]$ of $f$ such that $m(\gamma) = 0$, and a rational interval isolating $\gamma$.
	
	Since $g_0(\gamma)\ne0$ and $m$ is irreducible, we have $\gcd(g_0,m)=1$. 
	The extended Euclidean algorithm computes $u,v\in\Q[x]$ with $ug_0 + vm =1$.
	Hence $g_i = ug_0 g_i + vm g_i$, and so $g_i(\gamma)/g_0(\gamma)=u(\gamma)g_i(\gamma)$.
	Consequently, each coordinate $x_i=g_i(\gamma)/g_0(\gamma)$ equals $r_i(\gamma)$, where $r_i=(ug_i)\bmod m$ and $\deg r_i<\deg m$.
	Hence all coordinates belong to the same selected real field $\mk =\mathbb Q(\gamma)$, with $[\mk :\Q]=\deg m\le N$.
	
	Polynomial factorization over $\Q$~\cite{LenLenLov82}, real-root isolation~\cite[Section~10.2]{BasPolRoy06}, and the extended Euclidean algorithm over $\Q[x]$~\cite[Theorem~6.52]{vzGGer13} all have bit complexity polynomial in the degrees and in the coefficient encoding sizes.
	Since all degrees are $\mo(N)$, the resulting encoding sizes of the coefficients and of the root isolating interval are at most $C(k,\nu)(B_{\rm coef} + 1)$, and the total bit complexity of these steps is at most $F(k,\nu)(B_{\rm coef} + 1)^c$, for an absolute constant $c$.
	Enlarging the constants concludes the proof.
\end{proof}

In the next lemma, we bound the bit complexity of solving \eqref{eqn:hard_constraints} in one box.
Recall from Table~\ref{tab:parameters} that $J_p = \lceil 4096 (d+1)^2 (D_d + 1) / t \rceil$, and let $B_{J_p}:= 1 + \lceil \log_2(J_p + 1) \rceil$ be its encoding size.

\begin{lemma}
	\label{lem:hard_feas_complexity}
	For each box $Q$, the system \eqref{eqn:hard_constraints} with $p\in Q$ can be solved exactly in at most $F_{\rm hard}(d) B_{J_p}^c$ bit operations, for some parameter-dependent constant $F_{\rm hard}$ and an absolute constant $c$.
\end{lemma}

\begin{proof}
	We denote the variables in \eqref{eqn:hard_constraints} by $p_1, p_2, \ldots, p_D$.
	We write each orthogonality constraint in \eqref{eqn:hard_constraints} as $g_e(p) = 0$ for $e\in H$, and the norm constraint as $\sum_{j = 1}^D p_j^2 \le d$.
	For a box $Q = \Omega_{\ell_1} \times \cdots \times \Omega_{\ell_D}$, the constraint $p\in Q$ can be written as
	\begin{align*}
		p_j \ge \frac{2(\ell_j - 1)}{J_p} - 1 =: l_j, \quad
		p_j \le \frac{2\ell_j}{J_p} - 1 =: u_j, \quad \forall j \in [D].
	\end{align*}
	
	Solving \eqref{eqn:hard_constraints} in $Q$ is equivalent to finding a zero of the following non-negative quartic polynomial in $3D+1$ variables:
	\begin{align*}
		h(p, \alpha, \beta, \eta)&:= \sum_{e\in H} g_e^2(p) + \left(d - \eta^2 - \sum_{j = 1}^D p_j^2\right)^2 \\
		&\qquad + J_p^2 \sum_{j = 1}^D \left[(\alpha_j^2 + l_j - p_j)^2 + (\beta_j^2 + p_j - u_j)^2\right].
	\end{align*}
	Indeed, a vector $p$ can be extended to a zero $(p,\alpha,\beta,\eta)$ of $h$ if and only if $g_e(p) = 0$ for all $e\in H$, $\sum_{j = 1}^D p_j^2 \le d$, and $p_j\in [l_j, u_j]$ for all $j\in [D]$.
	Moreover, the zero set of $h$ is bounded.
	Indeed, if $h=0$, then $p\in Q$ and $\eta^2 \le d$.
	Moreover, $\alpha_j^2 = p_j - l_j$ and $\beta_j^2 = u_j - p_j$, so $\alpha_j^2 + \beta_j^2 = u_j - l_j = 2 / J_p$.
	
	Since $J_p l_j$ and $J_p u_j$ are integers and $|l_j|,|u_j|\le 1$, the polynomial $h$ has integer coefficients of absolute value at most $d^2+4(D+1)J_p^2$, and its number of monomials depends only on $d$.
	Thus, the total encoding size of the coefficients is at most $C_h(d) B_{J_p}$, for some parameter-dependent constant $C_h$.
	Proposition~\ref{prop:bounded_algebraic_sampling} with $\nu = 4$ and $k = 3D+1$ concludes the proof.
\end{proof}

We now bound the bit complexity of $\FixHardVariables$.

\begin{proposition}
	\label{prop:fix_hard_variables_complexity}
	Let $B_t$ be the encoding size of $t\in \Q\cap (0, 1)$.
	There is an implementation of $\FixHardVariables$ whose bit complexity $T_{\FixHardVariables}$ satisfies
	\begin{align*}
		T_{\FixHardVariables} \le F_{\FixHardVariables}(d) \left[(B_t + 1)^c + J_p^{D_d} B_{J_p}^c \right]
	\end{align*}
	for some parameter-dependent constant $F_{\FixHardVariables}$ and an absolute constant $c$.
	Moreover, each returned solution $p$ belongs to a field $\mk_p$ such that
	\begin{align*}
		[\mk_p : \Q] \le 6 \cdot 5^{3D}, \quad
		B_p \le C_p(d) B_{J_p},
	\end{align*}
	where $B_p$ is the representation size of $p$ in $\mk_p$, and $C_p$ is a parameter-dependent constant.
\end{proposition}

\begin{proof}
	Computing $J_p$ has bit complexity $F_1(d) (B_t + 1)^c$, for some parameter-dependent constant $F_1$ and an absolute constant $c$.
	By \eqref{eqn:upper_bound_num_boxes}, the number of boxes is at most $J_p^{D_d}$, since $D\le D_d = d(d-1)^2$.
	Together with Lemma~\ref{lem:hard_feas_complexity}, this gives the bound on $T_{\FixHardVariables}$.
	The bounds on $[\mk_p : \Q]$ and $B_p$ follow from Proposition~\ref{prop:bounded_algebraic_sampling} with $\nu = 4$ and $k = 3D+1$, applied to the polynomial $h$ in the proof of Lemma~\ref{lem:hard_feas_complexity}.
\end{proof}

Next, we bound the bit complexity of the ADP, starting with one branch.

\begin{lemma}
	\label{lem:one_branch_complexity_Gv}
	For a fixed branch of $\LinearOracle$, let $B_{\rm Gv}$ be the representation size of the Gram entries $G_{il}$ and objective values $v$ produced in $\SelectBranch$.
	Then
	\begin{align*}
		B_{\rm Gv} \le C_{\rm Gv}(d)(U + B_{\rm in})
	\end{align*}
	for some parameter-dependent constant $C_{\rm Gv}$.
\end{lemma}

\begin{proof}
	Fix one branch and one output $p$ of $\FixHardVariables$ for this branch.
	By Proposition~\ref{prop:fix_hard_variables_complexity}, the vector $p$ belongs to a field $\mk_p=\Q(\gamma)$ whose degree $[\mk_p:\Q]$ is bounded by a function of $d$, and the encoding sizes of the coordinates of $p$ are at most $C_p(d)B_{J_p}\le C(d)U$.
	The ADP performs only arithmetic operations and comparisons, so no further field extension is needed.
	
	To bound the encoding sizes of $G_{il}$ and $v$, let $Q_A$ be the product of the denominators of the entries of $A$, and let $Q_V= M 2^{J_0}$ be a common denominator of the numbers in free-variable finite set \eqref{eqn:free_variable_set}.
	Then
	\begin{align*}
		\log_2(Q_A) \le \mo(B_{\rm in}),
		\qquad
		\log_2(Q_V) \le C(d)U,
	\end{align*}
	for some parameter-dependent constant $C$.
	By Proposition~\ref{prop:fix_hard_variables_complexity} and arithmetic in $\mk_p$, the rational coordinate coefficients of all fixed hard-variable values and their pairwise products admit a common denominator $Q_H$ with $\log_2(Q_H)\le C(d)U$.
	We claim that the contributions to the Gram matrix and objective values have a common denominator
	\begin{align*}
		Q_*=Q_A Q_V^2 Q_H,
		\qquad
		\log_2(Q_*) \le C(d)(B_{\rm in}+U),
	\end{align*}
	and that the encoding sizes of the numerators satisfy the same bound.
	Indeed, every Gram entry and every objective value has the form
	\begin{align*}
		G_{il}=\sum_{j\in I}y_{ji}y_{jl},
		\qquad
		v=\sum_{j\in I}\sum_{i=1}^d A_{ji}y_{ji},
	\end{align*}
	for some subset $I\subseteq [n]$, where $y_{ji}$ denotes the $(j,i)$ entry of the partial solution, so $Q_*$ is a common denominator.
	These sums have at most $nd$ terms, so the encoding sizes of their numerators increase by at most $\mo(\log(nd))$, which is absorbed into $C(d)(B_{\rm in}+U)$.
	Since a merge combines disjoint sets of rows and $\KeepBest$ only selects among candidates, these bounds hold throughout the branch.
\end{proof}

\begin{proposition}
	\label{prop:ADP_complexity}
	There is an implementation of $\SelectBranch$ whose bit complexity $T_{\SelectBranch}$ satisfies
	\begin{align*}
		T_{\SelectBranch} \le F_{\SelectBranch}(d) n t^{-(d^2 + 3d + 2 + c)} (B_{\rm in} + 1)^{c} U^{d^2 + 5 d + 4 + 3c}
	\end{align*}
	for some parameter-dependent constant $F_{\SelectBranch}$ and an absolute constant $c$.
\end{proposition}

\begin{proof}
	Fix one branch and one output $p$ of $\FixHardVariables$ for this branch. 
	By Proposition~\ref{prop:fix_hard_variables_complexity}, the vector $p$ belongs to a field $\mk_p$ whose degree $[\mk_p:\Q]$ is bounded by a function of $d$.
	Every value computed by the ADP belongs to $\mk_p$.
	By Lemma~\ref{lem:one_branch_complexity_Gv}, every Gram entry $G_{il}$ and objective value $v$ has encoding size at most $C_{\rm Gv}(d)(B_{\rm in}+U)$, also after merges and dictionary updates.
	
	It remains to account for the rational discretization thresholds. 
	Let $l_{\max}$ be as in~\eqref{eqn:capacity_use}, and let $k_{\max}$ be the smallest integer $k$ with $\kappa_G (1+\theta)^k \ge 4d$.
	These are the largest exponents in the capacity and Gram-bin thresholds, and they satisfy
	\begin{align*}
		\max\{l_{\max}, k_{\max}\} \le C_{\rm exp}(d) t^{-1} U^2
	\end{align*}
	for some parameter-dependent constant $C_{\rm exp}$.
	Indeed, by Table~\ref{tab:parameters}, we have $H_{\rm tree} = \lceil \log_2(n) \rceil$, $\delta = t / (8(d+1))$, and $\tau = \delta / (128 d^2)$.
	The bound follows from $\log(1+x)\ge x / 2$ for $x\in (0, 1)$.
	Since the encoding sizes of $K_0$, $1+\alpha$, $1+\theta$, and $\kappa_G$ are at most $C_{\rm base}(d)(B_{\rm in}+U)$ for some parameter-dependent constant $C_{\rm base}$, all thresholds have encoding sizes at most
	\begin{align*}
		B_{\rm thre} := C_{\rm thre}(d)t^{-1} (B_{\rm in}+1) U^3
	\end{align*}
	for some parameter-dependent constant $C_{\rm thre}$.
	Consequently, arithmetic in $\mk_p$, comparisons, bin assignments, and dictionary operations can each be implemented in at most
	\begin{align*}
		F_{\rm work}(d)(B_{\rm thre}+1)^{c}
	\end{align*}
	bit operations, for some parameter-dependent constant $F_{\rm work}$ and an absolute constant $c$, by Lemma~\ref{lem:bit_complexity}.
	Here, by means of binary search, a Gram matrix bin assignment or a sparsity capacity usage rounding consists of a number of sign tests that is logarithmic in $k_{\max}$, $|U_c|$, and $1/\theta$.
	Each sign test compares an element of $\mk_p$ with a precomputed rational threshold and creates no new values, so Lemma~\ref{lem:bit_complexity} applies to each test with $L=1$.
	The number of these tests is at most $C_{\rm test}(d)U$ for some parameter-dependent constant $C_{\rm test}$, and it is absorbed by enlarging $F_{\rm work}$ and $c$.
	The construction of $\mv$ and of the thresholds is absorbed into the same bound.
	Finally, by \eqref{ub:branch_workload}, the ADP performs at most $W_{\rm ADP} \le f_{\rm bw}(d) n t^{-(d^2+3d+2)} U^{d^2+5d+4}$ operations.
	Multiplying this bound by the bit complexity of one operation concludes the proof.
\end{proof}

Next, we bound the bit complexity of $\LinearOracle$.
We first bound the cost of comparing the objective values of two candidates that may lie in different fields.

\begin{lemma}
	\label{lem:compare_obj_complexity}
	Let $x\in \mk_1$ and $y\in \mk_2$ with $[\mk_1 : \Q] \le \Delta_1$, $[\mk_2:\Q] \le \Delta_2$.
	Let $B_{\rm val}$ be an upper bound on the representation sizes of $x$ in $\mk_1$ and of $y$ in $\mk_2$.
	Then the sign of $x - y$ can be determined with bit complexity
	\begin{align*}
		F_{\rm cmp}(\Delta_1, \Delta_2)(B_{\rm val} + 1)^c
	\end{align*}
	for some parameter-dependent constant $F_{\rm cmp}$ and an absolute constant $c$. 
\end{lemma}

\begin{proof}
	The compositum $\mk_1 \mk_2$, that is, the smallest field that contains $\mk_1$ and $\mk_2$, has degree at most $\Delta_1 \Delta_2$~\cite[Chapter~V, Exercise~7]{Lan02}.
	One can compute a primitive element of $\mk_1\mk_2$ via resultants, its minimal polynomial via factorization~\cite{LenLenLov82}, and the representations of $x$ and $y$ in $\mk_1\mk_2$, as in the proof of Lemma~\ref{lem:bit_complexity} (see also \cite[Chapter~12]{BasPolRoy06}).
	Then, by sign determination~\cite[Section~10.3]{BasPolRoy06}, the sign of $x - y$ can be determined exactly in $F_{\rm cmp}(\Delta_1, \Delta_2)(B_{\rm val}+1)^c$ bit operations, for some parameter-dependent constant $F_{\rm cmp}$ and an absolute constant $c$. 
\end{proof}

\begin{proposition}
	\label{prop:linear_oracle_complexity}
	Let $B_{\rm in}$ be the encoding size of the input $(A, s, t)$, including the dimensions.
	There is an implementation of $\LinearOracle$ whose bit complexity $T_{\LinearOracle}$ satisfies
	\begin{align*}
		T_{\LinearOracle} \le F_{\LinearOracle}(d) n t^{-(d^3 - d^2 + 4d + 2 + c)} (B_{\rm in} + 1)^{c} U^{d^2 + 5d + 4 +3c}
	\end{align*}
	for some parameter-dependent constant $F_{\LinearOracle}$ and an absolute constant $c$.
\end{proposition}

\begin{proof}
    The cases $A = O$ and $d = 1$ can be handled within the stated bound.
    Thus, we assume that $A \ne O$ and $d\ge 2$.
	By \eqref{eqn:linear_oracle_raw_work}, the number of computational tasks of $\LinearOracle$ is at most
	\begin{align*}
		W_{\LinearOracle} \le f_{\rm lin}(d) n t^{-\beta_d} U^{\gamma_d},
	\end{align*}
	where $\beta_d := D_d+d^2+3d+2=d^3-d^2+4d+2$ and $\gamma_d := d^2+5d+4$.
	This bound includes the calls to $\FixHardVariables$ for all boxes $Q$ and the calls to $\ExactFinalize$ for all accepted root candidates.
	
	We first bound the input encoding size of $\ExactFinalize$.
	Consider an accepted candidate solution $Y$ reconstructed at the root of the ADP.
    By construction, $Y$ contains at most $D_d$ fixed hard-variable entries, while all remaining entries belong to the finite rational set $\mv$. 
    By Proposition~\ref{prop:fix_hard_variables_complexity}, the fixed hard-variable values belong to a common algebraic field whose degree is bounded in terms of $d$, and their encoding sizes are at most $C_p(d)B_{J_p}\le C_{\rm hv}(d)U$ for some parameter-dependent constant $C_{\rm hv}$.
    Moreover, every element of $\mv$ has encoding size at most $C_V(d)U$ for some parameter-dependent constant $C_V$.
    By Lemma~\ref{lem:one_branch_complexity_Gv}, the entries of $G^0=Y^\top Y$ have encoding sizes at most $C_{\rm Gv}(d)(B_{\rm in}+U)$.
    Since $B_{\rm in}$ is the total encoding size of the numerical input and $A$ has $n$ rows, we have $n\le B_{\rm in}$.
    Therefore, the total input encoding size required by $\ExactFinalize$ satisfies
    \begin{align*}
        B_{\rm tot-enc}
        \le C_0(d)(B_{\rm in}+nU)
        \le C_1(d)(B_{\rm in}+1)U,
    \end{align*}
    for some parameter-dependent constants $C_0$ and $C_1$.

	Next, by Propositions~\ref{prop:exact_finalize_complexity}, \ref{prop:fix_hard_variables_complexity}, and~\ref{prop:ADP_complexity}, together with Lemmas~\ref{lem:one_branch_complexity_Gv} and~\ref{lem:compare_obj_complexity}, the bit complexity of each computational task is at most
	\begin{align*}
		T_{\rm task}\le F_{\rm task}(d)(B_{\rm task}+1)^{c},
	\end{align*}
	for some parameter-dependent constant $F_{\rm task}$ and an absolute constant $c$, where $B_{\rm task}$ is the input encoding size of the task.
	By the proofs of these propositions, we have
	\begin{align*}
		B_{\rm task} \le C_{\rm task}(d)t^{-1}(B_{\rm in}+1)U^3
	\end{align*}
	for some parameter-dependent constant $C_{\rm task}$.
    Indeed, by Proposition~\ref{prop:fix_hard_variables_complexity}, the sampled hard variables have encoding sizes bounded by $C_2(d)U$.
    By Lemma~\ref{lem:one_branch_complexity_Gv}, the Gram entries and objective values in the ADP have encoding sizes bounded by $C_3(d)(B_{\rm in}+U)$. 
    Moreover, the proof of Proposition~\ref{prop:ADP_complexity} upper bounds the encoding sizes of the discretization thresholds by $C_4(d)t^{-1}(B_{\rm in}+1)U^3$.
    By the bound on $B_{\rm tot-enc}$ above, the input of $\ExactFinalize$ has total encoding size at most $C_1(d)(B_{\rm in}+1)U$.
    Here, $C_2$, $C_3$, and $C_4$ are parameter-dependent constants.
    These bounds give the bound on $B_{\rm task}$.
    
    Finally, by Propositions~\ref{prop:fix_hard_variables_complexity} and~\ref{prop:exact_finalize_complexity}, each candidate lies in a field whose degree is bounded in terms of $d$.
    By Lemma~\ref{lem:compare_obj_complexity}, comparisons between different candidates use temporary field extensions, while the better candidate is retained in its original field.
    Therefore, these comparisons do not accumulate field extensions.
	We conclude the proof by $T_{\LinearOracle} \le W_{\LinearOracle} T_{\rm task}$.
\end{proof}

Since $n\le B_{\rm in}$ and $U$ is logarithmic in $n$, $d$, and $1/t$, the bound in Proposition~\ref{prop:linear_oracle_complexity} is polynomial in $B_{\rm in}$ and $1/t$ for fixed $d$.
Together with Proposition~\ref{prop:linear_oracle_guarantee}, this proves Theorem~\ref{thm:linear_FPTAS}.

Finally, we prove Theorem~\ref{thm:main_alg_complexity}.

\begin{proof}[Proof of Theorem~\ref{thm:main_alg_complexity}]
	Algorithm~\ref{alg:main_alg} enumerates at most
	\begin{align*}
		M_W\le \left(2\left\lceil \frac{4rd}{\epsilon} \right\rceil + 1 \right)^{rd} \le \left(\frac{8rd + 3}{\epsilon}\right)^{rd}
	\end{align*}
	matrices $W$.
	For each $W$, the entries have the common denominator $J_W =\lceil 4rd/\epsilon \rceil$.
	To bound the total encoding size $B_A$ of $A = RW$, write each entry of $R$ as $R_{jk}=a_{jk}/b_{jk}$, where $a_{jk}\in\mathbb{Z}$ and $b_{jk}$ is a positive integer. 
    Since every entry of $W$ has denominator $J_W$, we can write
    \begin{align*}
        A_{ji}=\frac{1}{J_W} \sum_{k=1}^{r}\frac{a_{jk}w_{ki}}{b_{jk}},
    \end{align*}
    where $w_{ki} = J_W W_{ki}\in\Z$ and $|w_{ki}|\le J_W$.
    Summing the encoding sizes over all rows and columns gives
    \begin{align*}
        B_A \le C_0(r,d) \left( B_{\rm SPCA} +n( 1+\log_2(J_W + 1) ) \right)
        \le C_{\rm SPCA}(r,d) (B_{\rm SPCA}+1) U_\epsilon,
    \end{align*}
    for some parameter-dependent constants $C_0$ and $C_{\rm SPCA}$, where the last inequality follows from $n\le B_{\rm SPCA}$ and $\log_2(J_W+1)\le C_1(r,d)U_\epsilon$ for some parameter-dependent constant $C_1$.
	Hence the input $(A,s,t)$ of $\LinearOracle$ has encoding size $B_{\rm in}\le B_A+B_{\rm SPCA}+\mo(U_\epsilon)$.

	Applying Proposition~\ref{prop:linear_oracle_complexity} with $t=\epsilon/4$, each call to $\LinearOracle$ has bit complexity at most
    \begin{align*}
        F_{\LinearOracle}'(r,d) n \epsilon^{-(d^3 - d^2 + 4d + 2+c)} (B_{\rm SPCA} + 1)^{c} U_\epsilon^{d^2 + 5d + 4 + 4c},
    \end{align*}
	for some parameter-dependent constant $F_{\LinearOracle}'$ and an absolute constant $c$.
	Computing $RW$, obtaining the returned objectives, and comparing them exactly using Lemma~\ref{lem:compare_obj_complexity} are dominated by this bound.
	The value $\fnorm{R^\top X}^2$ is computed from the $rd$ sums $\sum_{j=1}^n R_{jk}X_{ji}$, each with $n$ terms.
	Clearing denominators as in the proof of Lemma~\ref{lem:one_branch_complexity_Gv}, their encoding sizes are polynomial in $B_{\rm SPCA}$ and $U_\epsilon$, because $n\le B_{\rm SPCA}$.
	Indeed, by Proposition~\ref{prop:fix_hard_variables_complexity} and Lemma~\ref{lem:exact_finalize_encoding}, the degrees of the field extensions are bounded in terms of $d$, and by Lemma~\ref{lem:exact_finalize_encoding} and the bound on $B_{\rm tot-enc}$ in the proof of Proposition~\ref{prop:linear_oracle_complexity}, the encoding sizes are bounded by a polynomial in $B_{\rm SPCA}$ and $U_\epsilon$.
	Thus, after enlarging the parameter-dependent constant and the absolute constant $c$ if necessary, these additional computations are absorbed into the stated bound.
	Multiplying by $M_W$ concludes the proof.
\end{proof}

\paragraph{Dependence on $n$.}
The proof of Theorem~\ref{thm:main_alg_complexity} gives a finer bound on the dependence on $n$.
For fixed $r$ and $d$, Algorithm~\ref{alg:main_alg} processes at most $F(r,d)\, n\, \epsilon^{-(rd + d^3 - d^2 + 4d + 2)} U_{\epsilon}^{d^2 + 5d + 4}$ states, merges, and finalization steps, by \eqref{eqn:linear_oracle_raw_work} and the bound on the number of matrices $W$.
Each of these steps uses a number of arithmetic operations and sign tests, in a number field whose degree is bounded in terms of $d$, that is at most a function of $d$ times $U_\epsilon$.
The number of square-root extensions in $\ExactFinalize$ and of exact comparisons of candidates in different fields is polylogarithmic in $n$.
Thus the number of steps is linear in $n$ up to a polylogarithmic factor.
For comparison, a straightforward dynamic program that processes the rows one at a time would need exact sparsity capacity usages of order $n^d$ and a Gram grid of mesh of order $\epsilon/(nd^2)$, since its rounding errors accumulate over all $n$ rows.
This gives a number of states of order $n^{d+d(d+1)/2}$ at each row.